\def\year{2019}\relax
\documentclass[letterpaper]{article} 
\usepackage{aaai19}  
\usepackage{times}  
\usepackage{helvet}  
\usepackage{courier}  
\usepackage{url}  
\usepackage{graphicx}  
\usepackage{url}            
\usepackage{booktabs}       
\usepackage{amsfonts}       
\usepackage{nicefrac}       
\usepackage{microtype}      
\usepackage{graphicx}
\usepackage{placeins}
\usepackage{tikz,graphics,color,float,epsf,caption,subcaption,fullpage}
\usepackage{amsmath,amssymb}
\usepackage{listings}
\usepackage{hhline}
\usepackage{multirow}
\usepackage{placeins}
\usepackage{multicol}
\usepackage{lipsum}
\usepackage{bbm}
\usepackage{breqn}
\usepackage[algoruled,boxed,lined]{algorithm2e}
\usepackage[export]{adjustbox}
\usepackage{wrapfig}
\graphicspath{ {images/} }

\newcommand{\specialcell}[2][l]{%
  \begin{tabular}[#1]{@{}l@{}}#2\end{tabular}}
\DeclareMathOperator*{\argmax}{arg\,max}

\begin{document}
\title{Towards Better Interpretability in Deep Q-Networks}
\author{
  Raghuram Mandyam Annasamy\\
  Carnegie Mellon University\\
  \texttt{rannasam@cs.cmu.edu} \\
  \And
  Katia Sycara\\
  Carnegie Mellon University\\
  \texttt{katia@cs.cmu.edu}}

\maketitle
\begin{abstract}
 Deep reinforcement learning techniques have demonstrated superior performance in a wide variety of environments. As improvements in training algorithms continue at a brisk pace, theoretical or empirical studies on understanding what these networks seem to learn, are far behind. In this paper we propose an interpretable neural network architecture for Q-learning which provides a global explanation of the model's behavior using key-value memories, attention and reconstructible embeddings. With a directed exploration strategy, our model can reach training rewards comparable to the state-of-the-art deep Q-learning models. However, results suggest that the features extracted by the neural network are extremely shallow and subsequent testing using out-of-sample examples shows that the agent can easily overfit to trajectories seen during training.
\end{abstract}

\section{Introduction}
The last few years have witnessed a rapid growth of research and interest in the domain of deep Reinforcement Learning (RL) due to the significant progress in solving RL problems \cite{arulkumaran2017brief}. Deep RL has been applied to a wide variety of disciplines ranging from game playing, robotics, systems to natural language processing and even biological data \cite{silver2017mastering,mnih2015human,levine2016end,kraska2018case,williams2017hybrid,choi2017coarse}. However, most applications treat neural networks as a black-box and the problem of understanding and interpreting deep learning models remains a hard problem. This is even more understudied in the context of deep reinforcement learning and only recently has started to receive attention. Commonly used visualization methods for deep learning such as saliency maps and t-SNE plots of embeddings have been applied to deep RL models \cite{greydanus2017visualizing,zahavy2016graying,mnih2015human}. However, there are a few questions over the reliability of saliency methods including, as an example, sensitivity to simple transformations of the input \cite{kindermans2017reliability}. The problem of generalization and memorization with deep RL models is also important. Recent findings suggest that deep RL agents can easily memorize large amounts of training data with drastically varying test performance and are vulnerable to adversarial attacks \cite{zhang2018study,zhang2018dissection,huang2017adversarial}. 

In this paper, we propose a neural network architecture for Q-learning using key-value stores, attention and constrained embeddings, that is easier to study than the traditional deep Q-network architectures. This is inspired by some of the recent work on Neural Episodic Control (NEC) \cite{pritzel2017neural} and distributional perspectives on RL \cite{bellemare2017distributional}. We call this model i-DQN for Interpretable DQN and study latent representations learned by the model on standard Atari environments from Open AI gym \cite{openai_gym_atari}.  Most current work around interpretability in deep learning is based on local explanations i.e. explaining network predictions for specific input examples \cite{lipton2016mythos}. For example, saliency maps can highlight important regions of the input that influence the output of the neural network. In contrast, global explanations attempt to understand the mapping learned by a neural network regardless of the input. We achieve this by constraining the latent space to be reconstructible and inverting embeddings of representative elements in the latent space (keys). This helps us understand aspects of the input space (images) that are captured in the latent space across inputs. Our visualizations suggest that the features extracted by the convolutional layers are extremely shallow and can easily overfit to trajectories seen during training. This is in line with the results of \cite{zhang2018study} and \cite{zhang2018dissection}. Although our main focus is to understand learned models, it is important that the models we analyze perform well on the task at hand. To this end, we show our model achieves training rewards comparable to Q-learning models like Distributional DQN \cite{bellemare2017distributional}. Our contribution in this work is threefold:
\begin{itemize}
\item We explore a different neural network architecture with key-value stores, constrained embeddings and an explicit soft-assignment step that separates representation learning and Q-value learning (state aggregation).
\item We show that such a model can improve interpretability in terms of visualizations of the learned keys (cluster), attention maps and saliency maps. Our method attempts to provide a global explanation of the model's behavior instead of explaining specific input examples (local explanations).  We also develop a few examples to test the generalization behavior.
\item We show that the model's uncertainty can be used to drive exploration that reaches reasonably high rewards with reduced sample complexity (training examples) on some of the Atari environments.
\end{itemize}

\begin{figure*}[t!]
\centering
\includegraphics[width=0.8\textwidth]{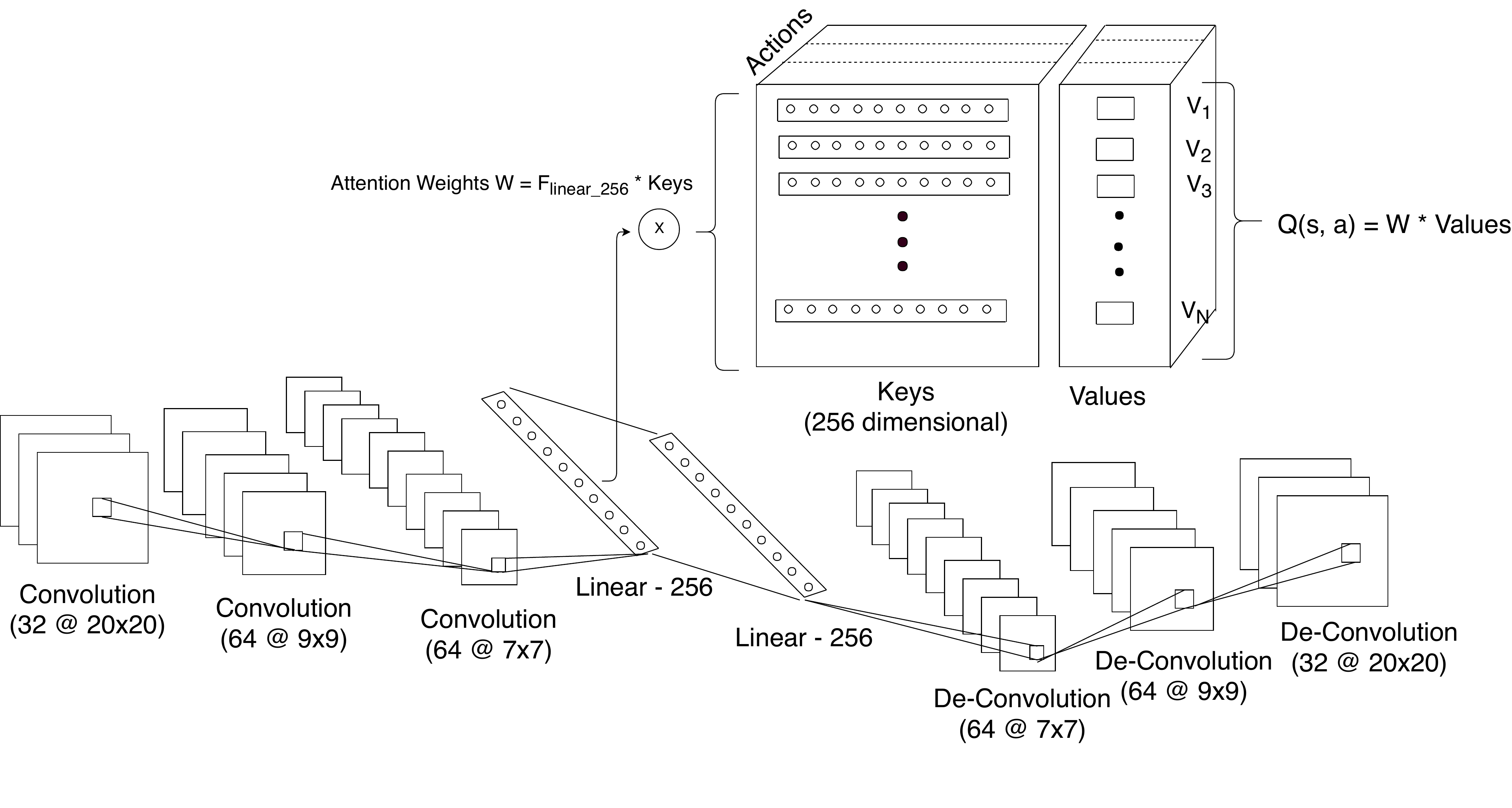}
\caption{Model Architecture- Interpretable DQN (i-DQN)}
\label{fig:arch}
\end{figure*}

\section{Related Work}
Many attempts have been made to tackle the problem of interpretability with deep learning, largely in the supervised learning case. \cite{zhang2018visual} carry out an in-depth survey on interpretability with Convolutional Neural Networks (CNNs). Our approach to visualizing embeddings is in principle similar to the work of \cite{dosovitskiy2016inverting} on inverting visual representations. They train a neural network with deconvolution layers using HOG, SIFT and AlexNet embeddings as input and their corresponding real images as ground truth (the sole purpose of this network being visualization). Saliency maps are another popular type of method that generate local explanations which generally use gradient-like information to identify salient parts of the image. The different ways of computing saliency maps are covered exhaustively in \cite{zhang2018visual}. Few of these have been applied in the context of deep reinforcement learning. \cite{zahavy2016graying} use the Jacobian of the network to compute saliency maps on a Q-value network. Perturbation based saliency maps using a continuous mask across the image and also using object segmentation based masks have been studied in the context of deep-RL \cite{greydanus2017visualizing,iyer2018transparency,odrl}. In contrast to these approaches, our method is based on a global view of the network. Given a particular action and expected returns, we invert the corresponding key to try and understand visual aspects being captured by the embedding regardless of the input state. More recently, \cite{verma2018programmatically} introduce a new method that finds interpretable programs that can best explain the policy learned by a neural network- these programs can also be treated as global explanations for the policy networks.

Architecturally, our network is similar to the network first proposed by \cite{pritzel2017neural}. The authors describe their motivation as speeding up the learning process using a semi-tabular representation with Q-value calculations similar to the tabular Q-learning case. This is to avoid the inherent slowness of gradient descent and reward propagation. Their model learns to attend over a subset of states that are similar to the current state by tracking all the states recently seen (up-to half-million states) using a k-d tree. However, their method does not have any notion of clustering or fixed Q-values. Our proposed method is also similar to \citeauthor{bellemare2017distributional}'s work on categorical/distributional DQN. The difference is that in our model the cluster embeddings (keys) for different Q-values are accessible freely (for analysis and visualization) because of the explicit soft-assignment step, whereas it is almost impossible to find such representations while having fully-connected layers like in \cite{bellemare2017distributional}. Although we do not employ any iterative procedure (like refining keys; we train fully using backpropagation), works on combining deep embeddings with unsupervised clustering methods \cite{xie2016unsupervised,chang2017deep} (joint optimization/iterative refinement) have started to pick up pace and show better performance compared to traditional clustering methods.

Another important direction that is relevant to our work is that of generalizing behavior of neural networks in the reinforcement learning setting. \cite{henderson2017deep} discuss in detail about general problems of deep RL research and evaluation metrics used for reporting. \cite{zhang2018study,zhang2018dissection} perform systematic experimental studies on various factors affecting generalization behavior such as diversity in training seeds and randomness in environment rewards. They conclude that deep RL models can easily overfit to random reward structures or when there is insufficient training diversity and careful evaluation techniques (such as isolated training and testing seeds) are needed.





\section{Proposed Method}
We follow the usual RL setting and assume the environment can be modelled as a Markov Decision Process (MDP) represented by the 5-tuple $(S, A, T, R, \gamma)$, where $S$ is the state space, $A$ is the action space, $T(s'|s,a)$ is the state transition probability function, $R(s,a)$ is the reward function and $\gamma \in [0,1)$ is the discount factor. A policy $\pi: S \rightarrow A$ maps every state to a distribution over actions. The value function $V^{\pi}(s_t)$ is the expected discounted sum of rewards by following policy $\pi$ from state $s_t$ at time $t$, $V^{\pi}(s_t) = E[\sum_{i=0}^T \gamma^i r_{t+i}]$. Similarly, the Q-value (action-value) $Q^{\pi}(s_t,a)$ is the expected return starting from state $s_t$, taking action $a$ and then following $\pi$. Q-value function can be recursively estimated using the Bellman equation $Q^{\pi}(s_t,a) = E[r_t + \gamma \max_{a'} Q(s_{t+1}, a')]$ and $\pi^*$ is the optimal policy which achieves the highest $Q^{\pi}(s_t,a)$ over all policies $\pi$.
\begin{figure*}[t]
\centering
\begin{subfigure}[t]{0.42\textwidth}
  \centering
    \includegraphics[width=1.0\textwidth]{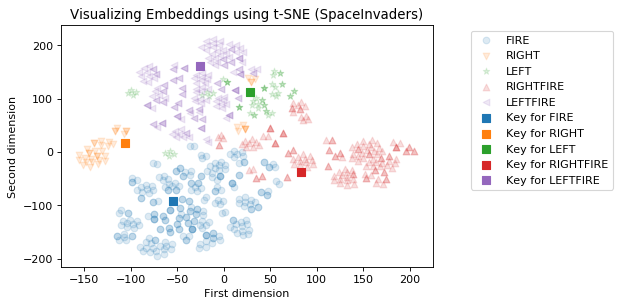}
    \caption{SpaceInvaders}
\end{subfigure}%
\begin{subfigure}[t]{0.42\textwidth}
  \centering
    \includegraphics[width=1.0\textwidth]{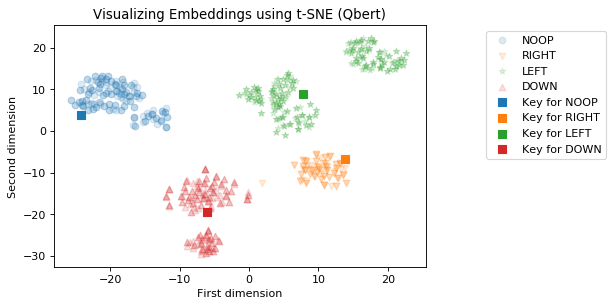}
    \caption{Qbert}
\end{subfigure}
\begin{subfigure}[t]{0.42\textwidth}
  \centering
    \includegraphics[width=1.0\textwidth]{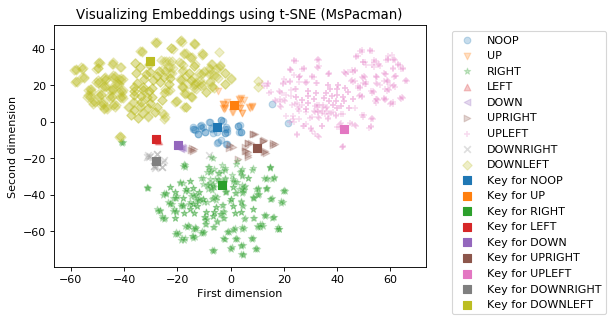}
    \caption{MsPacman}
    \label{fig:tsne-pacman}
\end{subfigure}%
\begin{subfigure}[t]{0.42\textwidth}
  \centering
    \includegraphics[width=1.0\textwidth]{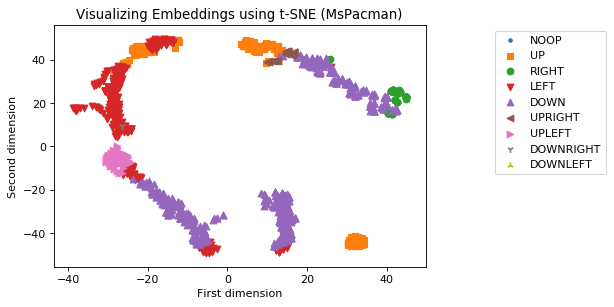}
    \caption{MsPacman, DDQN}
    \label{fig:tsne-dqn}
\end{subfigure}
\caption{Visualizing keys, state embeddings using t-SNE: i-DQN, Q-value 25 (a)-(c); Double DQN(d)}
\label{fig:tsne}
\end{figure*}

Similar to the traditional DQN architecture \cite{mnih2015human}, any state $s_t$ (a concatenated set of input frames) is encoded using a series of convolutional layers each followed by a non-linearity and finally a fully-connected layer at the end $ h(s_{t}) = Conv(s_{t}) $. This would usually be followed by some non-linearity and a fully-connected layer that outputs Q-values. Instead, we introduce a restricted key-value store over which the network learns to attend as shown in Figure~\ref{fig:arch}.

Intuitively, the model is trying to learn two things. First, it learns a latent representation $h(s_t)$ that captures important visual aspects of the input images. At the same time, the model also attempts to learn an association between embeddings of states $h(s_t)$ and embeddings of keys in the key-value store. This would help in clustering the state (around the keys) based on the scale of expected returns (Q-values) from that state. We can think of the $N$ different keys $h^a$ (for a given action $a$) weakly as the cluster centers for the corresponding Q-values, attention weights $w^a(s_t)$ as a soft assignment between embeddings for current state $h(s_t)$ and embeddings for different Q-values $\{h_1^a, h_2^a, \cdots h_N^a\}$. This explicit association step helps us in understanding the model in terms of attention maps and visualizations of the cluster centers (keys). 

The key-value store is restricted in terms of size and values of the store. Each action $a \in A$ has a fixed number of key-value pairs (say $N$) and the value associated with every key is also held constant. $N$ values $\{v_1, v_2, \cdots, v_N\}$ are sampled uniformly at random from $(V_{min}, V_{max})$ (usually $(-25, 25)$) once and the same set of values are used for all the actions. All of the keys ($N \times A$) are also initialized randomly. To compute attention, the embeddings $h(s_t)$ for a state $s_t$ are compared to all the keys $\{h_1^a, h_2^a, \cdots, h_N^a\}$ in the store for a particular action $(a)$ using a softmax over their dot products.

\begin{align}
  w(s_t)_i^a = \dfrac{\exp{(h(s_{t})\cdot h_i^a)}}{\sum_{j} \exp{(h(s_{t})\cdot h_j^a)}} \label{attention_eq}
\end{align}

These attention weights over keys and their corresponding value terms are then used to calculate the Q-values.
\[
  Q(s_{t}, a) = \sum_i w_i^a(s_t) v_i
\]
Now that we have Q-values, we can define the different losses that can be used to train the network,
\begin{itemize}
    \item Bellman Error ($\mathcal{L}_{bellman}$): The usual value function estimation error. 
    \[ \mathcal{L}_{bellman}(\theta) = \big(Q(s_t, a, \theta) - Y_t\big)^2 \]
    where $Y_t = R(s_t,a,s_{t+1}) + \gamma \max_a' Q(s_{t+1}, a', \theta)$
    \item Distributive Bellman Error ($\mathcal{L}_{distrib}$): We force the distributive constraint on attention weights between current and next states similar to \cite{bellemare2017distributional} using values $\{v_1, v_2, \cdots, v_N\}$ as supports of the distribution. The distributive loss is defined as the KL divergence between $\phi \mathcal{T} w(s_{t+1})^{a^*} $ and $w(s_t)^a$ where $\mathcal{T}$ is the distributional Bellman operator and $\phi$ is the projection operator and $a^*$ is best action at state $s_{t+1}$ i.e. $a^* = \argmax_a Q(s_{t+1}, a)$.
    \begin{align}
    \mathcal{L}_{distrib}(\theta) =& \mathcal{D}_{KL} (\phi \mathcal{T} w(s_{t+1})^{a^*}, w(s_t)^a) \\
       =& - \sum_i \phi \mathcal{T} w(s_{t+1})^{a^*}_i \cdot  w(s_t)^a_i \label{distrib_eq}
    \end{align}
    Equation (\ref{distrib_eq}) is simply the cross entropy loss (assuming $w(s_{t+1})^{a^*}$ to be constant with respect to $\theta$, similar to the assumption for $Y_t$ in Bellman error).
    
    \item Reconstruction Error ($\mathcal{L}_{reconstruct}$): We also constrain the embeddings $h(s_t)$ for any state to be reconstructible. This is done by transforming $h(s_t)$ using a fully-connected layer and then followed by a series of non-linearity and deconvolution layers.
    \begin{align}
        h^{dec}(s_t) = W^{dec}h(s_t) \label{deconv_eq}
    \end{align}
    \[
        \hat{s}_t = Deconv(h^{dec}(s_t))
    \]
    The mean squarred error between reconstructed image $\hat{s}_t$ and original image $s_t$ is used,
    \[
        \mathcal{L}_{reconstruct}(\theta) = \dfrac{1}{2} || \hat{s}_t - s_t ||^2_2
    \]
    \item Diversity Error ($\mathcal{L}_{diversity}$): The diversity error forces attention over different keys in a batch. This is important because training can collapse early with the network learning to focus on very few specific keys (because both the keys and attention weights are being learned together). We could use KL-divergence between the attention weights but \cite{lin2017structured} develop an elegant solution to this in their work.
    \[
        \mathcal{L}_{diversity}(\theta) = ||(AA^T - I)||^2
    \]
    where $A$ is a 2D matrix of size (batch size, $N$) and each row of $A$ is the attention weight vector $w(s_t)^a$. It drives $AA^T$ to be diagonal (no overlap between keys attended to within a batch) and $l$-2 norm of $w$ to be $1$. Because of softmax, the $l$-1 norm is also $1$ and so ideally the attention must peak at exactly one key however in practice it spreads over as few keys as possible.
Finally, the model is trained to minimize a weighted linear combination of all the four losses.
\begin{dmath*}
\mathcal{L}_{final}(\theta) = \lambda_1 \mathcal{L}_{bellman}(\theta) + \lambda_2 \mathcal{L}_{distrib}(\theta) + \lambda_3 \mathcal{L}_{reconstruct}(\theta)  + \lambda_4 \mathcal{L}_{diversity}(\theta)
\end{dmath*}
\end{itemize} 

\begin{table*}[tb!]
  \begin{center}
  {\small
  \begin{tabular}{p{2.5cm}|cc|ccccc}
    \hline
    \multirow{2}{2.5cm}{\textbf{Environment}} & \multicolumn{2}{c|}{\textbf{10M frames}} & \multicolumn{5}{c}{\textbf{Reported Scores (final)}}\\
    & \textbf{DDQN} & \textbf{i-DQN} & \textbf{DDQN} & \textbf{Distrib. DQN} & \textbf{Q-ensemble} & \textbf{Bootstrap DQN} & \textbf{Noisy DQN} \\
    \hline
    Alien & 1,533.45 & 2,380.72 & 3,747.7 & \textbf{4,055.8} & 2,817.6 & 2,436.6 & 2,394.90 \\
    Freeway & 22.5 & 28.79 & 33.3 & 33.6 & \textbf{33.96} & 33.9 & 32\\
    Frostbite & 754.48 & \textbf{3,968.45} & 1,683.3 & 3,938.2 & 1,903.0 & 2,181.4 & 583.6 \\ 
    Gravitar & 279.89	& 517.33 & 412 & \textbf{681} & 318 & 286.1 & 443.5  \\
    MsPacman & 2,251.43 & \textbf{6,132.21} & 2,711.4 & 3,769.2 & 3,425.4 & 2,983.3 & 2,501.60\\
    Qbert & 10,226.93 & \textbf{19,137.6} & 15,088.5 & 16,956.0 & 14,198.25 & 15,092.7 & 15,276.30 \\ 
    Space Invaders & 563.2 & 979.45 & 2,525.5 & \textbf{6,869.1} & 2,626.55  & 2,893 & 2,145.5\\ 
    Venture & 70.87 & 985.11 & 98 & \textbf{1,107.0} & 67 & 212.5 & 0  \\ \hline
  \end{tabular}
  }%
  \end{center}
  \caption{Training scores (averaged over 100 episodes, 3 seeds). Scores for Double DQN , Distributional DQN and Noisy DQN are from \cite{hessel2017rainbow}; Scores for Bootstrap-DQN are as reported in the original paper \cite{osband2016deep}; Scores for UCB style exploration with Q-ensembles are from \cite{chen2018ucb}}
  \label{table:heuristic_scores}
\end{table*}

\section{Experiments and Discussions}
We report the performance of our model on eight Atari environments \cite{openai_gym_atari}- Alien, Freeway, Frostbite, Gravitar, MsPacman, Qbert, SpaceInvaders,  and Venture, in Table~\ref{table:heuristic_scores}. Using the taxonomoy of Atari games from \cite{bellemare2016unifying} and \cite{ostrovski2017count}, seven of the eight environments tested (all except SpaceInvaders) are considered hard exploration problems. Additionally, three of them (Freeway, Gravitar and Venture) are hard to explore because of sparse rewards. Since our focus is on interpretability, we do not carry out an exhaustive performance comparison. We simply show that training rewards achieved by i-DQN model are comparable to some of the state-of-the-art models. This is important because we would like our deep-learning models to be interpretable but also remain competitive at the same time.  We look at scores against other exploration baselines for Q-learning that do not involve explicit reward shaping/exploration bonuses- Bootstrap DQN \cite{osband2016deep}, Noisy DQN \cite{fortunato2017noisy} and Q-ensembles \cite{chen2018ucb}.

\subsection{Directed exploration}
We use the uncertainty in attention weights to drive exploration during training. $U(s_t, a)$ is an approximate upper confidence on the Q-values. Similar to \cite{chen2018ucb} we select the action maximizing a UCB style confidence interval,
{
\begin{align*}
Q(s_{t}, a) =& \sum_i w_i^a(s_t) v_i \\
U(s_t, a) =& \sqrt{Q(s_t, a)^2 - \sum_i w_i^a(s_t) v_i^2} \\
a_t =& \argmax_{a\in A} Q(s_t, a) + \lambda_{\text{exp}}\; U(s_t, a)
\end{align*}
}
Table~\ref{table:heuristic_scores} compares i-DQN's performance (with directed exploration) against a baseline Double DQN implementation (which uses epsilon-greedy exploration) at 10M frames. Double DQN (DDQN), Distributional DQN (Distrib. DQN), Bootstrap DQN and Noisy DQN agents are trained for up to 50M steps which translates to 200M frames \cite{hessel2017rainbow,fortunato2017noisy,osband2016deep}. The Q-ensemble agent using UCB-style exploration is trained for up to 40M frames \cite{chen2018ucb}. We see that on some of the games, our model reaches higher training rewards within 10M frames compared to Double DQN, Distributional DQN  models. Also, our model is competitive with the final scores reported by other exploration baselines like Bootstrap DQN, Noisy DQN and Q-ensembles, and and even performs better on some environments (5 out of 8 games). The training time for i-DQN is roughly 2x slower because of the multiple loss functions compared to our implementation of Double DQN.

\subsection{What do the keys represent?}
The keys are latent embeddings (randomly initialized) that behave like cluster centers for the particular action-return pairs (latent space being $\mathbb{R}^{256}$). Instead of training using unsupervised methods like K-means or mixture models, we use the neural network itself to find these points using gradient descent. For example, the key for action right; Q-value 25 (Figure \ref{fig:tsne-pacman}) is a cluster center that represents the latent embeddings for all states where the agent expects a return of 25 by selecting action right. These keys partition the latent space into well formed clusters as shown in Figure~\ref{fig:tsne}, suggesting that embeddings also contain action-specific information crucial for an RL agent. On the other hand, Figure~\ref{fig:tsne-dqn} shows embeddings for DDQN which are not easily separable (similar to the visualizations in \cite{mnih2015human}). Since we use simple dot-product based distance for attention, keys and state embeddings must lie in a similar space and this can be seen in the t-SNE visualization i.e. keys (square boxes) lie within the state embeddings (Figure~\ref{fig:tsne}). The fact that the keys lie close to their state embeddings is essential to interpretability because state embeddings satisfy reconstructability constraints.
\begin{figure}[!t]
\centering
\begin{subfigure}[t]{0.12\textwidth}
  \centering
    \includegraphics[width=1.0\textwidth]{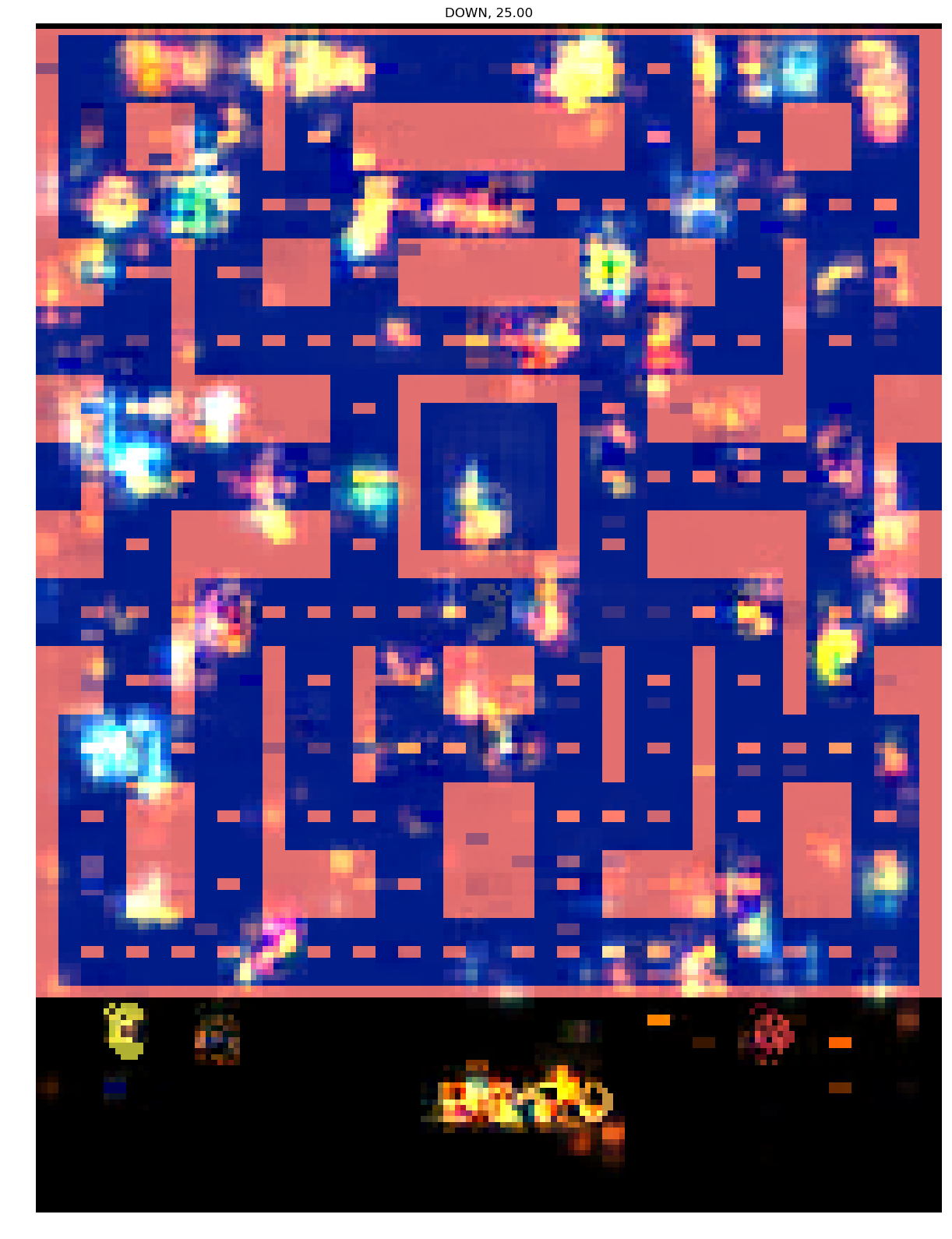}
    \caption{Down}
\end{subfigure} \hspace*{-0.6em}
\begin{subfigure}[t]{0.12\textwidth}
  \centering
    \includegraphics[width=1.0\textwidth]{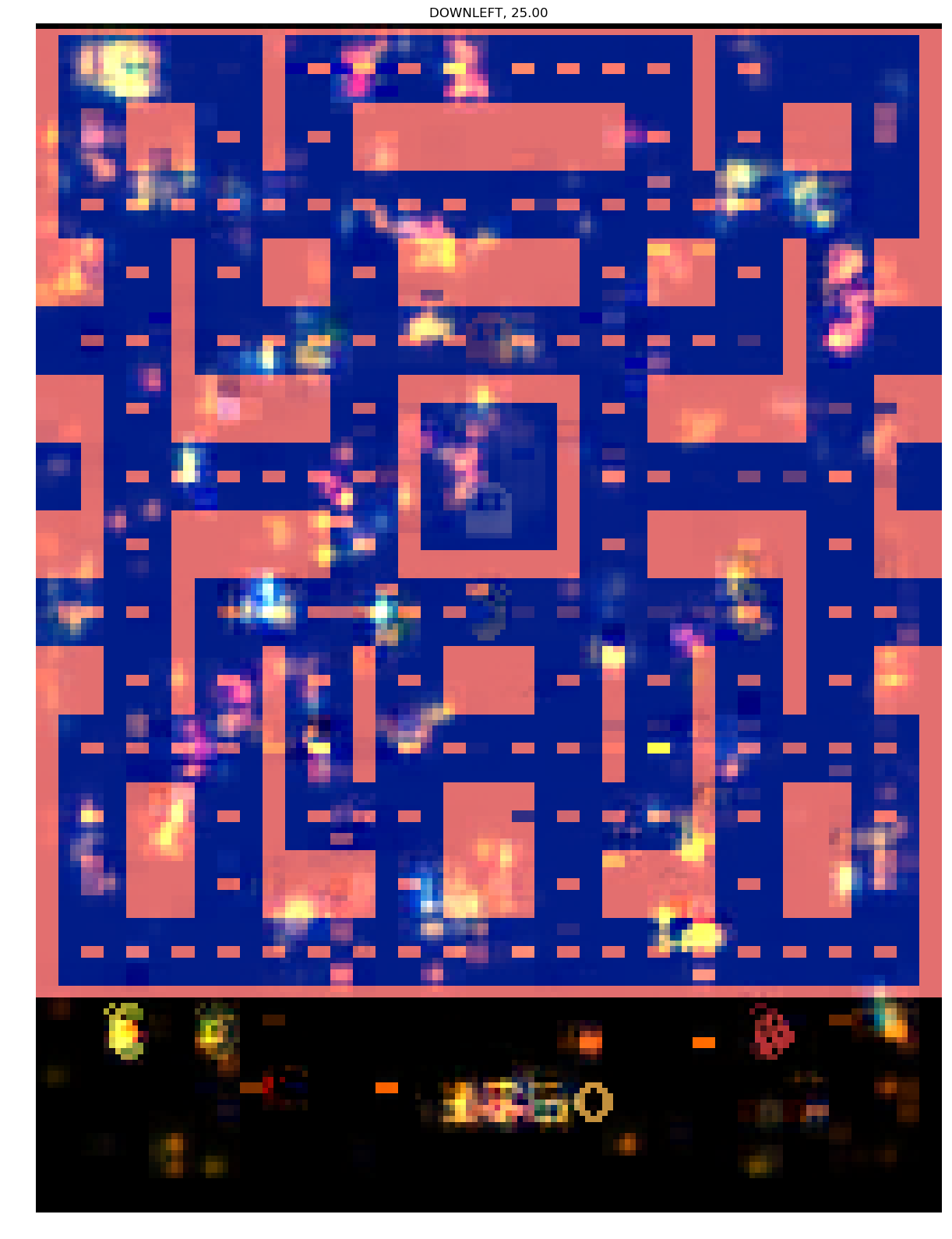}
    \caption{Downleft}
\end{subfigure} \hspace*{-0.6em}
\begin{subfigure}[t]{0.12\textwidth}
  \centering
    \includegraphics[width=1.0\textwidth]{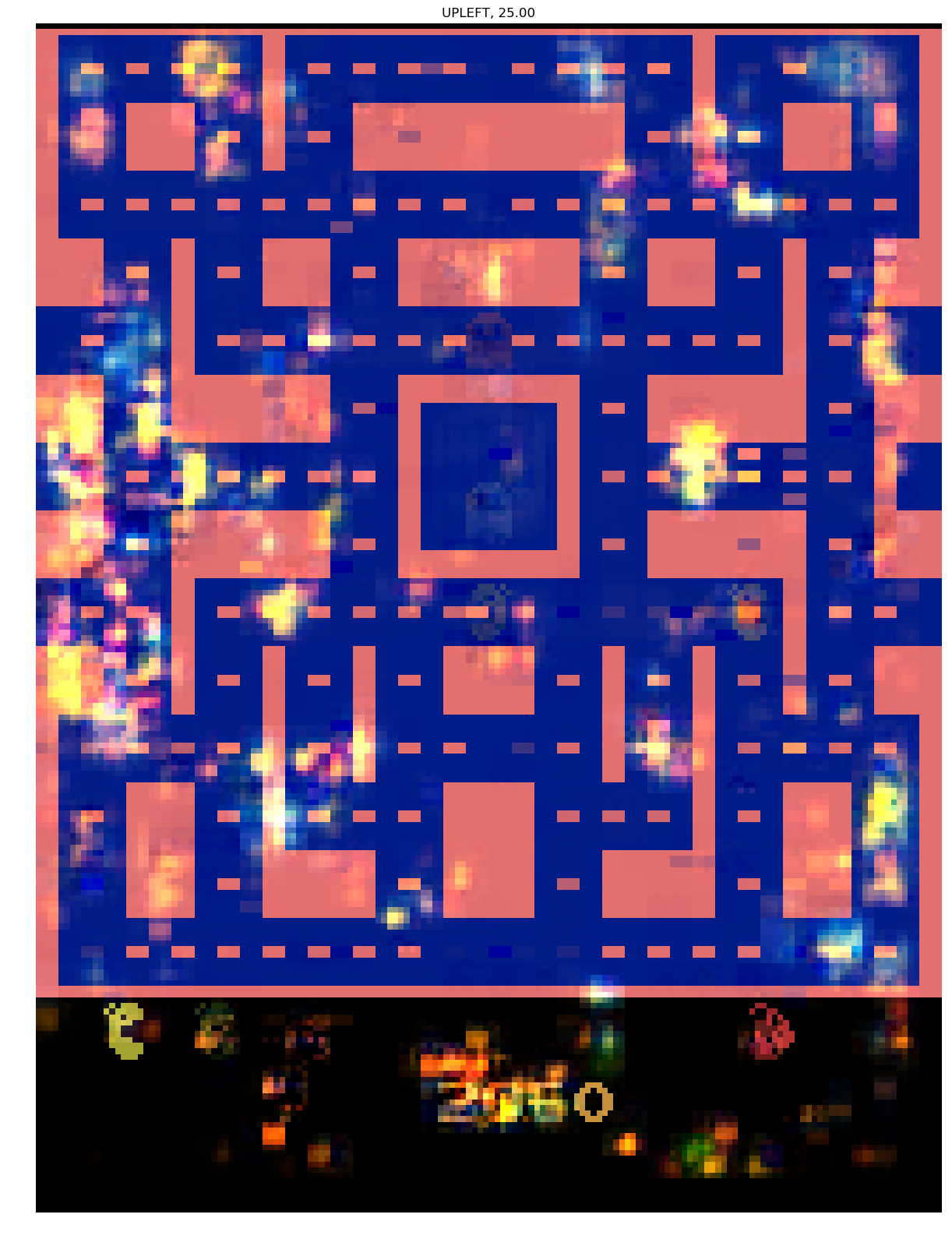}
    \caption{Upleft}
\end{subfigure} \hspace*{-0.6em}
\begin{subfigure}[t]{0.12\textwidth}
  \centering
    \includegraphics[width=1.0\textwidth]{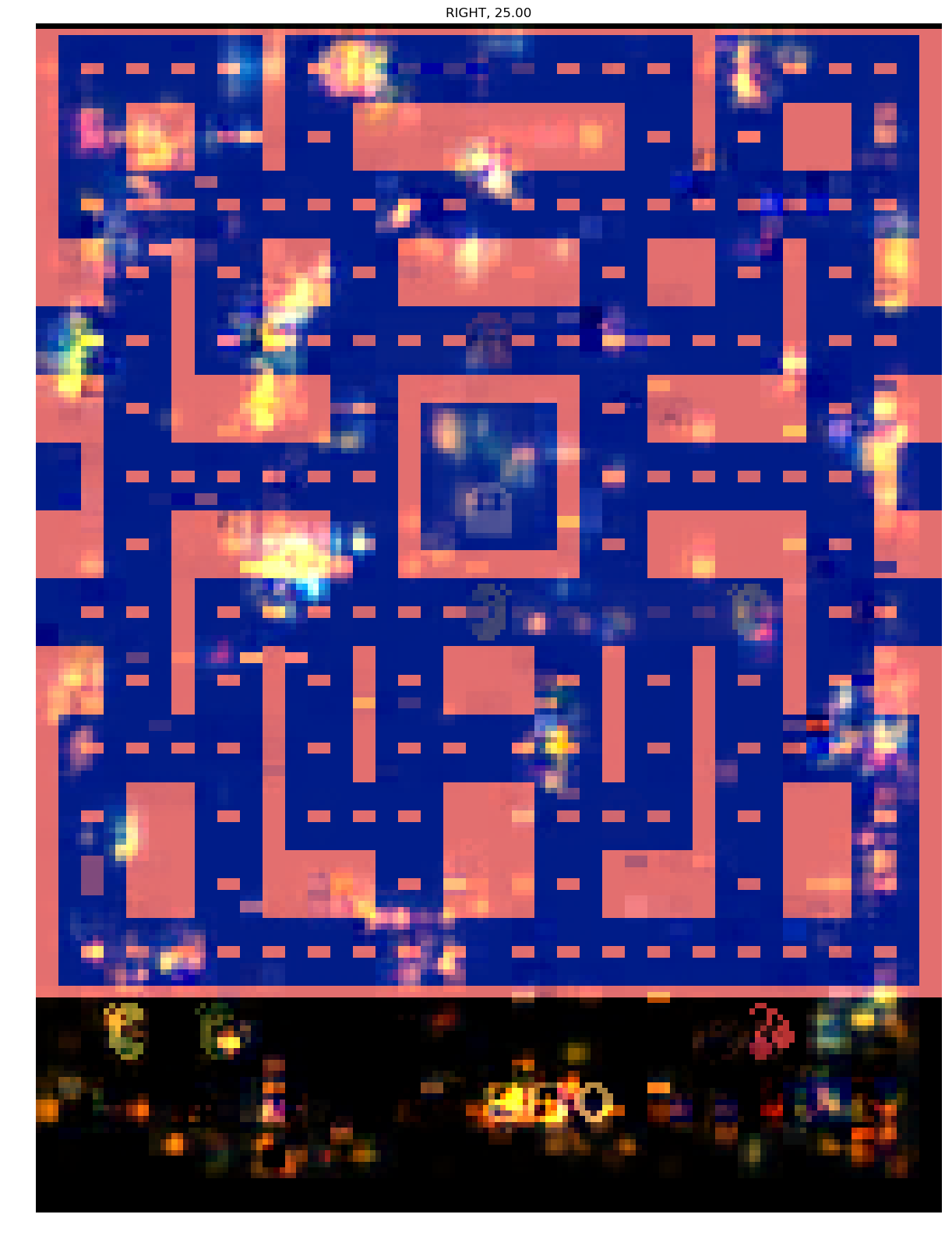}
    \caption{Right}
    \label{fig:invert-key-25-pacman-right}
\end{subfigure}
\caption{MsPacman, Inverting keys for Q-value 25}
\label{fig:invert-key-25-pacman}
\begin{subfigure}[t]{0.10\textwidth}
  \centering
    \includegraphics[width=1.0\textwidth]{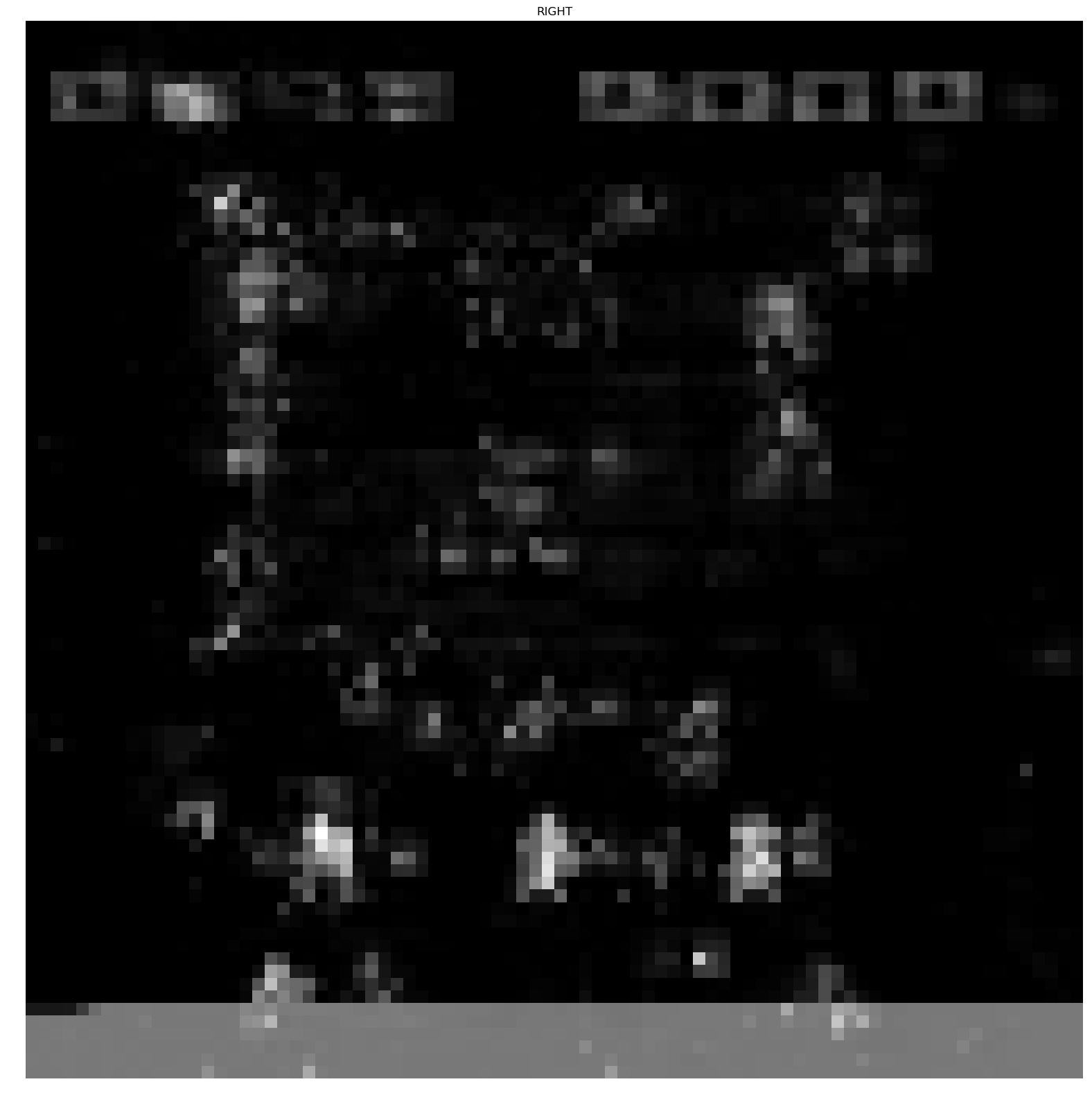}
    \caption{Right}
\end{subfigure} \hspace*{-1em}
\begin{subfigure}[t]{0.10\textwidth}
  \centering
    \includegraphics[width=1.0\textwidth]{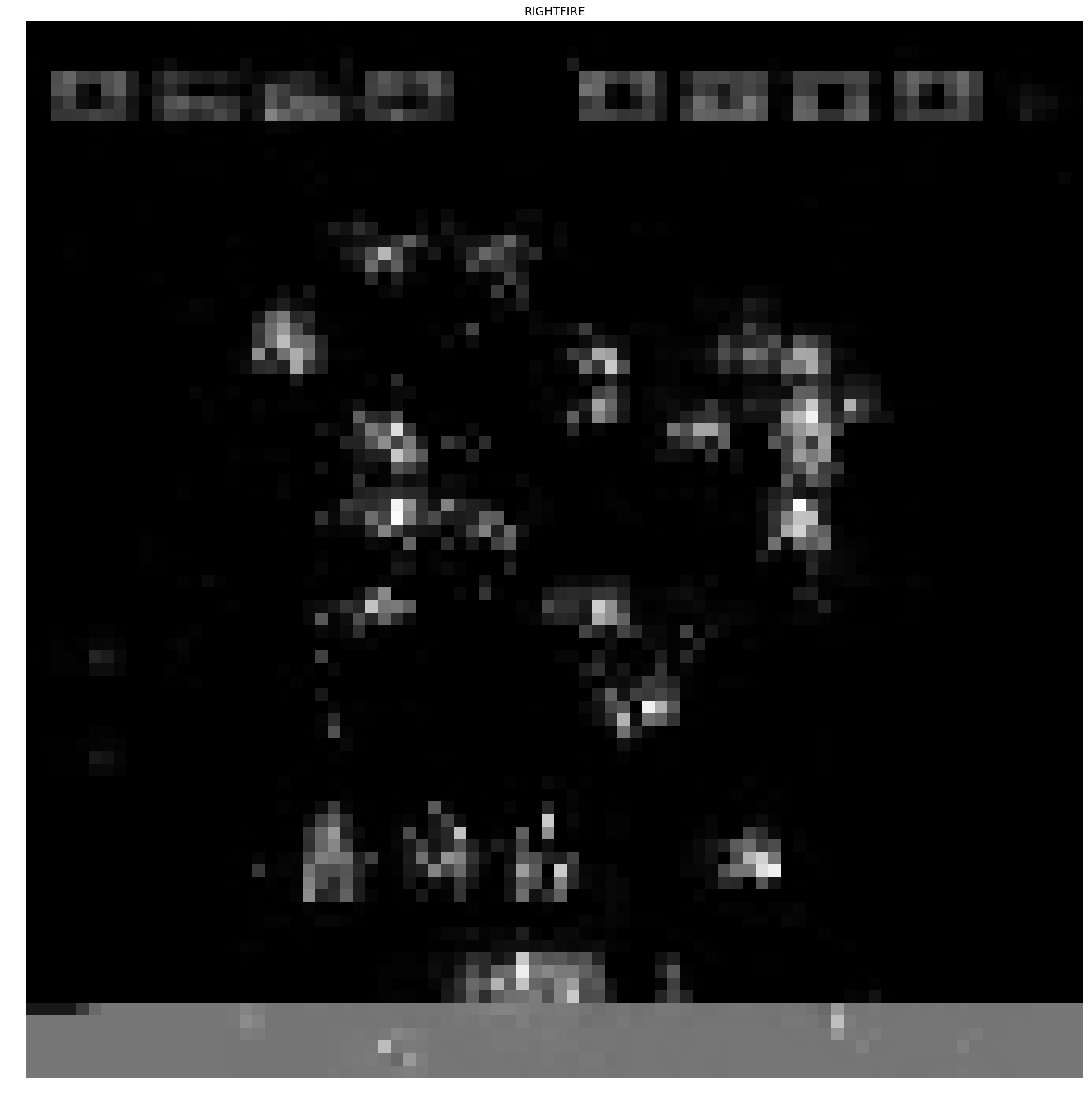}
    \caption{R-Fire}
\end{subfigure} \hspace*{-1em}
\begin{subfigure}[t]{0.10\textwidth}
  \centering
    \includegraphics[width=1.0\textwidth]{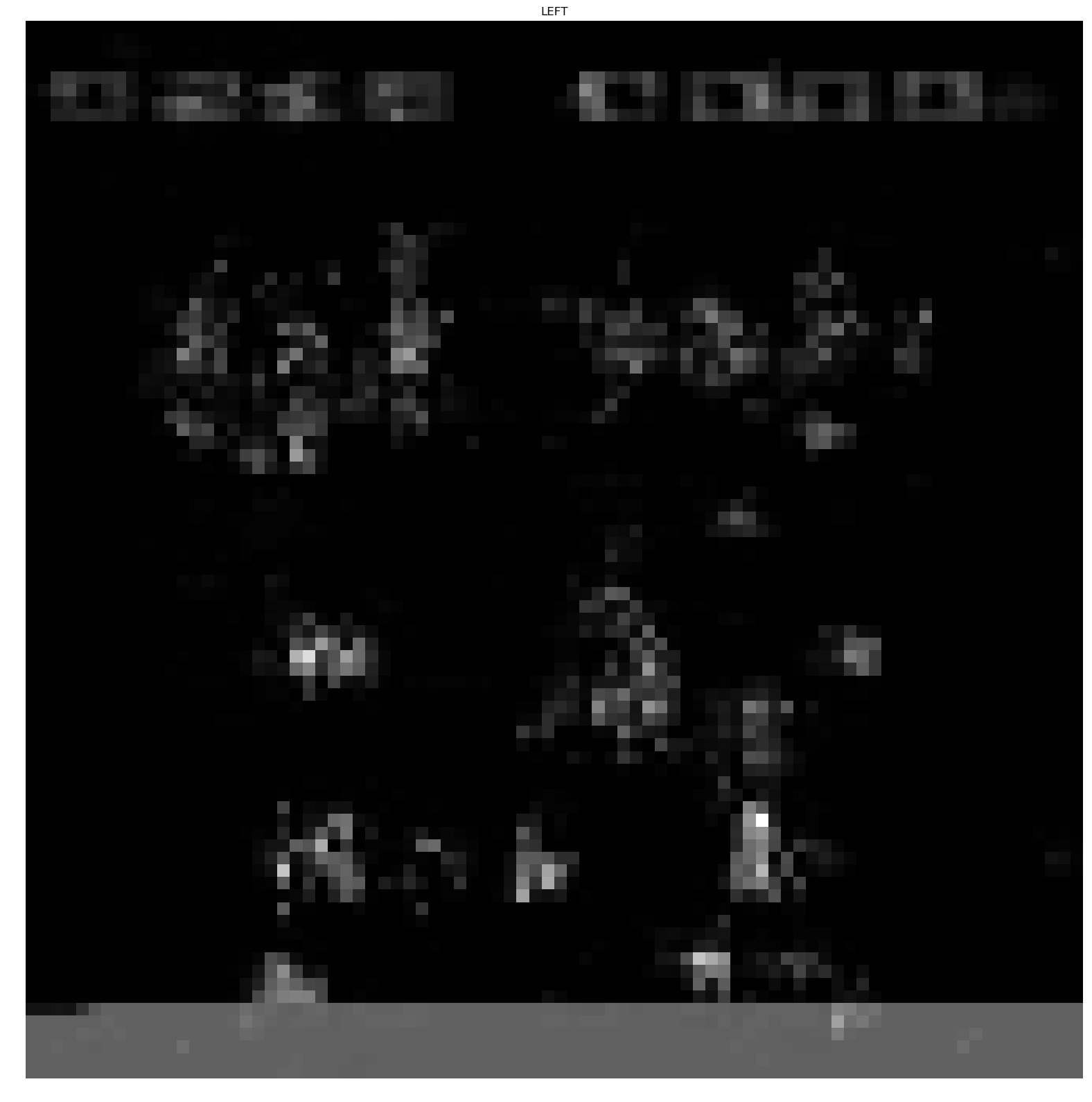}
    \caption{Left}
\end{subfigure} \hspace*{-1em}
\begin{subfigure}[t]{0.10\textwidth}
  \centering
    \includegraphics[width=1.0\textwidth]{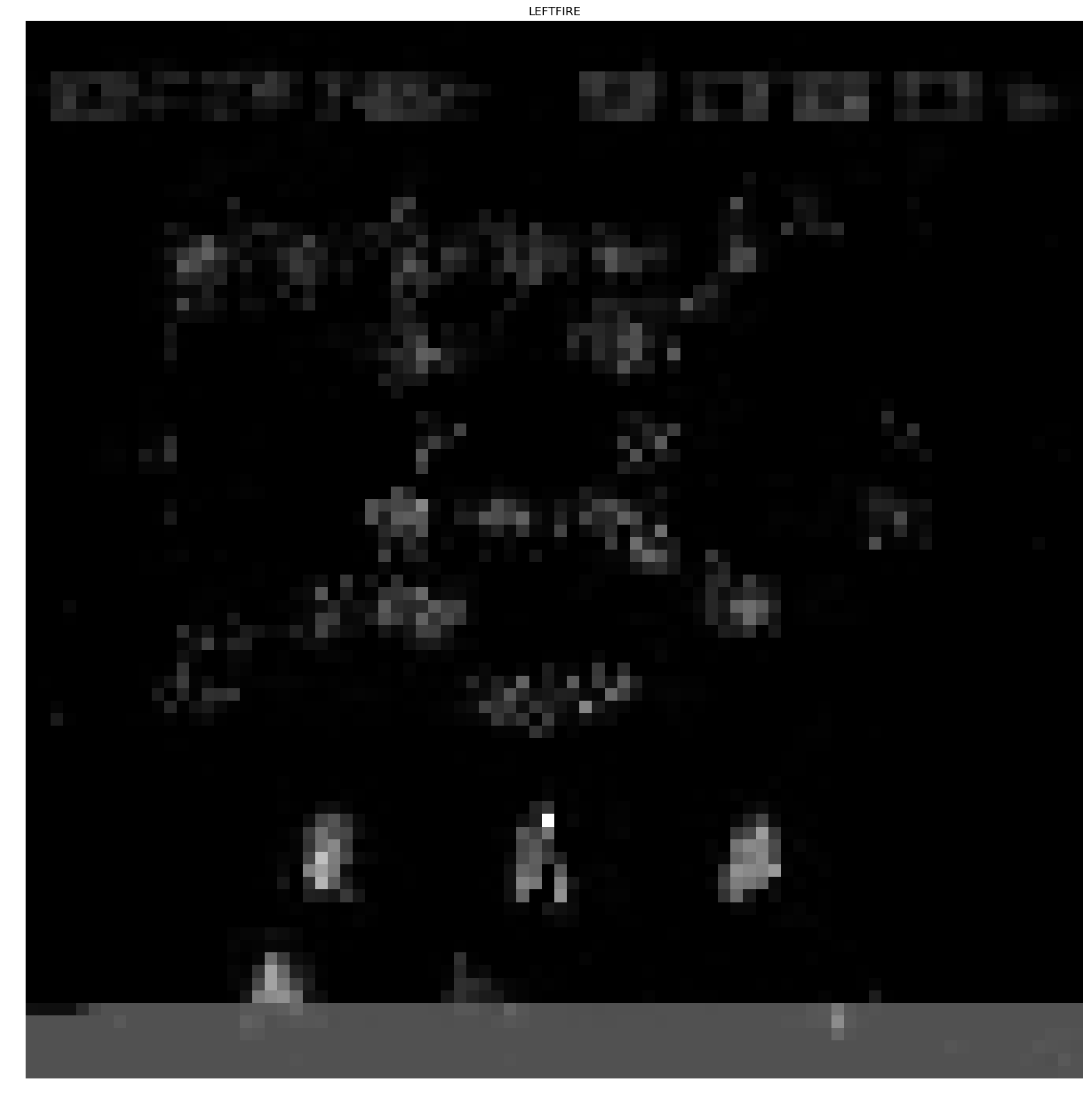}
    \caption{L-Fire}
\end{subfigure} \hspace*{-1em}
\begin{subfigure}[t]{0.10\textwidth}
  \centering
    \includegraphics[width=1.0\textwidth]{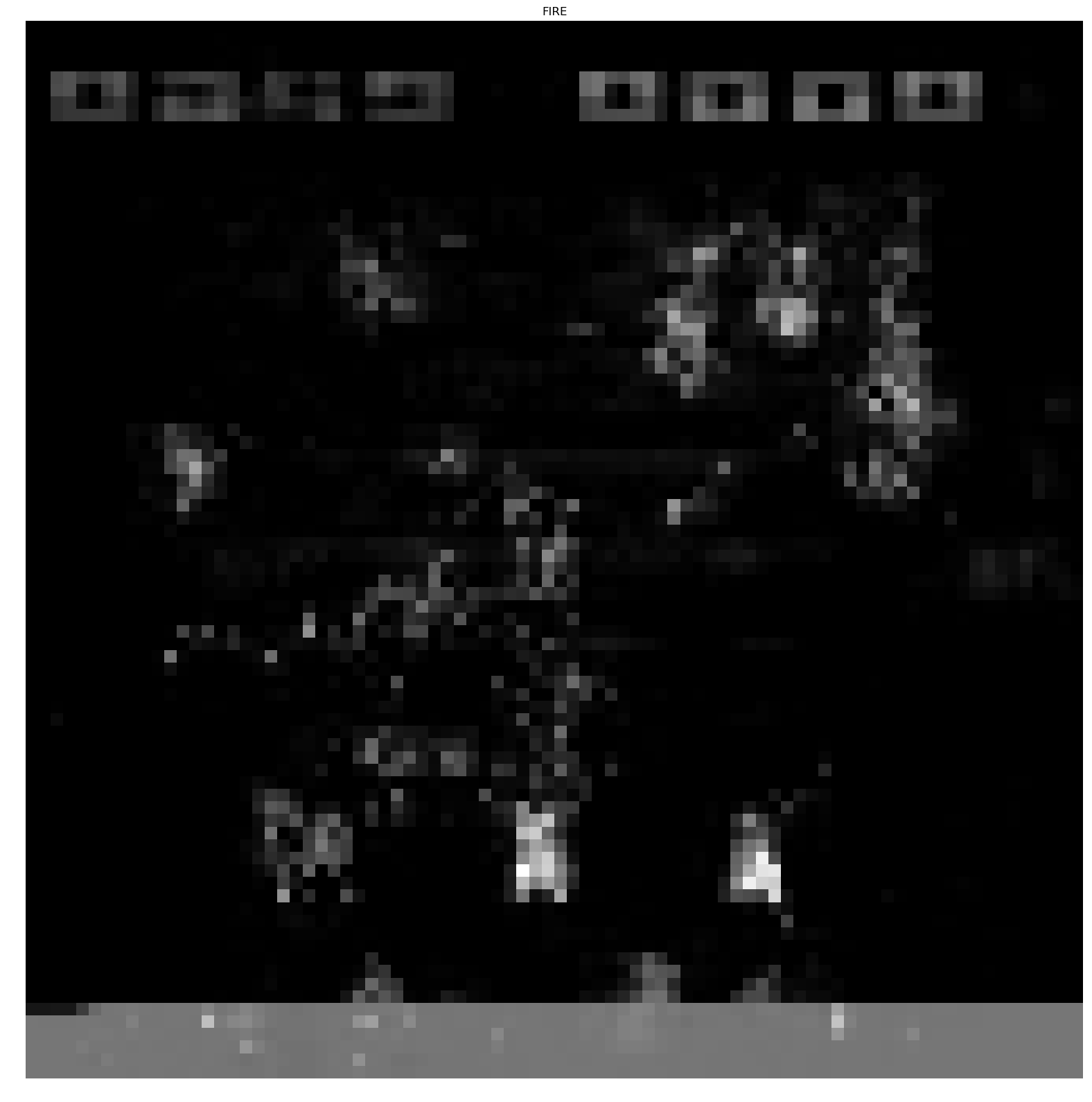}
    \caption{Fire}
\end{subfigure}
\caption{SpaceInvaders, Inverting keys for Q-value 25}
\label{fig:invert-key-25-spaceinvaders}
\end{figure}

\begin{figure*}[t!]
\centering
\begin{subfigure}[t]{0.2\textwidth}
  \centering
  \captionsetup{labelformat=empty}
  \captionsetup{position=top}
  \caption{Input State}
  \begin{tikzpicture}
    \node[anchor=south west,inner sep=0] (image) at (0,0) {\includegraphics[width=1.0\textwidth]{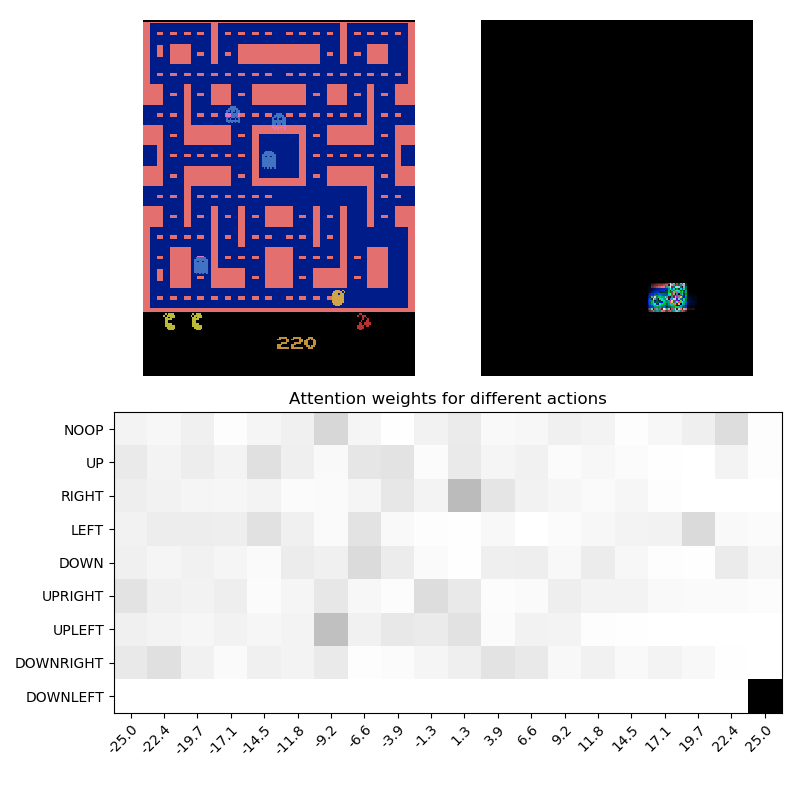}};
    \begin{scope}[x={(image.south east)},y={(image.north west)}]
        \draw[yellow,very thick,rounded corners] (0.38,0.58) rectangle (0.48,0.68);
    \end{scope}
  \end{tikzpicture}
\end{subfigure} \hspace*{-.5em}
\begin{subfigure}[t]{0.2\textwidth}
  \centering
  \captionsetup{labelformat=empty}
  \captionsetup{position=top}
  \caption{Input State}
  \begin{tikzpicture}
    \node[anchor=south west,inner sep=0] (image) at (0,0) {\includegraphics[width=1.0\textwidth]{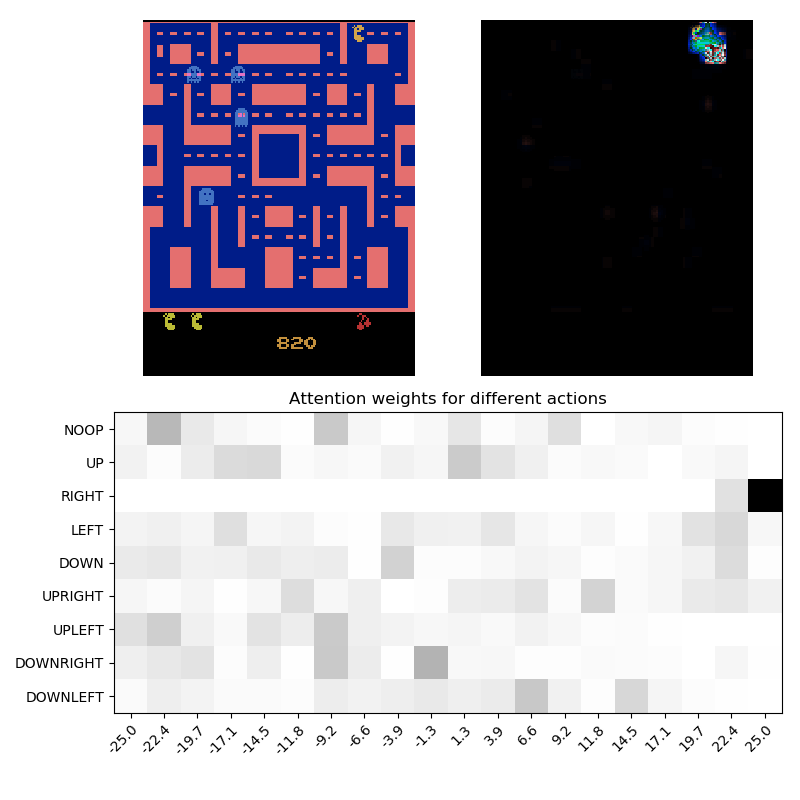}};
    \begin{scope}[x={(image.south east)},y={(image.north west)}]
        \draw[yellow,very thick,rounded corners] (0.4,0.9) rectangle (0.5,1.0);
    \end{scope}
  \end{tikzpicture}
\end{subfigure} \hspace*{-.5em}
\begin{subfigure}[t]{0.2\textwidth}
  \centering
  \captionsetup{labelformat=empty}
  \captionsetup{position=top}
  \caption{Input State}
  \begin{tikzpicture}
    \node[anchor=south west,inner sep=0] (image) at (0,0) {\includegraphics[width=1.0\textwidth]{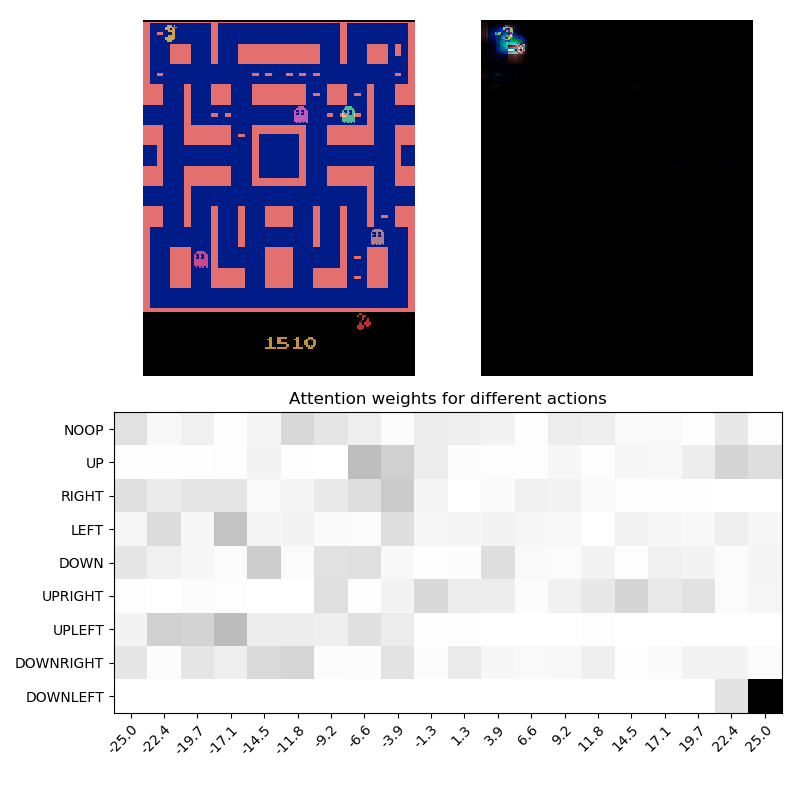}};
    \begin{scope}[x={(image.south east)},y={(image.north west)}]
        \draw[yellow,very thick,rounded corners] (0.18,0.9) rectangle (0.28,1.0);
    \end{scope}
  \end{tikzpicture}
\end{subfigure} \hspace*{-.5em}
\begin{subfigure}[t]{0.2\textwidth}
  \centering
  \captionsetup{labelformat=empty}
  \captionsetup{position=top}
  \caption{Input State}
  \begin{tikzpicture}
    \node[anchor=south west,inner sep=0] (image) at (0,0) {\includegraphics[width=1.0\textwidth]{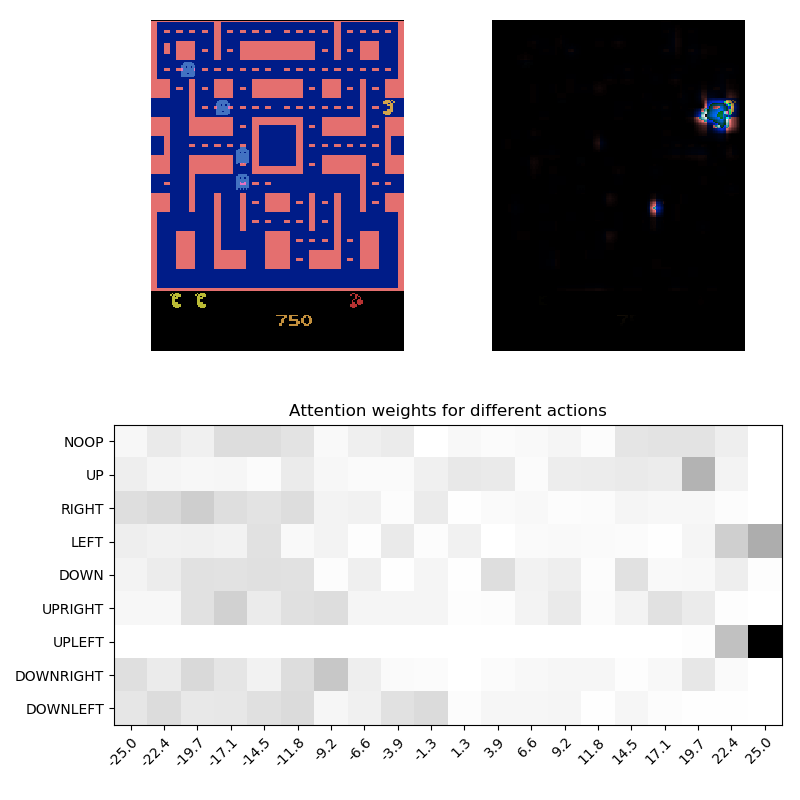}};
    \begin{scope}[x={(image.south east)},y={(image.north west)}]
        \draw[yellow,very thick,rounded corners] (0.44,0.8) rectangle (0.54,0.9);
    \end{scope}
  \end{tikzpicture}
\end{subfigure} \hspace*{-.5em}
\begin{subfigure}[t]{0.2\textwidth}
  \centering
  \captionsetup{labelformat=empty}
  \captionsetup{position=top}
  \caption{Input State}
  \begin{tikzpicture}
    \node[anchor=south west,inner sep=0] (image) at (0,0) {\includegraphics[width=1.0\textwidth]{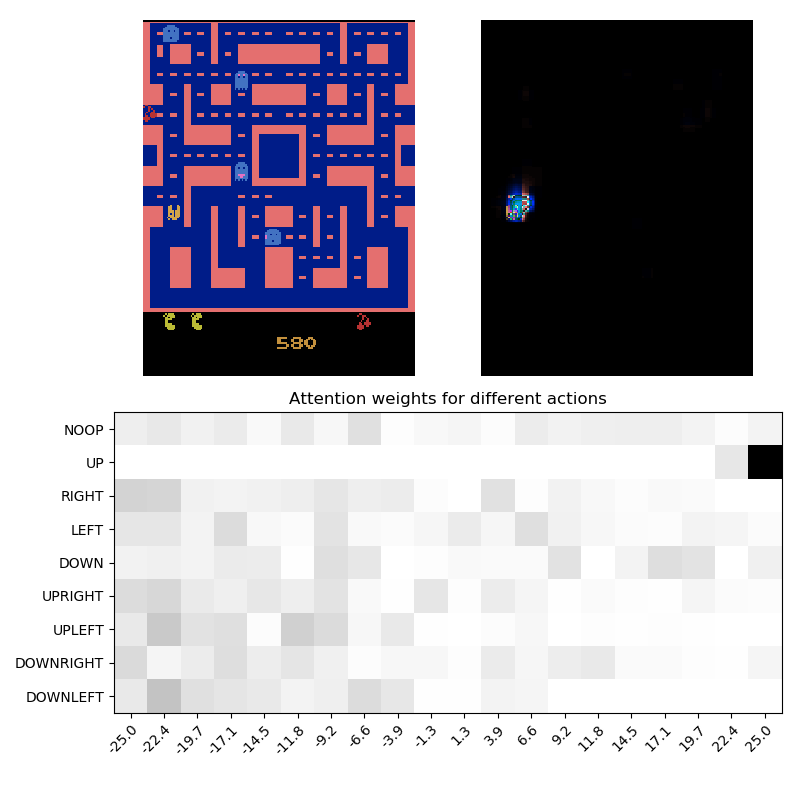}};
    \begin{scope}[x={(image.south east)},y={(image.north west)}]
        \draw[yellow,very thick,rounded corners] (0.18,0.68) rectangle (0.28,0.78);
    \end{scope}
  \end{tikzpicture}
\end{subfigure}
 \\
 \setcounter{subfigure}{0}
\begin{subfigure}[t]{0.15\textwidth}
  \centering
  \begin{tikzpicture}
    \node[anchor=south west,inner sep=0] (image) at (0,0) {\includegraphics[width=1.0\textwidth]{pacman-actions/DOWNLEFT-act-19.png}};
    \begin{scope}[x={(image.south east)},y={(image.north west)}]
        \draw[yellow,very thick,rounded corners] (0.65,0.18) rectangle (0.85,0.33);
    \end{scope}
  \end{tikzpicture}
  \caption{(Downleft, 25)}
  \label{fig:invert-agree-first}
\end{subfigure} \hspace*{2em}
\begin{subfigure}[t]{0.15\textwidth}
  \centering
  \begin{tikzpicture}
    \node[anchor=south west,inner sep=0] (image) at (0,0) {\includegraphics[width=1.0\textwidth]{pacman-actions/RIGHT-act-19.png}};
    \begin{scope}[x={(image.south east)},y={(image.north west)}]
        \draw[yellow,very thick,rounded corners] (0.7,0.85) rectangle (0.9,1.0);
    \end{scope}
  \end{tikzpicture}
  \caption{(Right, 25)}
\end{subfigure} \hspace*{2em}
\begin{subfigure}[t]{0.15\textwidth}
  \centering
  \begin{tikzpicture}
    \node[anchor=south west,inner sep=0] (image) at (0,0) {\includegraphics[width=1.0\textwidth]{pacman-actions/DOWNLEFT-act-19.png}};
    \begin{scope}[x={(image.south east)},y={(image.north west)}]
        \draw[yellow,very thick,rounded corners] (0.05,0.85) rectangle (0.25,1.0);
    \end{scope}
  \end{tikzpicture}
  \caption{(Downleft, 25)}
\end{subfigure} \hspace*{2em}
\begin{subfigure}[t]{0.15\textwidth}
  \centering
  \begin{tikzpicture}
    \node[anchor=south west,inner sep=0] (image) at (0,0) {\includegraphics[width=1.0\textwidth]{pacman-actions/UPLEFT-act-19.png}};
    \begin{scope}[x={(image.south east)},y={(image.north west)}]
        \draw[yellow,very thick,rounded corners] (0.8,0.65) rectangle (1.0,0.8);
    \end{scope}
  \end{tikzpicture}
  \caption{(Upleft, 25)}
\end{subfigure} \hspace*{2em}
\begin{subfigure}[t]{0.15\textwidth}
  \centering
  \begin{tikzpicture}
    \node[anchor=south west,inner sep=0] (image) at (0,0) {\includegraphics[width=1.0\textwidth]{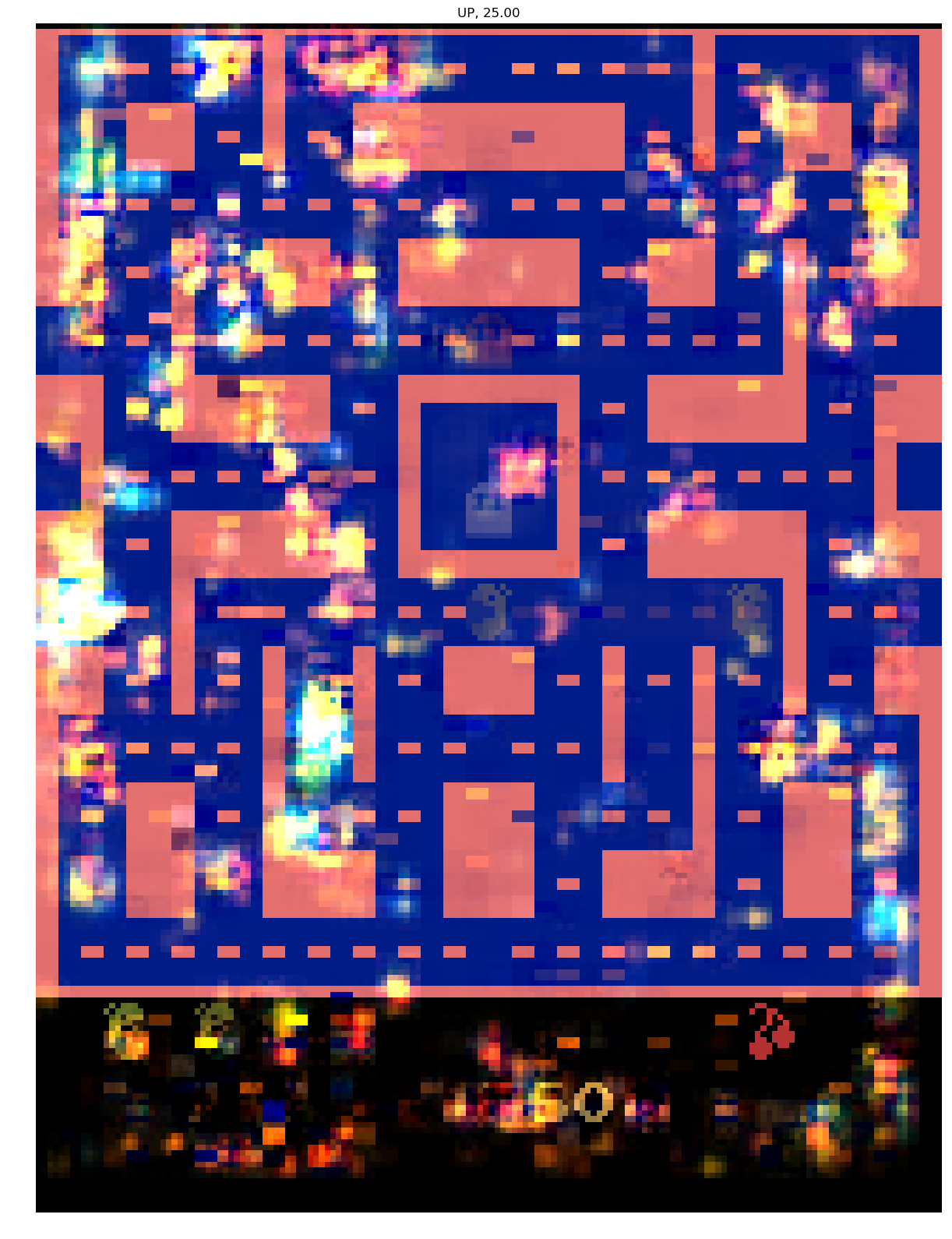}};
    \begin{scope}[x={(image.south east)},y={(image.north west)}]
        \draw[yellow,very thick,rounded corners] (0.05,0.41) rectangle (0.25,0.56);
    \end{scope}
  \end{tikzpicture}
  \caption{(Up, 25)}
\end{subfigure}
\caption{MsPacman, examples where agent's decision agrees with the reconstructed image}
\label{fig:invert-agree}
\end{figure*}

\subsection{Inversion of keys}
Although keys act like cluster centers for action-return pairs, it is difficult to interpret them in the latent space. By inverting keys, we attempt to find important aspects of input space (images) that influence the agent to choose particular action-return pair ($Deconv(h^a_i)$). These are `global explanations' because inverting keys is independent of the input. For example, in MsPacman, reconstructing keys for different actions (fixing return of 25) indicates yellow blobs at many different places for each action (Figure~\ref{fig:invert-key-25-pacman}). We hypothesize that these correspond to the Pacman object itself and that the model memorizes its different positions to make its decision i.e. the yellow blobs in Figure~\ref{fig:invert-key-25-pacman-right} correspond to different locations of Pacman and for any input state where Pacman is in one of those positions, the agent selects action right expecting a return of 25. Figure~\ref{fig:invert-agree} shows such examples where the agent's action-return selection agrees with reconstructed key (red boxes indicate Pacman's location). Similarly, in SpaceInvaders, the agent seems to be looking at specific combinations of shooter and alien ship positions that were observed during training (Figure~\ref{fig:invert-key-25-spaceinvaders}).

\begin{figure}[t]
    \raggedright
\begin{subfigure}[t]{0.11\textwidth}
  \raggedright
    \includegraphics[width=1.0\textwidth]{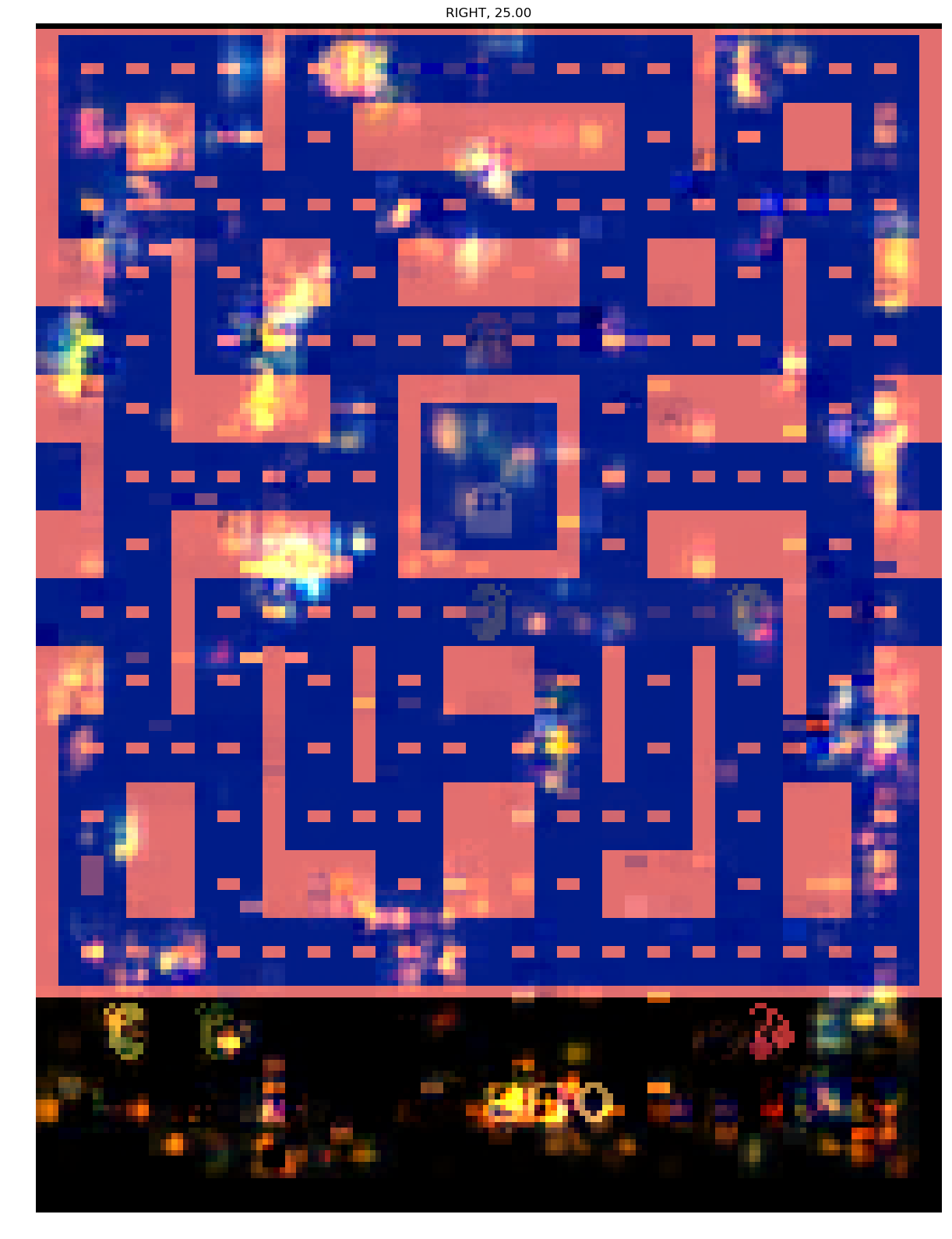}
\end{subfigure} \hspace*{-.1em}
\begin{subfigure}[t]{0.11\textwidth}
  \raggedright
    \includegraphics[width=1.0\textwidth]{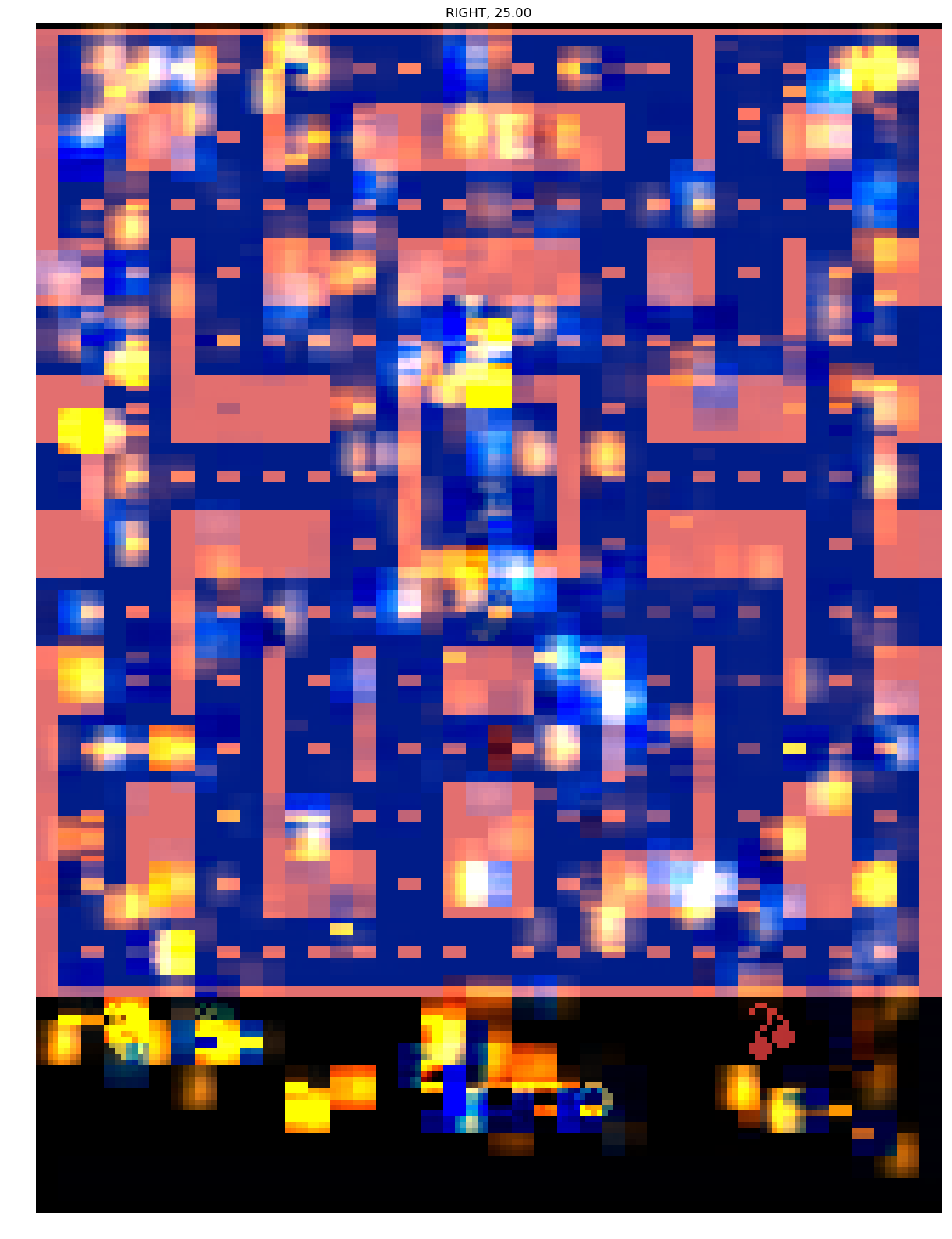}
\end{subfigure} \hspace*{-.1em}
\begin{subfigure}[t]{0.11\textwidth}
  \raggedright
    \includegraphics[width=1.0\textwidth]{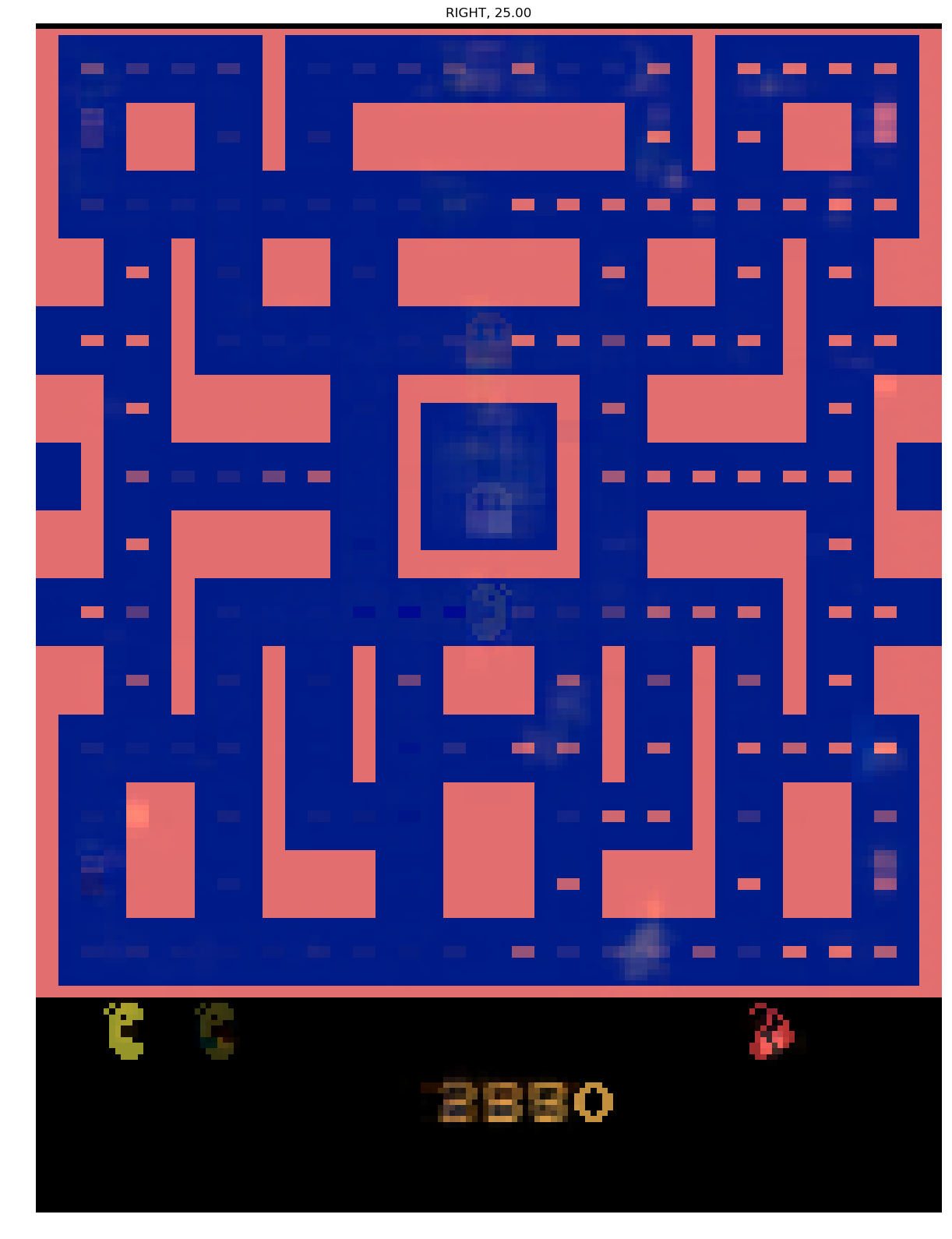}
\end{subfigure} \hspace*{-.1em}
\begin{subfigure}[t]{0.11\textwidth}
  \raggedright
    \includegraphics[width=1.0\textwidth]{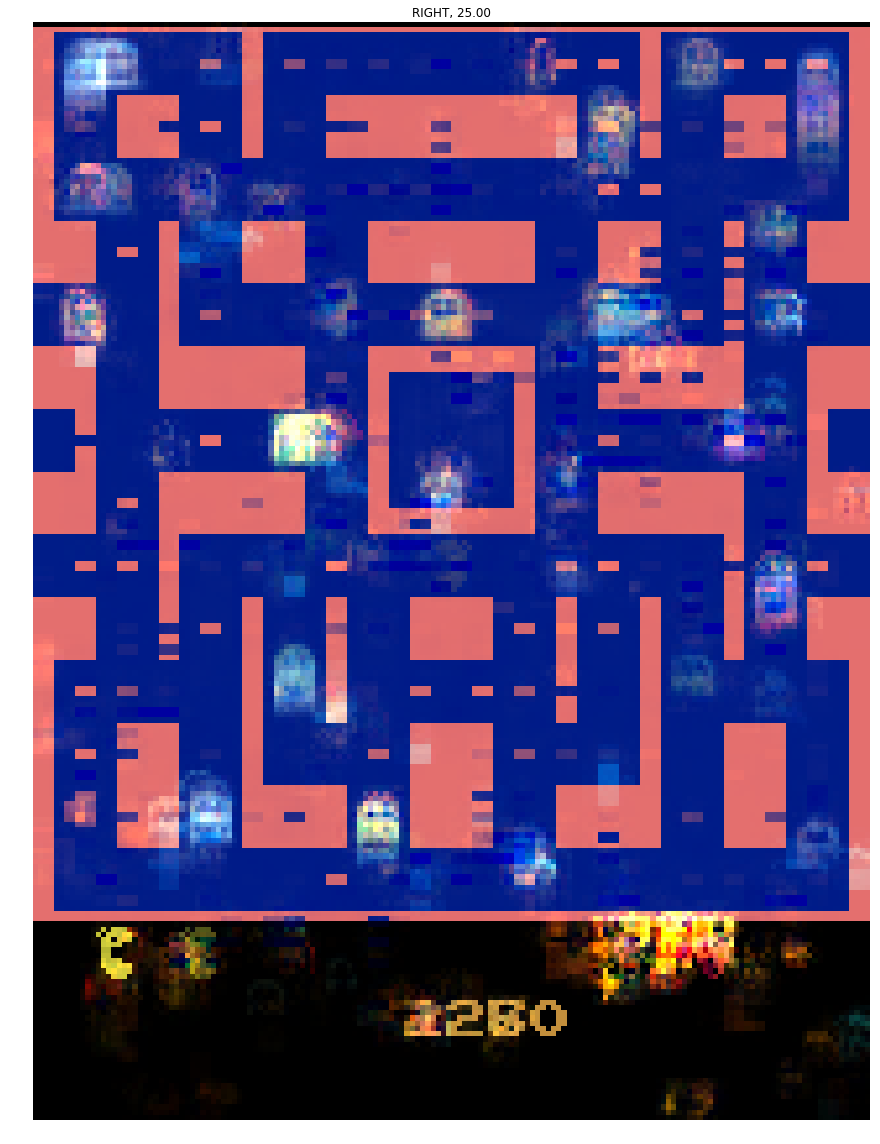}
\end{subfigure} 
\caption{\raggedright Reconstruction: AE, VAE $\beta=0$, VAE $\beta=0.01$, Dense VAE}
\label{fig:invert-methods}
\end{figure}

The keys have never been observed by the deconvolution network during training and so the reconstructions depend upon its generalizability. Interestingly, reconstructions for action-return pairs that are seen more often tend to be less noisy with less artifacts. This can be observed in Figure~\ref{fig:invert-key-25-pacman} for Q-value 25 where actions Right, Downleft and Upleft nearly 65\% of all actions taken by the agent. We also look at the effect of different reconstruction techniques keeping the action-return pair fixed (Figure~\ref{fig:invert-methods}). Variational autoencoder with $\beta$ set to 0 yields sharper looking images but increasing $\beta$ which is supposed to bring out disentanglement in the embeddings yields reconstructions with almost no objects. Dense VAE with $\beta=0$ is a slightly deeper network similar to \cite{oh2015action} and seems to reconstruct slightly clearer shapes of ghosts and pacman.
\begin{table}[t!]
 \centering
 {\small
  \setlength\tabcolsep{1.9pt}
  \begin{tabular}{lcccc}
    \hline
    \textbf{Agreement} & AE & \specialcell[t]{VAE \\($\beta=0$)} & \specialcell[t]{VAE \\($\beta=0.01$)} & \specialcell[t]{Dense VAE\\ ($\beta=0$)} \\ \hline
    \specialcell[t]{MsPacman (Color) \rule{0pt}{2.6ex}} & \textbf{30.76} & 29.78 & 16.8 & 23.73 \\
    \specialcell[t]{MsPacman (Gray,\\ Rescaled)} & 19.87 & 18.14 & 10.97 & 14.56\\ \hline
  \end{tabular}
  }%
  \caption{Evaluating visualizations: Agreement scores}
  \label{table:agreeability}
\end{table}
\begin{figure*}[t!]
\centering
\begin{subfigure}[t]{\textwidth}
\centering
\begin{adjustbox}{minipage=\linewidth,scale=0.9}
\centering
\begin{subfigure}[t]{0.14\textwidth}
  \centering
    \includegraphics[width=1.0\textwidth]{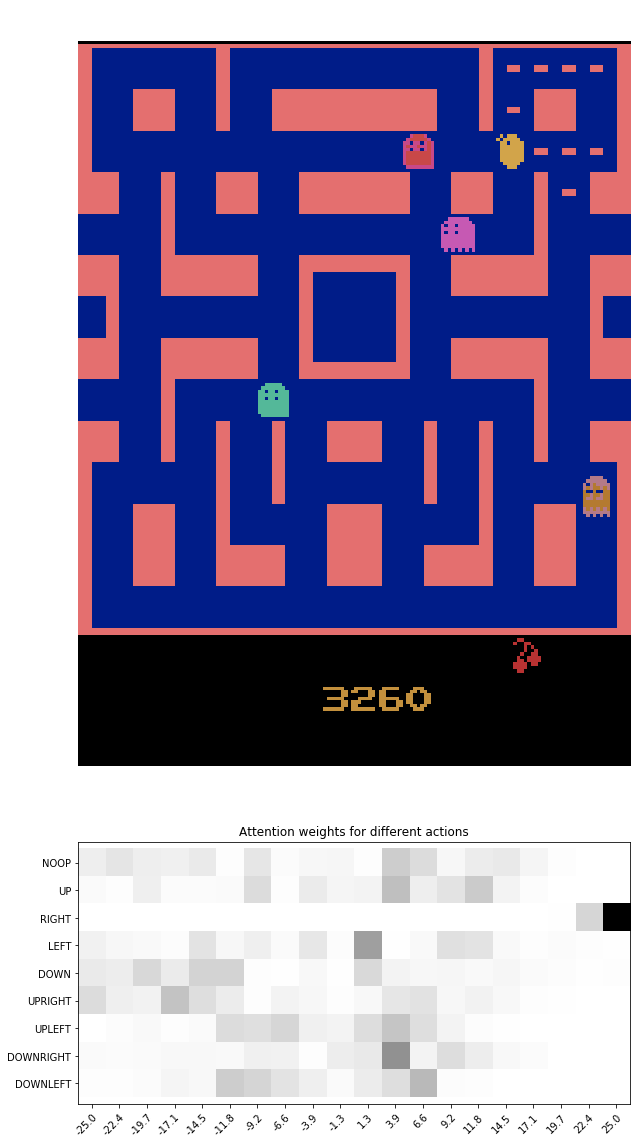}
\end{subfigure} \hspace*{-.5em}
\begin{subfigure}[t]{0.14\textwidth}
  \centering
    \includegraphics[width=1.0\textwidth]{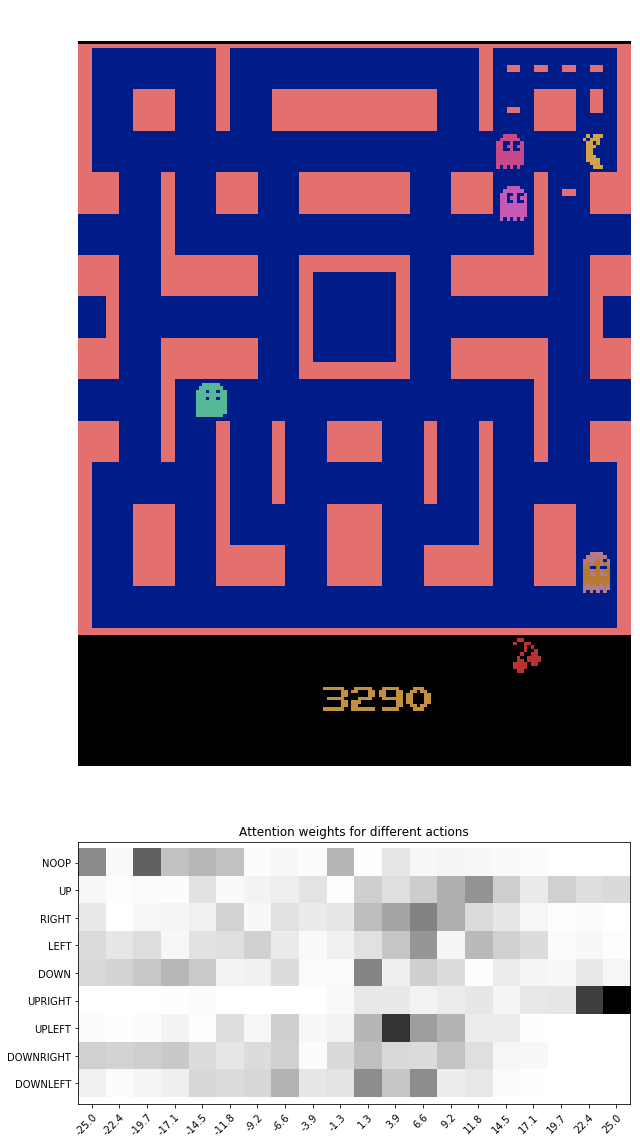}
\end{subfigure} \hspace*{-.5em}
\begin{subfigure}[t]{0.14\textwidth}
  \centering
    \includegraphics[width=1.0\textwidth]{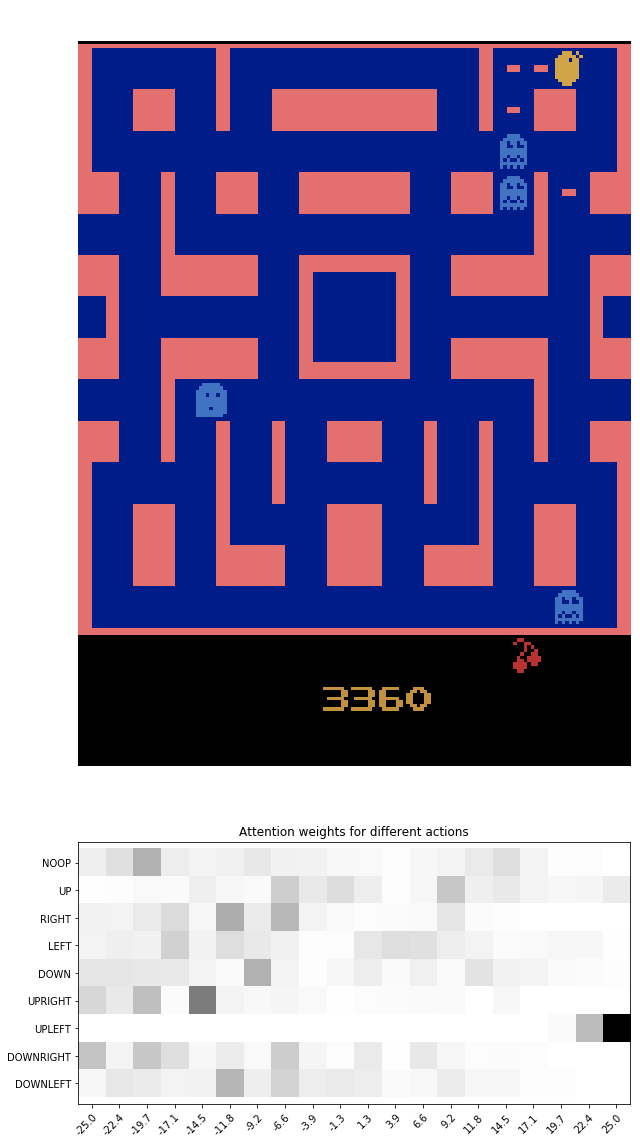}
\end{subfigure} \hspace*{-.5em}
\begin{subfigure}[t]{0.14\textwidth}
  \centering
    \includegraphics[width=1.0\textwidth]{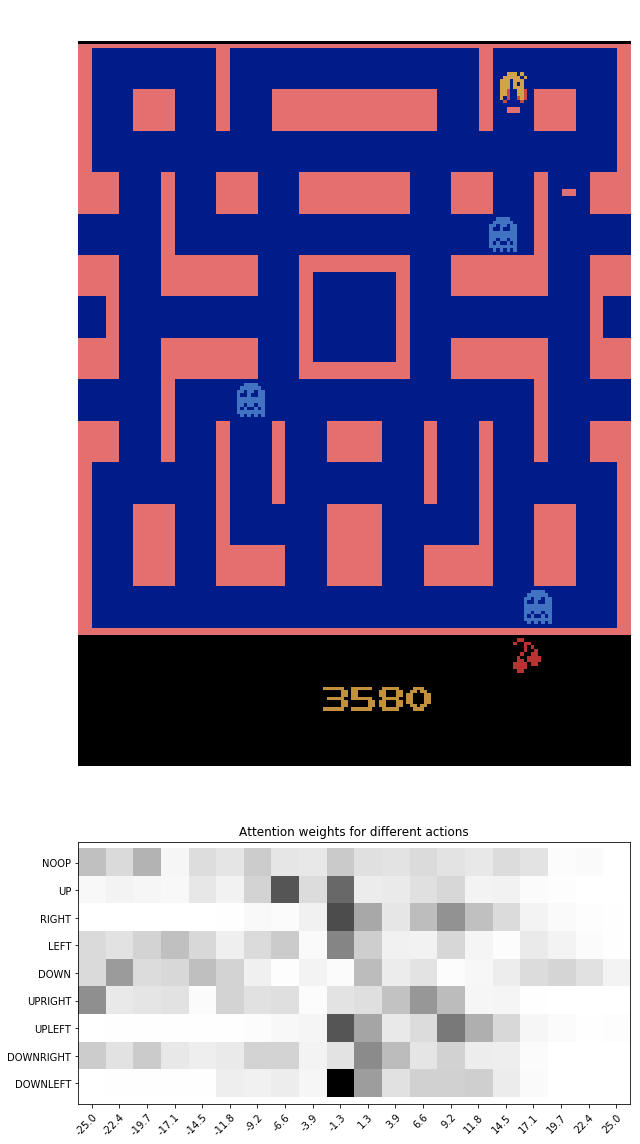}
\end{subfigure} \hspace*{-.5em}
\begin{subfigure}[t]{0.14\textwidth}
  \centering
    \includegraphics[width=1.0\textwidth]{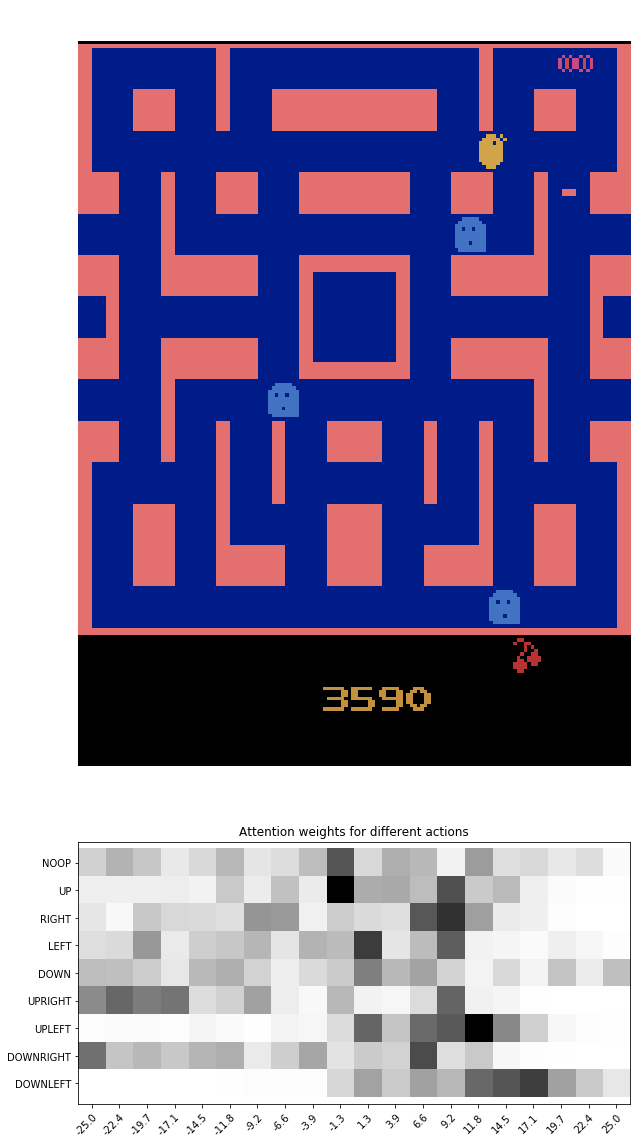}
\end{subfigure} \hspace*{-.5em}
\begin{subfigure}[t]{0.14\textwidth}
  \centering
    \includegraphics[width=1.0\textwidth]{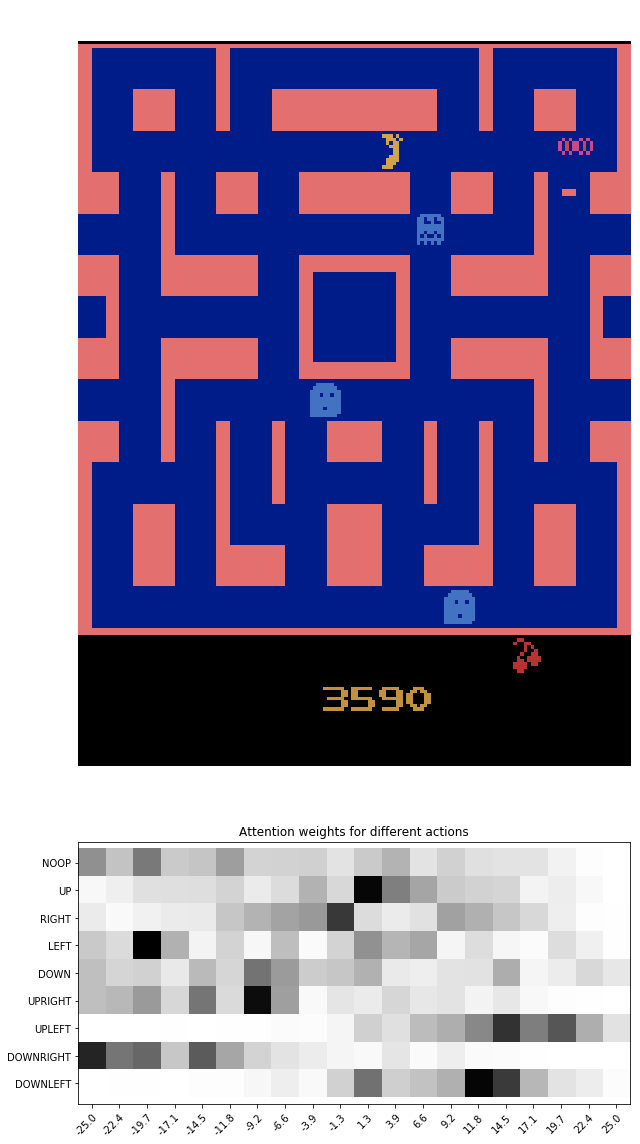}
\end{subfigure} \hspace*{-.5em}
\begin{subfigure}[t]{0.14\textwidth}
  \centering
    \begin{tikzpicture}
    \node[anchor=south west,inner sep=0] (image) at (0,0) {\includegraphics[width=1.0\textwidth]{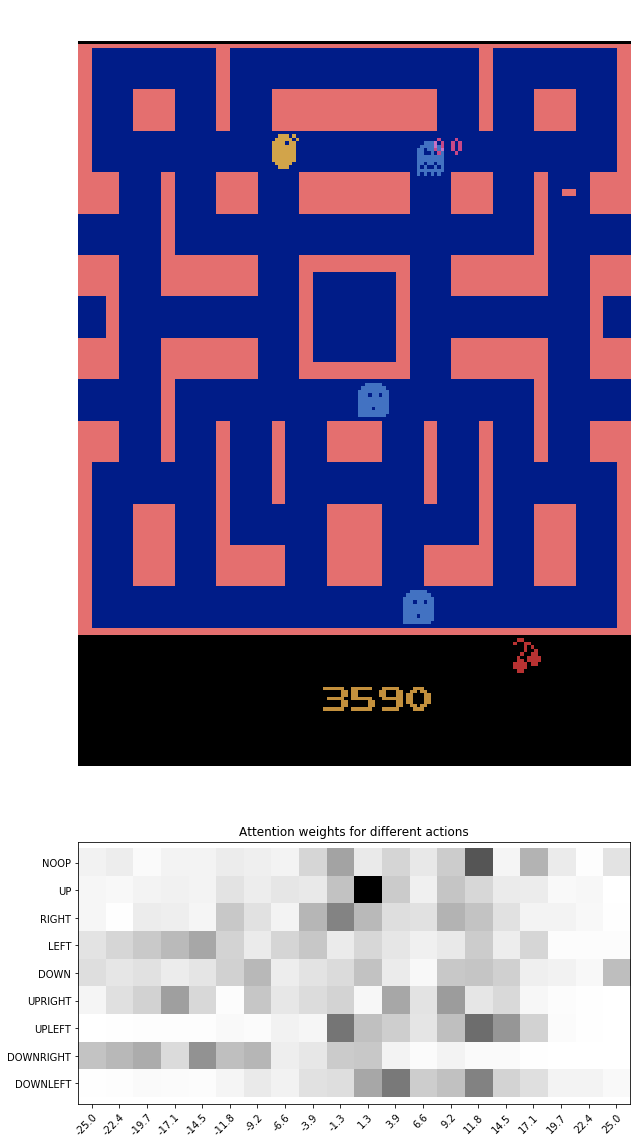}};
    \begin{scope}[x={(image.south east)},y={(image.north west)}]
        \draw[yellow,very thick,rounded corners] (0.82,0.77) rectangle (0.97,0.87);
    \end{scope}
  \end{tikzpicture}
\end{subfigure}
\end{adjustbox}
\caption{Adversarial example}
\label{fig:pacman-test-1}
\end{subfigure}\\
\begin{subfigure}[t]{\textwidth}
\centering
\begin{adjustbox}{minipage=\linewidth,scale=0.9}
\centering
\begin{subfigure}[t]{0.14\textwidth}
  \centering
    \includegraphics[width=1.0\textwidth]{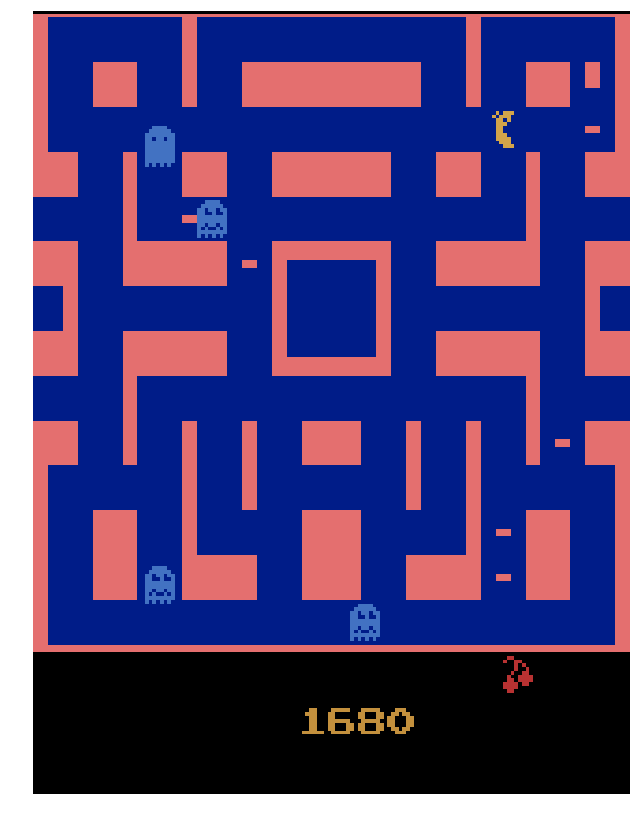}
\end{subfigure} \hspace*{-.5em}
\begin{subfigure}[t]{0.14\textwidth}
  \centering
    \includegraphics[width=1.0\textwidth]{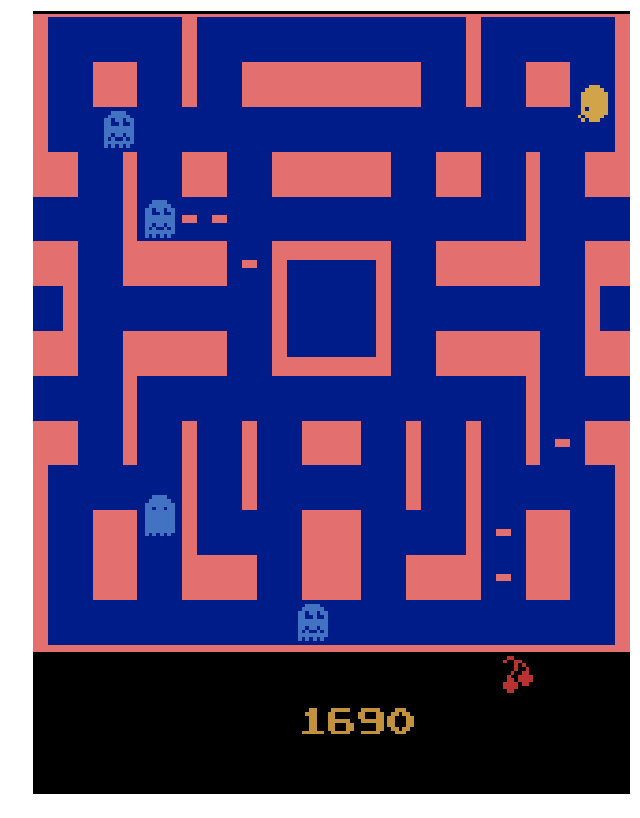}
\end{subfigure} \hspace*{-.5em}
\begin{subfigure}[t]{0.14\textwidth}
  \centering
    \includegraphics[width=1.0\textwidth]{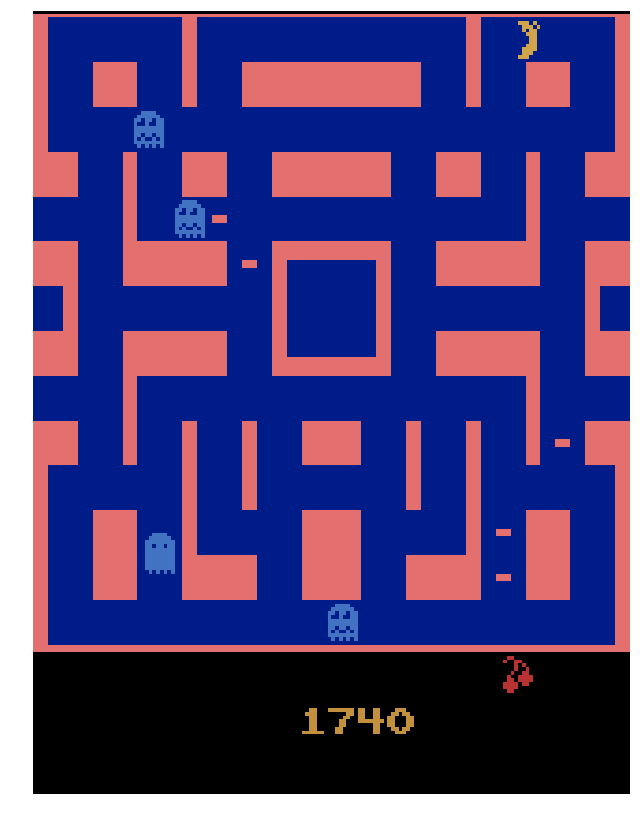}
\end{subfigure} \hspace*{-.5em}
\begin{subfigure}[t]{0.14\textwidth}
  \centering
    \includegraphics[width=1.0\textwidth]{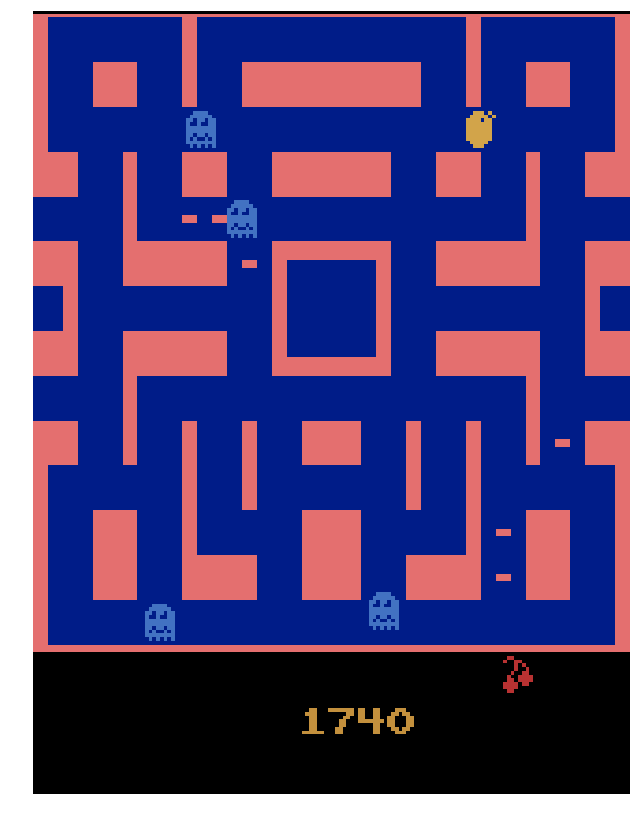}
\end{subfigure} \hspace*{-.5em}
\begin{subfigure}[t]{0.14\textwidth}
  \centering
    \includegraphics[width=1.0\textwidth]{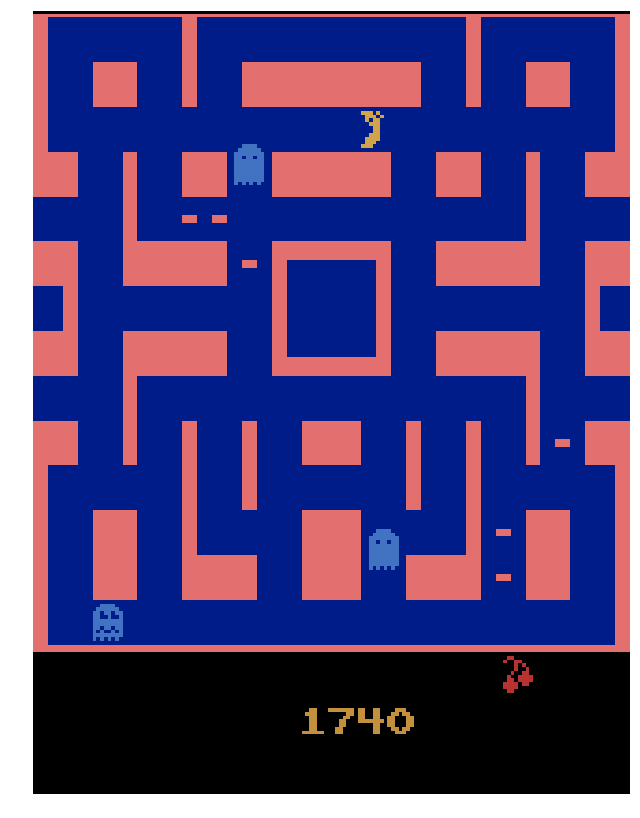}
\end{subfigure} \hspace*{-.5em}
\begin{subfigure}[t]{0.14\textwidth}
  \centering
    \includegraphics[width=1.0\textwidth]{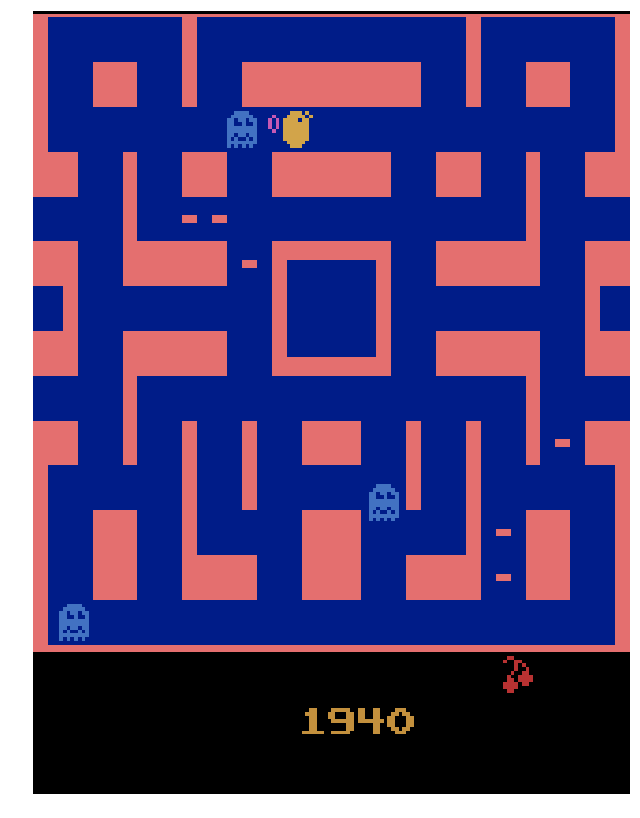}
\end{subfigure}
\end{adjustbox}
\caption{Trajectory during training}
\label{fig:pacman-train-1}
\end{subfigure}\\
\begin{subfigure}[t]{\textwidth}
\centering
\begin{adjustbox}{minipage=\linewidth,scale=0.9}
\centering
\begin{subfigure}[t]{0.14\textwidth}
  \centering
    \includegraphics[width=1.0\textwidth]{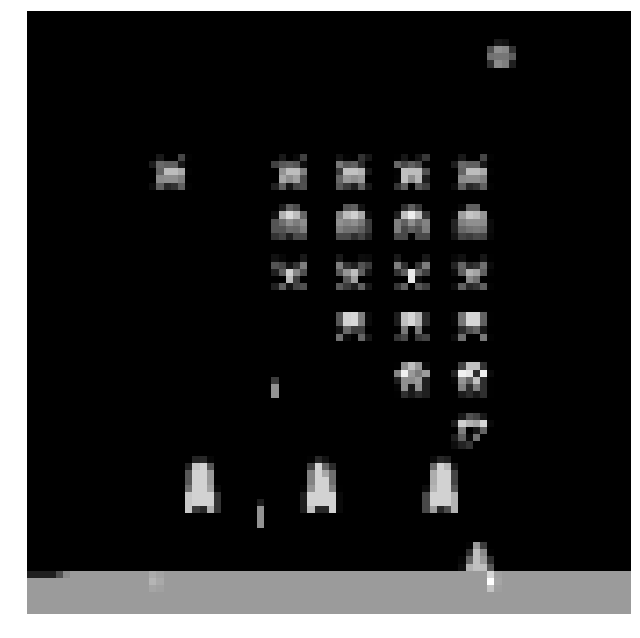}
\end{subfigure} \hspace*{-.5em}
\begin{subfigure}[t]{0.14\textwidth}
  \centering
    \includegraphics[width=1.0\textwidth]{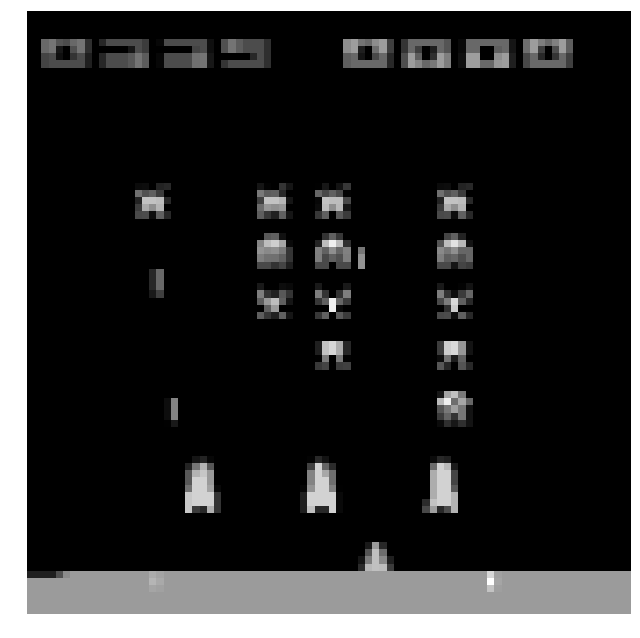}
\end{subfigure} \hspace*{-.5em}
\begin{subfigure}[t]{0.14\textwidth}
  \centering
    \includegraphics[width=1.0\textwidth]{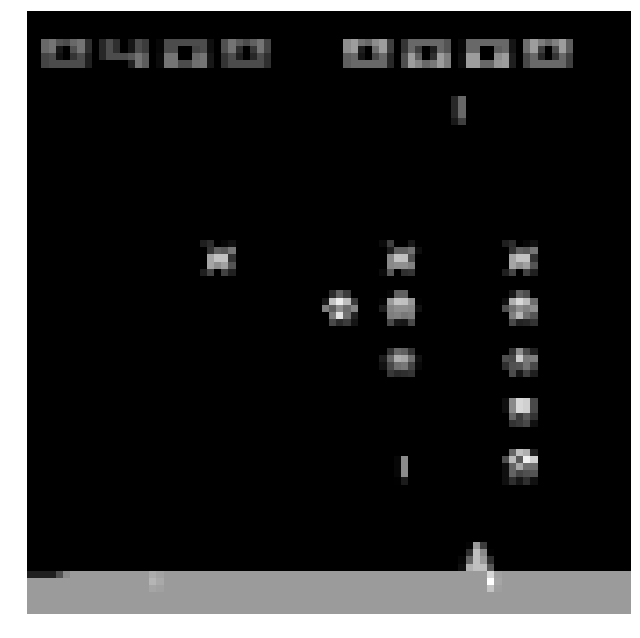}
\end{subfigure} \hspace*{-.5em}
\begin{subfigure}[t]{0.14\textwidth}
  \centering
    \includegraphics[width=1.0\textwidth]{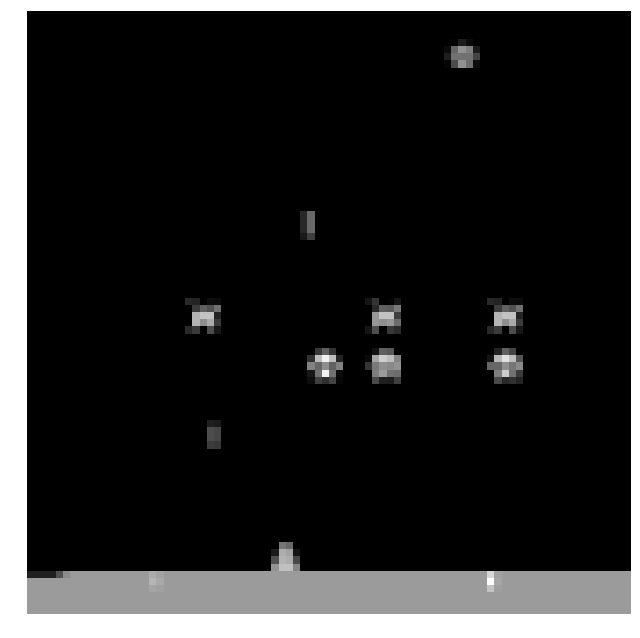}
\end{subfigure} \hspace*{-.5em}
\begin{subfigure}[t]{0.14\textwidth}
  \centering
    \includegraphics[width=1.0\textwidth]{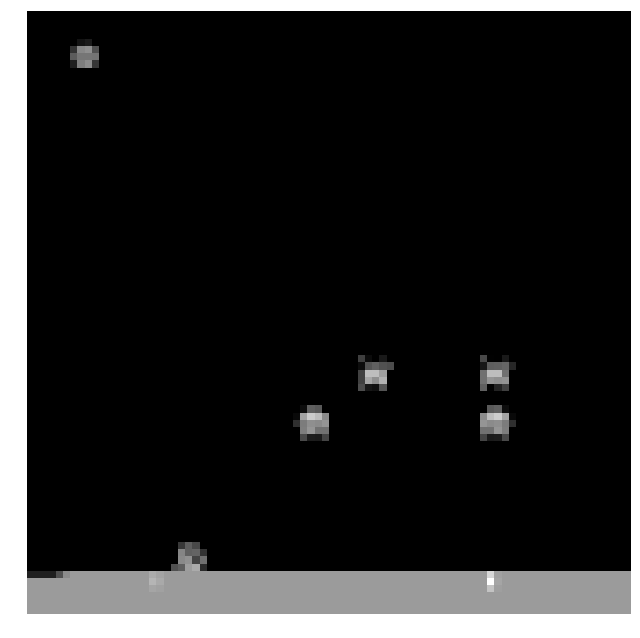}
\end{subfigure} \hspace*{-.5em}
\begin{subfigure}[t]{0.14\textwidth}
  \centering
    \includegraphics[width=1.0\textwidth]{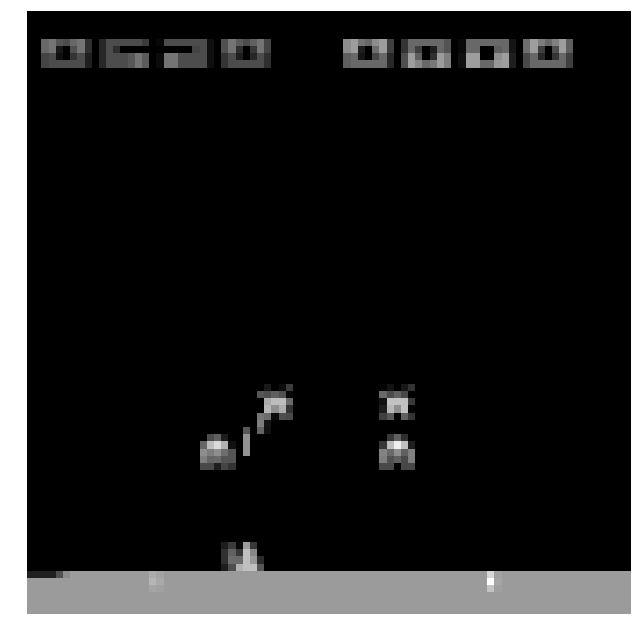}
\end{subfigure} \hspace*{-.5em}
\begin{subfigure}[t]{0.14\textwidth}
  \centering
    \includegraphics[width=1.0\textwidth]{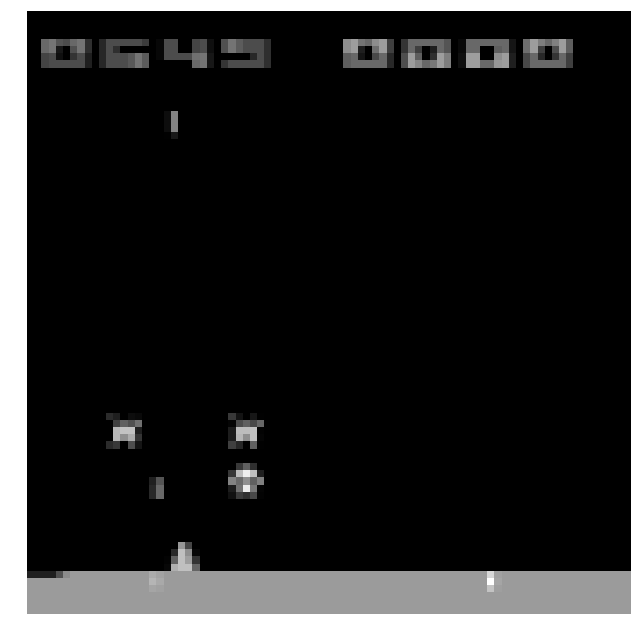}
\end{subfigure}
\end{adjustbox}
\caption{Adversarial example}
\label{fig:spaceinvaders-test-1}
\end{subfigure}
\begin{subfigure}[t]{\textwidth}
\centering
\begin{adjustbox}{minipage=\linewidth,scale=0.9}
\centering
\begin{subfigure}[t]{0.14\textwidth}
  \centering
    \includegraphics[width=1.0\textwidth]{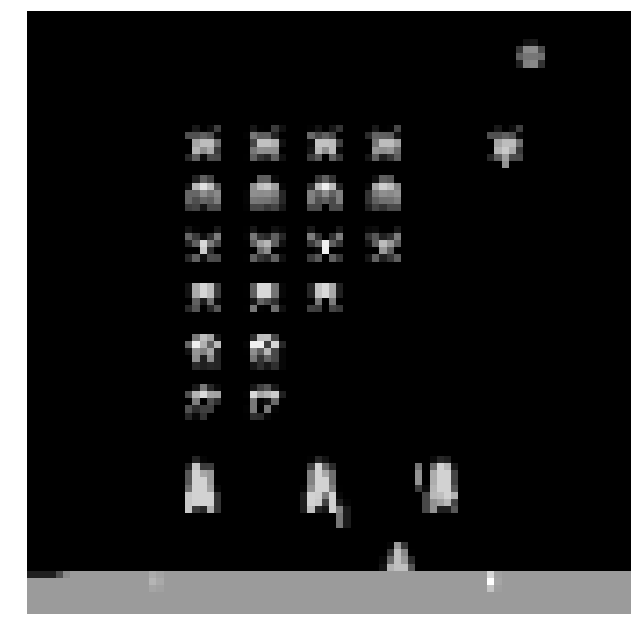}
\end{subfigure} \hspace*{-.5em}
\begin{subfigure}[t]{0.14\textwidth}
  \centering
    \includegraphics[width=1.0\textwidth]{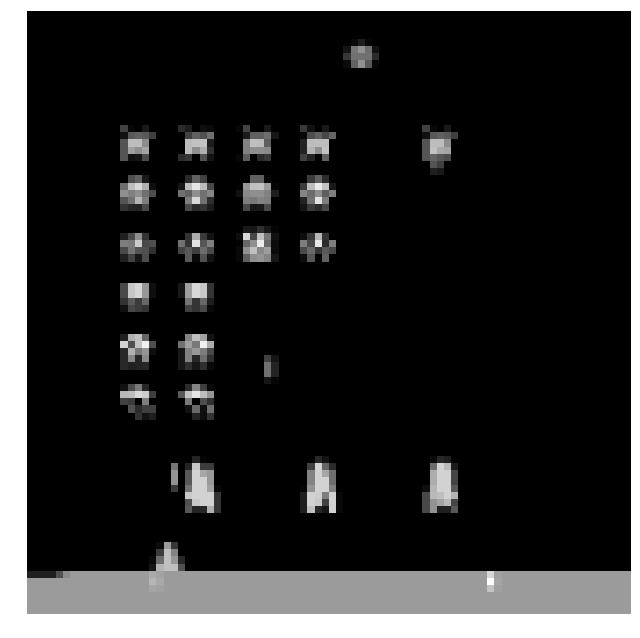}
\end{subfigure} \hspace*{-.5em}
\begin{subfigure}[t]{0.14\textwidth}
  \centering
    \includegraphics[width=1.0\textwidth]{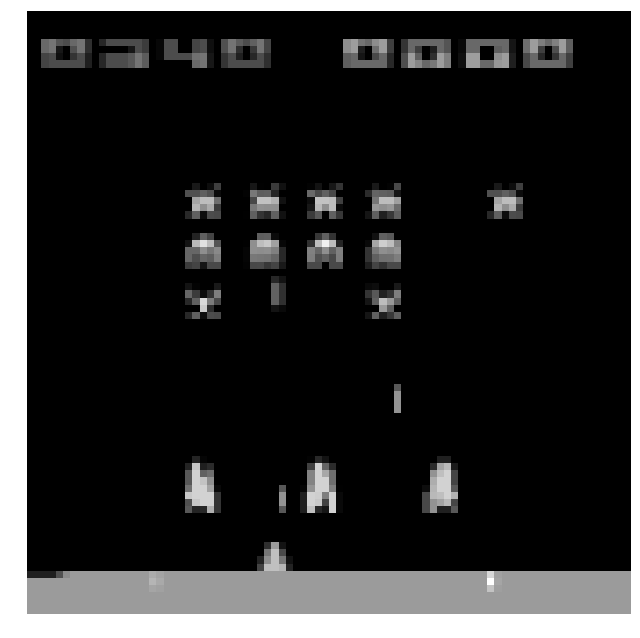}
\end{subfigure} \hspace*{-.5em}
\begin{subfigure}[t]{0.14\textwidth}
  \centering
    \includegraphics[width=1.0\textwidth]{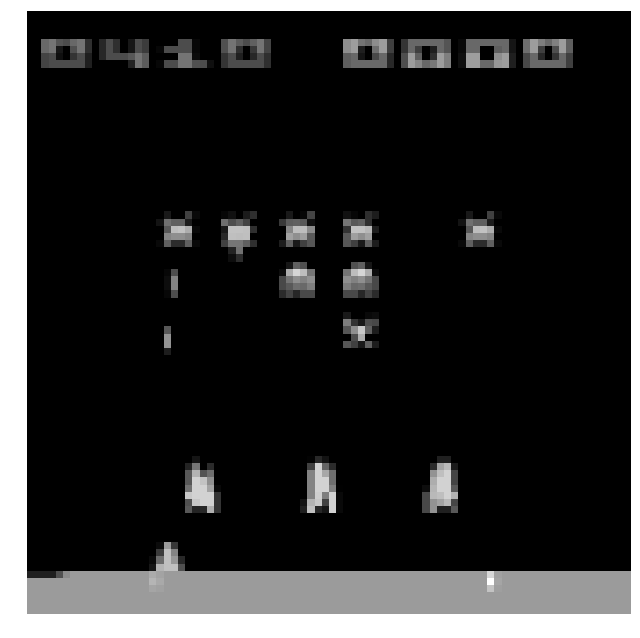}
\end{subfigure} \hspace*{-.5em}
\begin{subfigure}[t]{0.14\textwidth}
  \centering
    \includegraphics[width=1.0\textwidth]{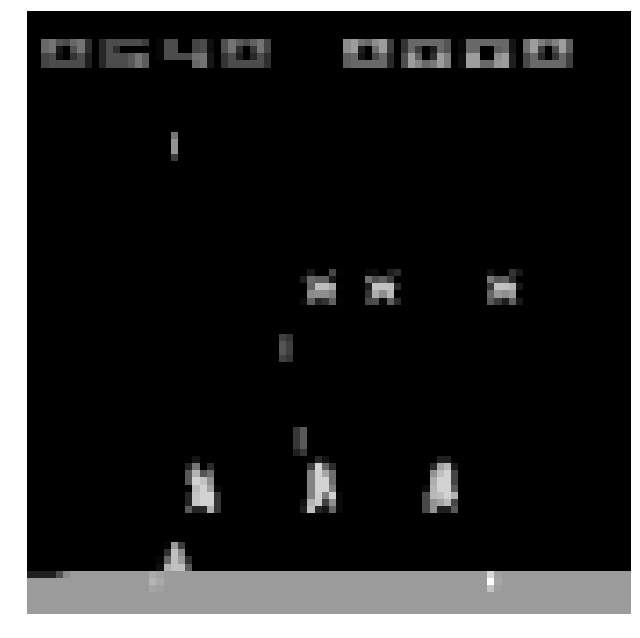}
\end{subfigure} \hspace*{-.5em}
\begin{subfigure}[t]{0.14\textwidth}
  \centering
    \includegraphics[width=1.0\textwidth]{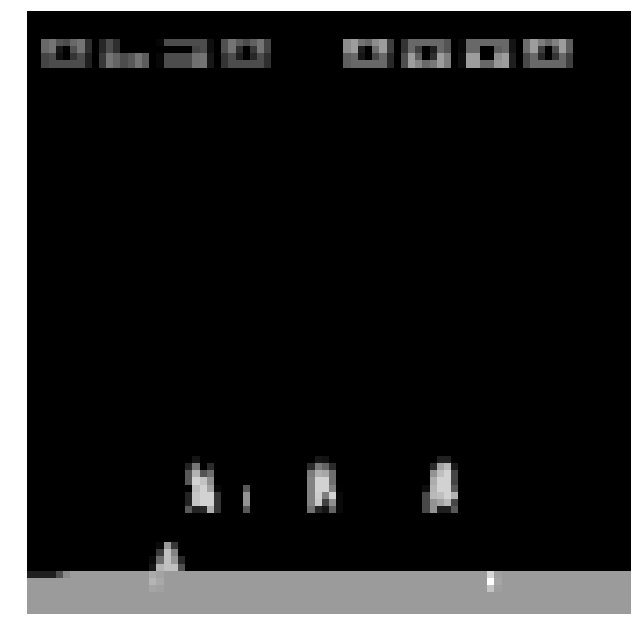}
\end{subfigure} \hspace*{-.5em}
\begin{subfigure}[t]{0.14\textwidth}
  \centering
    \includegraphics[width=1.0\textwidth]{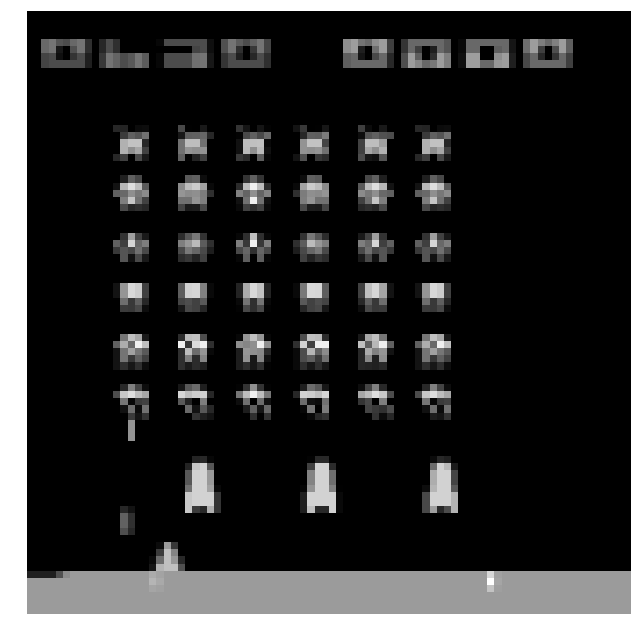}
\end{subfigure}
\end{adjustbox}
\caption{Trajectory during training}
\label{fig:spaceinvaders-train-1}
\end{subfigure}
\caption{Adversarial examples for MsPacman (a)-(b) and SpaceInvaders (c)-(d)}
\end{figure*}
\subsection{Evaluating the reconstructions}
To understand the effectiveness of these visualizations, we design a quantitative metric that measures the agreement between actions taken by the agent and the actions suggested using the reconstructed images.
With a fully trained model, we reconstruct images $\{s_1^a, s_2^a, \cdots s_N^a\}$ from the keys $\{h_1^a, h_2^a, \cdots h_N^a\}$ for all actions $a \in A$. In every state $s_t$, we induce another distribution on the Q-values using the cosine similarity in the image space,
\begin{align*}
w'(s_t)_i^a = Softmax\big(\dfrac{s_t \cdot s_i^a}{||s_t||_2 \cdot ||s_i^a||_2}\big)
\end{align*}
similar to $w(s_t)_i^a$ (which is also a distribution over Q-values but in the latent space). Using $w'(s_t)_i^a$, we can compute $Q'(s_t, a)$ and $U'(s_t, a)$ as before and select an action $a'_t = \argmax_{a\in A} Q'(s_t, a) + \lambda_{exp} U'(s_t, a)$. Using $a_t$ and $a'_t$, we define our metric of agreeability as 
\begin{align*}
Agreement = \dfrac{\mathbbm{1}_{a_t = a'_t}}{\mathbbm{1}_{a_t = a'_t} + \mathbbm{1}_{a_t \neq a'_t}}
\end{align*}
where $\mathbbm{1}$ is the indicator function. We measure this across multiple rollouts (5) using $a_t$ and average them . In Table~\ref{table:agreeability}, we report Agreement as a percentage for different encoder-decoder models. Unfortunately, the best agreement between the actions selected using the distributions in the image space and latent space is around $31\%$ for the unscaled color version of MsPacman. In MsPacman, the agent has 9 different actions and a random strategy would expect to have an Agreement of $\sim 11\%$. However, if the agreement scores were high (80-90\%), that would suggest that the Q-network is indeed learning to memorize configurations of objects seen during training. One explanation for the gap is that reconstructions rely heavily on generalizability to unseen keys.

\subsection{Adversarial examples that show memorization}
Looking at the visualizations and rollouts of a fully trained agent, we hand-craft a few out-of-sample environment states to examine the agent's generalization behavior. For example, in MsPacman, since visualizations suggest that the agent may be memorizing pacman's positions (also maybe ghosts and other objects), we simply add an extra pellet adjacent to a trajectory seen during training (Figure~\ref{fig:pacman-test-1}). The agent does not clear the additional pellet and simply continues to execute actions performed during training (Figure~\ref{fig:pacman-train-1}). Most importantly, the agent is extremely confident in taking actions initially (seen in attention maps Figure~\ref{fig:pacman-test-1}) which suggest that the extra pellet was probably not even captured by the embeddings. Similarly, in case of SpaceInvaders, the agent has a strong bias towards shooting from the leftmost-end (seen in Figure~\ref{fig:invert-key-25-spaceinvaders}). This helps in clearing the triangle like shape and moving to the next level (Figure~\ref{fig:spaceinvaders-train-1}). However, when triangular positions of spaceships are inverted, the agent repeats the same strategy of trying to shoot from left and fails to clear ships (Figure~\ref{fig:spaceinvaders-test-1}). These examples indicate that the features extracted by the convolutional channels seem to be shallow. The agent does not really model interactions between objects. For example in MsPacman, after observing 10M frames, it does not know general relationships between pacman and pellet or ghosts. Even if optimal Q-values were known, there is no incentive for the network to model these higher order dependencies when it can work with situational features extracted from finite training examples. \cite{zhang2018study} also report similar results on simple mazes where an agent trained on insufficient environments tends to repeat training trajectories on unseen mazes.

\subsection{Sensitivity to hyperparameters}
I-DQN's objective function introduces four hyperparameters for weighting the different loss components ($\lambda_1$: for bellman error, $\lambda_2$: distributional error, $\lambda_3$: reconstruction error and $\lambda_4$: diversity error). The diversity error forces attention over multiple keys (examples for $\lambda_4 = 0$ and $\lambda_4 = 0.01$ are shown in the supplementary material). In general, we found the values $\lambda_1 =1.0$, $\lambda_2 = 1.0$, $\lambda_3 = 0.05$, $\lambda_4 = 0.01$ to work well across games (detailed list of hyperparameters and their values is reported in the supplementary material). We ran experiments for different settings of $\lambda_1$, $\lambda_2$ and $\lambda_3$ keeping $\lambda_4 = 0.01$ constant on Ms Pacman (averaged over 3 trials). In general, increasing $\lambda_3$ (coefficient on reconstruction error) to $0.5$ and $5.0$ yields visually better quality reconstructions but poorer scores- $3,245.3$ and $3,013.1$ respectively (drop by $\sim 45$\%). Increasing $\lambda_1 = \lambda_2 = 10.0 $ also drops the score and yields poor reconstructions (sometimes without the objects of the game which loses interpretability) but converges quite quickly ($5,267.15$, drop by $\sim 12$\%). Out of $\lambda_1$ (bellman loss) and $\lambda_2$ (distributional loss), $\lambda_2$  seems to play a more important in reaching higher scores compared to $\lambda_1$ [$\lambda_1=1.0$, $\lambda_2=10.0$ score: $4,916.0$ ; $\lambda_1=10.0$, $\lambda_2=1.0$, score: $4,053.6$]. So, the setting $\lambda_1 =1.0$, $\lambda_2 = 1.0$, $\lambda_3 = 0.05$, $\lambda_4 = 0.01$ seems to find the right balance between the q-value learning losses and regularizing losses. For the exploration factor, we tried a few different values $\lambda_{\text{exp}} = \{0.1,  0.01, 0.001\}$ and it did not have a significant effect on the scores.

\section{Conclusion}
In this paper, we propose an interpretable deep Q-network model (i-DQN) that can be studied using a variety of tools including the usual saliency maps, attention maps and reconstructions of key embeddings that attempt to provide global explanations of the model's behavior. We also show that the uncertainty in soft cluster assignment can be used to drive exploration effectively and achieve high training rewards comparable to other models. Although the reconstructions do not explain the agent's decisions perfectly, they provide a better insight into the kind of features extracted by convolutional layers. This can be used to design interesting adversarial examples with slight modifications to the state of the environment where the agent fails to adapt and instead repeats action sequences that were performed during training. This is the general problem of overfitting in machine learning but is more acute in the case of reinforcement learning because the process of collecting training examples depends largely on the agent's biases (exploration). There are many interesting directions for future work. For example, we know that the reconstruction method largely affects the visualizations and other methods such as generative adversarial networks (GANs) \cite{goodfellow2014generative} can model latent spaces more smoothly and could generalize better to unseen embeddings. Another direction is to see if we can automatically detect the biases learned by the agent and design meaningful adversarial examples instead of manually crafting test cases.

\FloatBarrier
\section{Acknowledgments}
This work was funded by awards AFOSR FA9550-15-1-0442 and NSF IIS- 1724222.

\bibliographystyle{aaai}
\bibliography{ref}

\newpage
\section{Supplemental Material}
In this supplemental material, we include details of the experiment setup, hyperparameters and experiment results in terms of training curves. We present additional visualizations for the other games discussed in the paper and a few more adversarial examples for MsPacman. We also include an ablation study on the effect of the different loss functions.

\section{Experimental Setup}
The i-DQN training algorithm is described in Algorithm~\ref{alg:training} (including the directed exploration strategy and visualization step). As discussed in the paper, the i-DQN model attempts to minimize a weighted linear combination of multiple losses. 
\begin{dmath*}
\mathcal{L}_{final}(\theta) = \lambda_1 \mathcal{L}_{bellman}(\theta) + \lambda_2 \mathcal{L}_{distrib}(\theta) + \lambda_3 \mathcal{L}_{reconstruct}(\theta)  + \lambda_4 \mathcal{L}_{diversity}(\theta)
\end{dmath*}
\begin{algorithm}[h!]
\SetAlgoNoLine
    \tcp{initialization}
    Sample $N$ independent values $\{v_1, v_2, \cdots, v_N\}$ uniformly at random from $(V_{min}, V_{max})$ \\
    Initialize network parameters $\theta$, \\
    \qquad Conv/Deconv layers using Xavier initialization \\
    \qquad Linear layers and Keys $h_i^a \sim \mathcal{N}(0, 0.1)$ \\
 \For{$t=1$ \KwTo $T$}{
    \tcp{action selection}
    Compute embedding for $s_t$: $h(s_t) = Conv(s_t)$
    \For{action, $a=1$ \KwTo $A$}{
        Compare $h(s_t)$ to keys $\{h_1^a, h_2^a, \cdots, h_N^a\}$: \\ 
        \qquad $d(h(s_t), h_i^a) = h(s_t)\cdot h_i^a$ \\
        $w(s_t)^a_i = Softmax(d(h(s_t, h_i^a))$
    }
    Compute $Q(s_t, a)$ and $U(s_t, a)$ \\
    Select $a_t = \argmax_{a\in A} Q(s_t, a) + \lambda_{exp} U(s_t, a)$ and store transition. \\
    \tcp{update network}
    Sample random mini-batch $(s_t, a_t, s_{t+1}, R_t)$ \\
    Compute losses $\mathcal{L}_{bellman}$, $\mathcal{L}_{distrib}$, $\mathcal{L}_{reconstruct}$, $\mathcal{L}_{diversity}$ and  $\mathcal{L}_{final}(\theta)$ and perform gradient update step on network parameters $\theta$
}
\tcp{visualization}
To visualize: pick action $a$, value index $i$ and perform $\hat{s^a_i} = Deconv(h^a_i)$
 \caption{Training the i-DQN model}
 \label{alg:training}
\end{algorithm}

\begin{figure*}[t!]
\centering
\begin{subfigure}[t]{0.3\textwidth}
  \raggedright
    \includegraphics[width=1.0\textwidth]{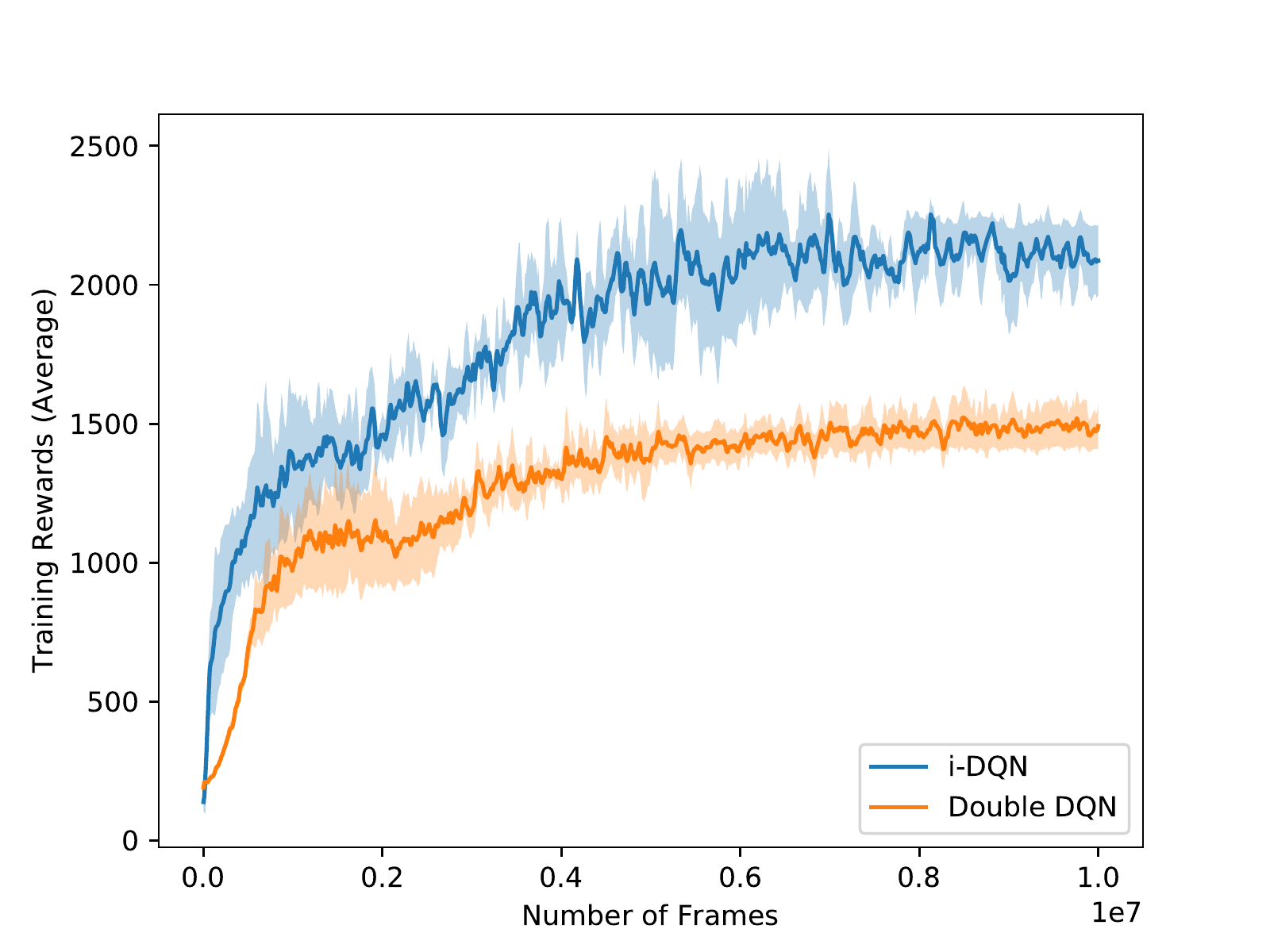}
    \caption{Alien}
\end{subfigure} \hspace{-1.5em}
\begin{subfigure}[t]{0.3\textwidth}
  \raggedright
    \includegraphics[width=1.0\textwidth]{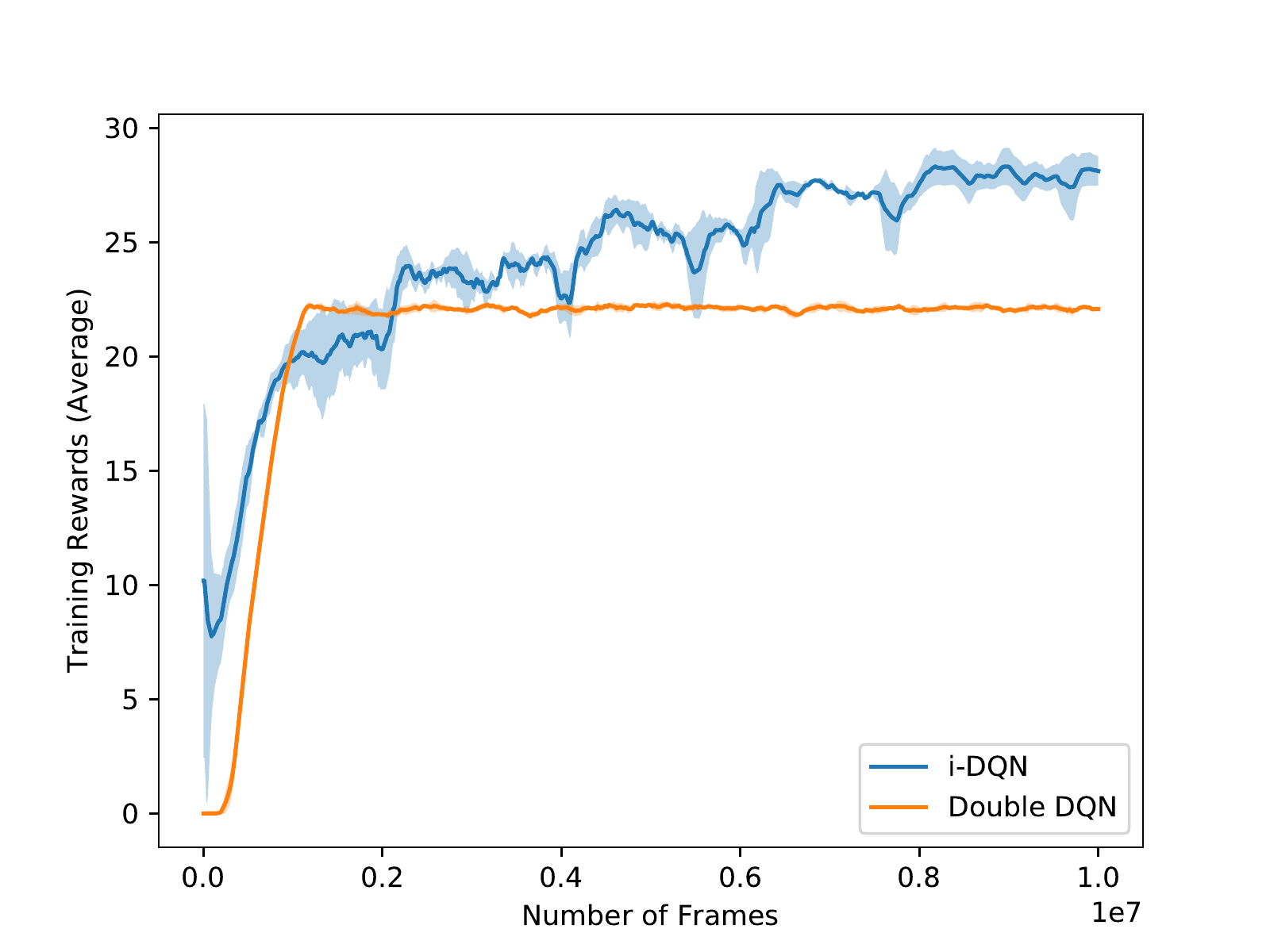}
    \caption{Freeway}
\end{subfigure} \hspace{-1.5em}
\begin{subfigure}[t]{0.3\textwidth}
  \raggedright
    \includegraphics[width=1.0\textwidth]{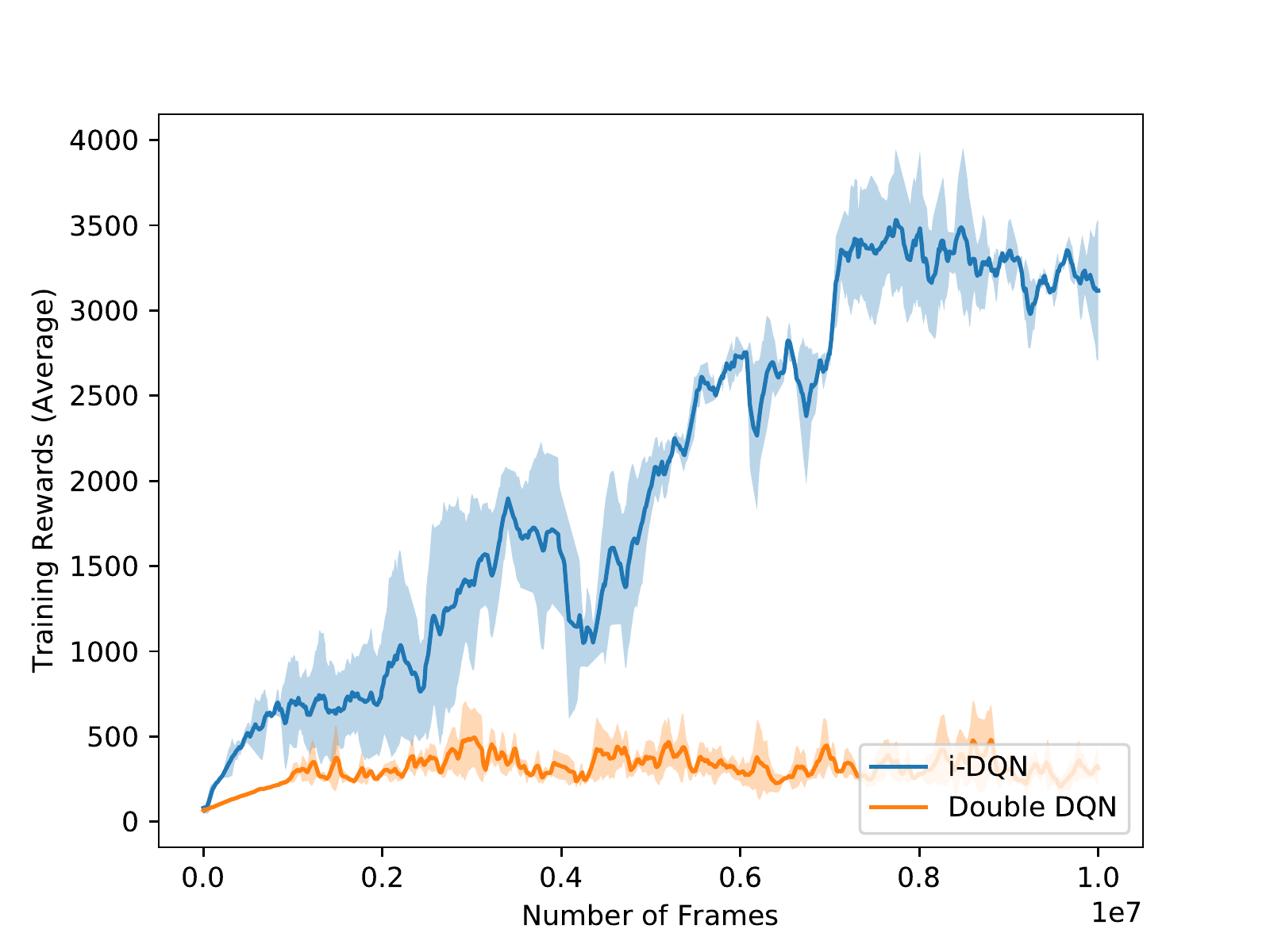}
    \caption{Frostbite}
\end{subfigure}\\
\begin{subfigure}[t]{0.3\textwidth}
  \raggedright
    \includegraphics[width=1.0\textwidth]{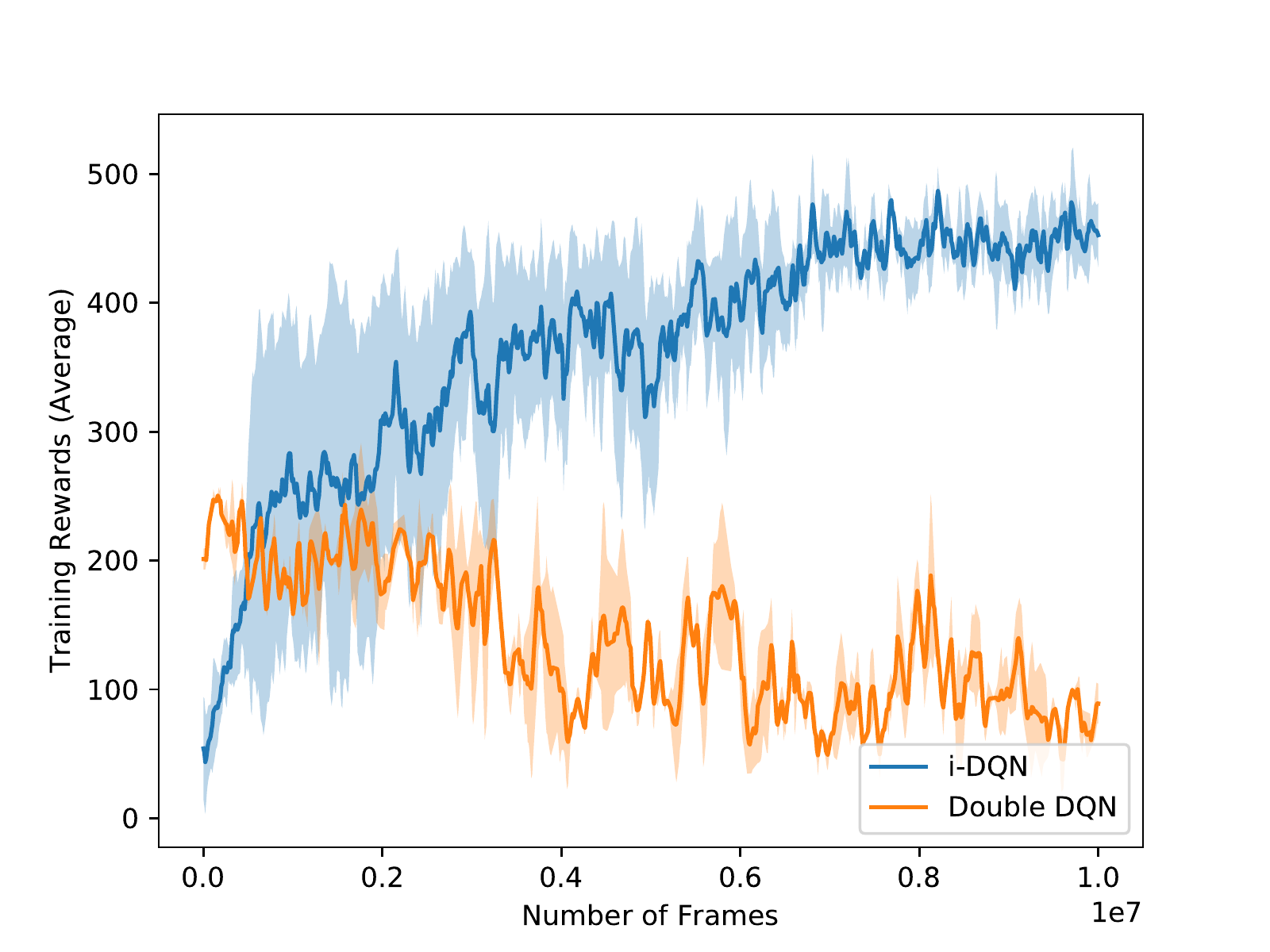}
    \caption{Gravitar}
\end{subfigure} \hspace{-1.5em}
\begin{subfigure}[t]{0.3\textwidth}
  \raggedright
    \includegraphics[width=1.0\textwidth]{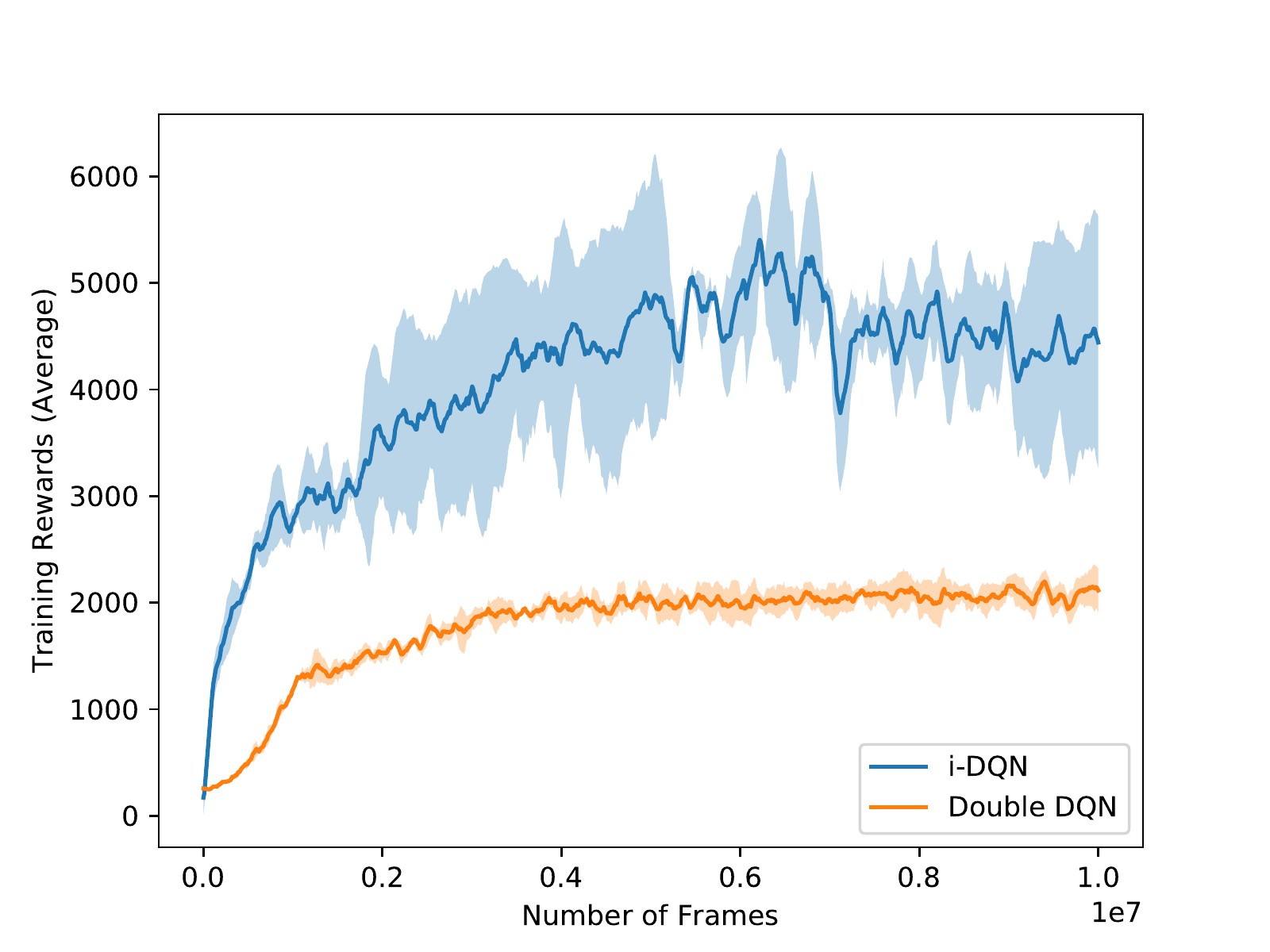}
    \caption{MsPacman}
\end{subfigure} \hspace{-1.5em}
\begin{subfigure}[t]{0.3\textwidth}
  \raggedright
    \includegraphics[width=1.0\textwidth]{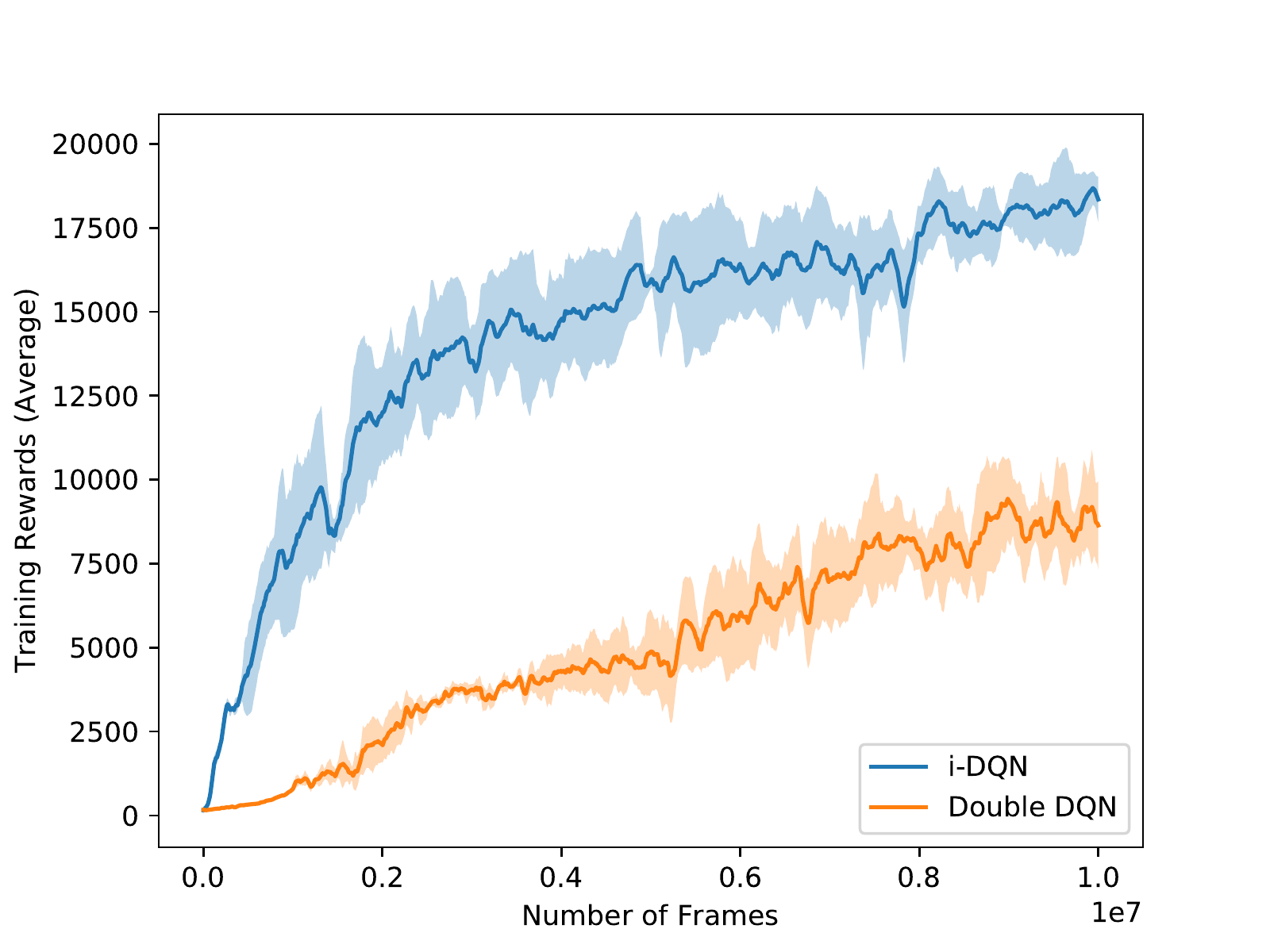}
    \caption{Qbert}
\end{subfigure}\\
\begin{subfigure}[t]{0.3\textwidth}
  \raggedright
    \includegraphics[width=1.0\textwidth]{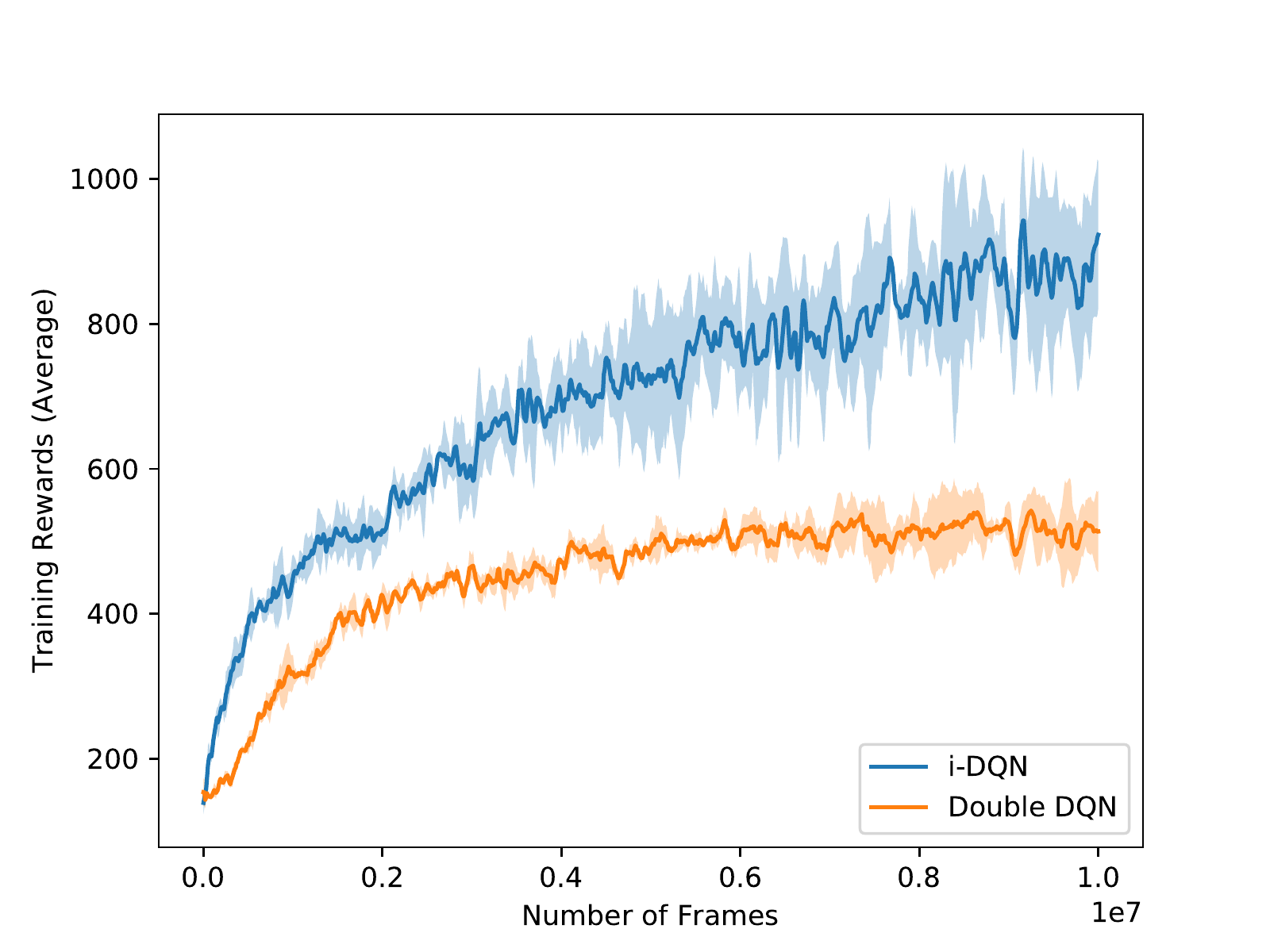}
    \caption{SpaceInvaders}
\end{subfigure} \hspace{-1.5em}
\begin{subfigure}[t]{0.3\textwidth}
  \raggedright
    \includegraphics[width=1.0\textwidth]{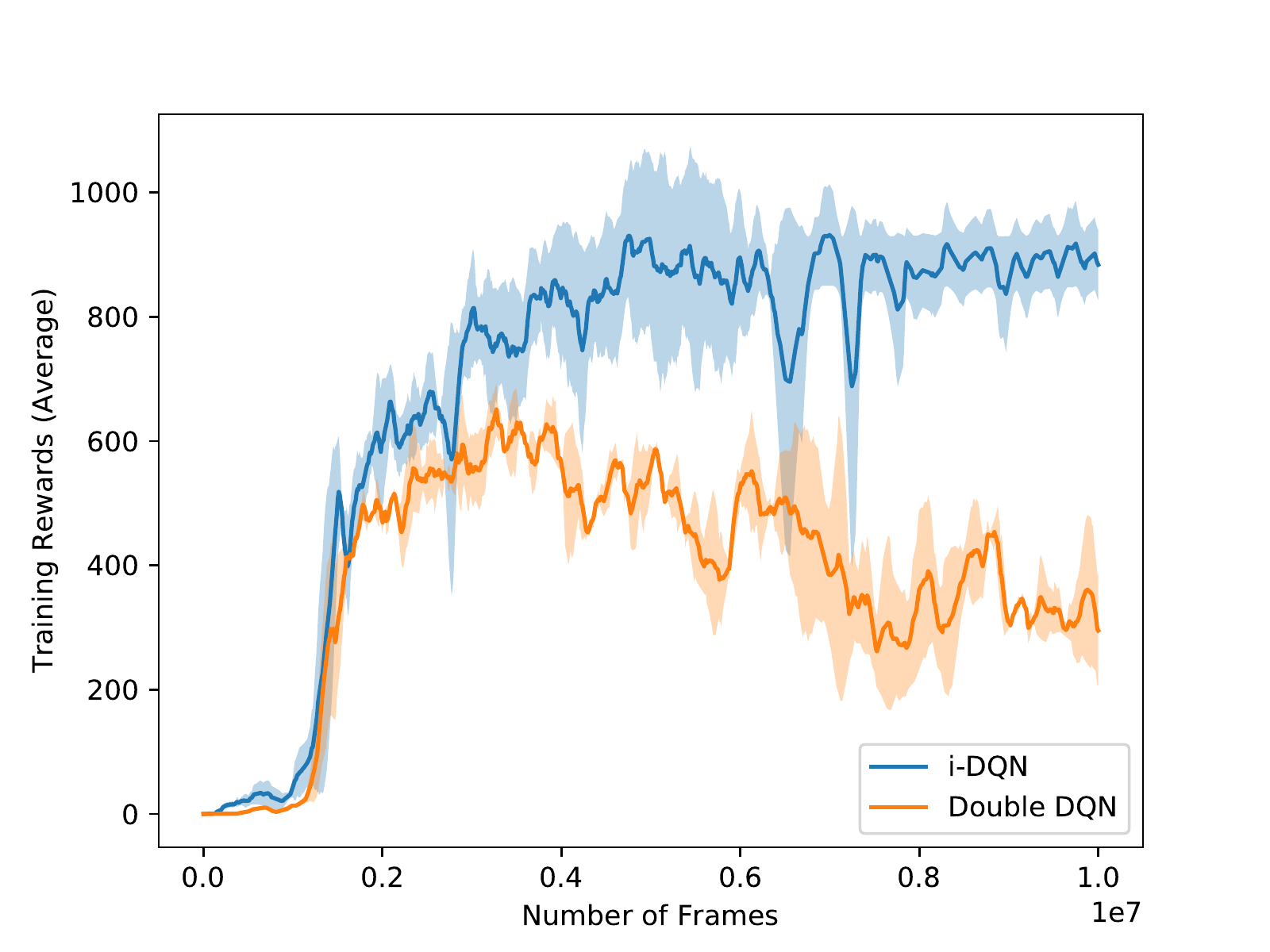}
    \caption{Venture}
\end{subfigure} \hspace{-1.5em}
\caption{Training curves comparing average rewards (over 100 episodes) for i-DQN and double DQN (the shaded region indicates variance over 3 random seeds). The blue curves are the i-DQN model using a directed exploration strategy while the green curve is our implementation of double DQN model using epislon greedy exploration.}
\label{fig:training-curves}
\end{figure*}

Table~\ref{table:hyperparams} lists all the hyperparameters (including network architecture), values and their description. Most of the training procedure and hyperparameter values are similar to the settings in \cite{van2016deep} and \cite{bellemare2017distributional}. Figure~\ref{fig:training-curves} shows the training curves for i-DQN and double DQN models over three random seeds for eight different Atari environments- Alien, Freeway, Frostbite, Gravitar, MsPacman, Qbert, SpaceInvaders,  and Venture. The results reported in the paper are averaged over the best results for individual runs for each seed. Although training i-DQN is slow, the directed exploration strategy helps to quickly reach scores competitive with the state-of-the-art deep Q-learning based models.  \\
\begin{table*}[t!]
\centering
  \begin{tabular}{p{4.5cm}p{3cm}p{4.5cm}}
  \hline
    \textbf{Hyperparameter} & \textbf{Value} & \textbf{Description}\\ \hline
    Number of Keys per Action ($N$) & 20 & Total keys in key-value store for $A$ actions $ = N \times A$ \\
    Range of Values $(V_{min}$, $V_{max})$ & $(-25, 25)$ & The range from which the values of key-value store are sampled \\
    Exploration Factor $\lambda_{exp}$ & $0.01$ & Controls the ucb style action selection using confidence intervals \\
    Embedding Size & 256 & \specialcell[t]{Dimensions of latent space where \\ keys and state embeddings lie} \\
    Network Channels & $(32, 8 \times 8, 4), (64, 4 \times 4, 2), (64, 3 \times 3, 1)$ & Convolution layers (Channels, Filter, Stride) \\
    Network Activations & ReLU &  \\
    Discount Factor & 0.99 & \\
    Batch Size & 32 & \\
    Optimizer & Adam & \specialcell[t]{learning rate$=0.00025$, \\beta1$=0.9$, beta2$=0.999$,\\ weight decay$=0$} \\ 
    Gradient Clipping & 10 & Using gradient norm \\
    Replay Buffer Size & 10K & \\
    Training Frequency & Every 4 steps & \\
    Target Network Sync Frequency & 1000 & Fully replaced with weights from source/training network \\
    Frame Preprocessing & Standard & Grayscaling, Downsampling (84, 84), Frames stacked: 4, Repeat Actions: 4 \\
    Reward Clipping & No & \\
    Loss Factors & $\lambda_{1}=1.0$, $\lambda_{2}=1.0$, $\lambda_{3}=0.05$, $\lambda_{4}=0.01$ & The weights on each loss function\\ \hline
  \end{tabular}
  \caption{List of hyper-parameters and their values}
  \label{table:hyperparams}
\end{table*}

\textit{Attention Maps}: During training, the i-DQN model learns to attend over keys in the key-value store using a dot-product attention.
\begin{align*}
  w(s_t)_i^a = \dfrac{\exp{(h(s_{t})\cdot h_i^a)}}{\sum_{j} \exp{(h(s_{t})\cdot h_j^a)}} \label{attention_eq}
\end{align*}
We can visualize these weights in terms of attention heat-maps where each row is a distribution over Q-values (because of the softmax) for a particular action. Darker cells imply higher attention weights, for example in Figure~\ref{fig:diversity}, the agent expects a negative return of -25 with very high probability for action 'Left'. \\

\textit{Saliency Maps}: They usually highlight the most sensitive regions around the input that can cause the neural network to change its predictions. Following \cite{greydanus2017visualizing}, we define saliency maps as follows,
\begin{align*}
    \phi(s_t, i, j) = s_t \odot (1-M(i,j)) + A(s_t, \sigma) \odot M(i,j) \\
    S(s_t, a, i, j) = \dfrac{1}{2}||Q(\phi(s_t, i, j), a) - Q(s_t, a)||^2
\end{align*}
$\phi(s_t, i, j)$ defines the perturbed image for state $s_t$ at pixel location $(i,j)$, $M(i, j)$ is the saliency mask usually a Gaussian centered at $(i,j)$ defining the sensitive region, $A(s_t, \sigma)$ is a Gaussian blur of $s_t$ and $S(s_t, a, i, j)$ denotes the sensitivity of $Q(s_t, a)$ to pixel location $(i,j)$.

\section{Analysing Loss Functions}
\begin{figure*}[t!]
    \centering
    \begin{tabular}[t]{cc}
\begin{subfigure}{0.4\textwidth}
    \centering
    \smallskip
    \includegraphics[width=1.0\linewidth]{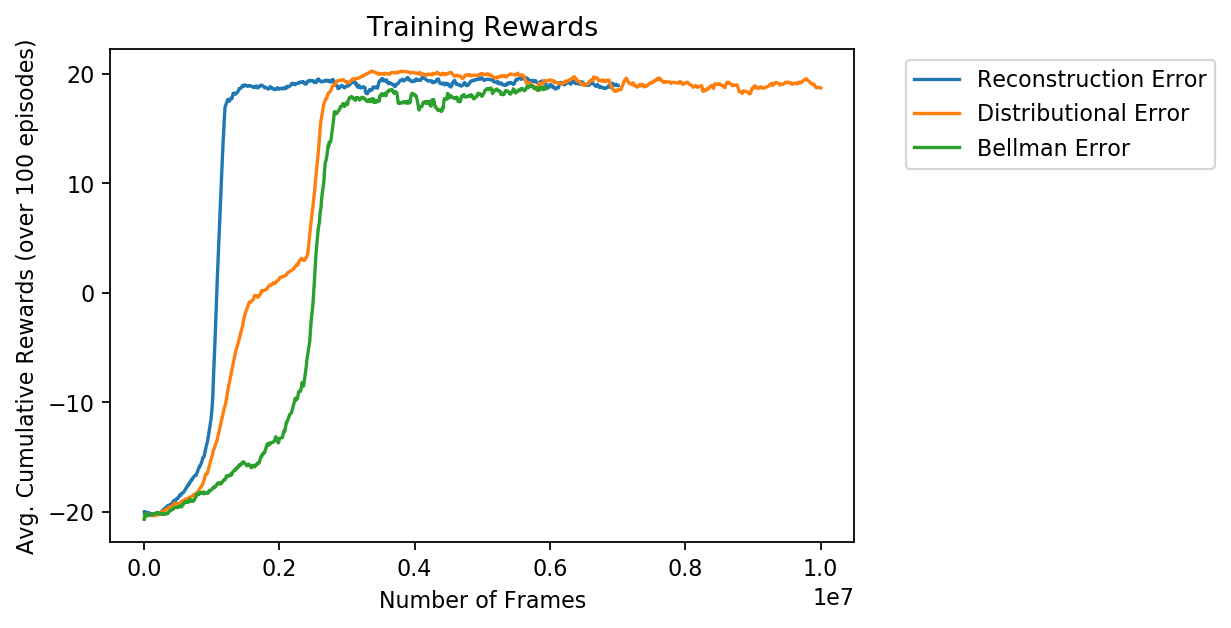}
    \caption{Ablation study with training rewards, keeping the diversity error constant (red curve is diversity + bellman losses, green curve is diversity + bellman + distributional losses and finally the blue curve is diversity + bellman + distributional + reconstruction losses)} 
    \label{fig:ablation}
\end{subfigure}
    &
        \begin{tabular}{c}
        \smallskip
            \begin{subfigure}[t]{0.35\textwidth}
                \centering
                \includegraphics[width=1.0\textwidth]{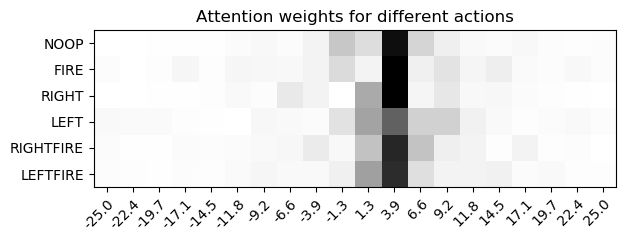}
                \caption{Attention Maps: Without diversity error}
                \label{fig:nodiversity}
            \end{subfigure}\\
            \begin{subfigure}[t]{0.35\textwidth}
                \centering
                \includegraphics[width=1.0\textwidth]{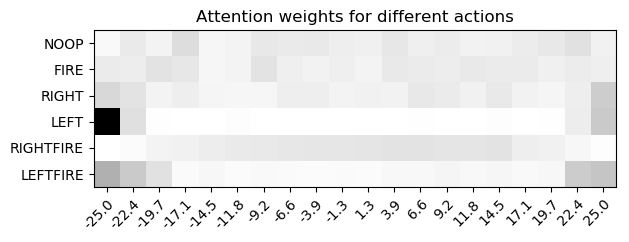}
                \caption{Attention Maps: With diversity error}
                \label{fig:diversity}
            \end{subfigure}
        \end{tabular}\\
    \end{tabular}
    \caption{Understanding the effect of different losses (PongNoFrameskip-v4 environment)}
\end{figure*}

Figure~\ref{fig:ablation} shows the effect of different loss functions. We keep the diversity error constant and incrementally add each of the loss functions. Figure~\ref{fig:nodiversity} and Figure~\ref{fig:diversity} show the effect of the diversity loss. Without the diversity loss, the network can learn to attend over a small set of the keys for each action. However, the diversity loss encourages attention across elements in a batch to be diverse and ideally, the attention weights should concentrate on a single key (because of l-$1$ and l-$2$ norm constraints) but concentrate over very few keys in practice .

\begin{figure*}[t!]
\centering
\begin{subfigure}[t]{0.12\textwidth}
  \centering
    \includegraphics[width=1.0\textwidth]{pacman-actions/DOWN-act-19.png}
    \caption{Down}
\end{subfigure} \hspace*{-0.25em}
\begin{subfigure}[t]{0.12\textwidth}
  \centering
    \includegraphics[width=1.0\textwidth]{pacman-actions/DOWNLEFT-act-19.png}
    \caption{Downleft}
\end{subfigure} \hspace*{-0.25em}
\begin{subfigure}[t]{0.12\textwidth}
  \centering
    \includegraphics[width=1.0\textwidth]{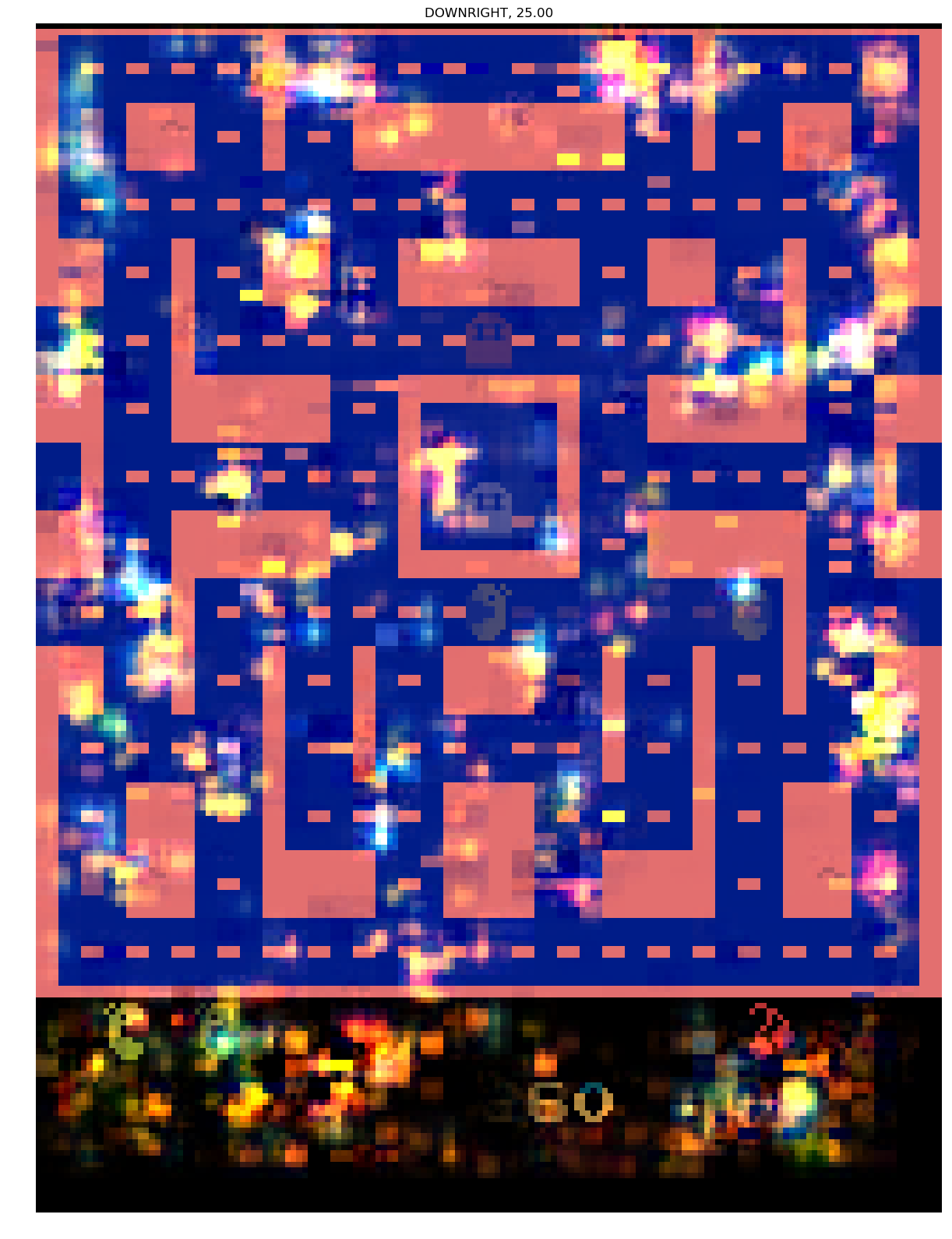}
    \caption{Downright}
\end{subfigure} \hspace*{-0.25em}
\begin{subfigure}[t]{0.12\textwidth}
  \centering
    \includegraphics[width=1.0\textwidth]{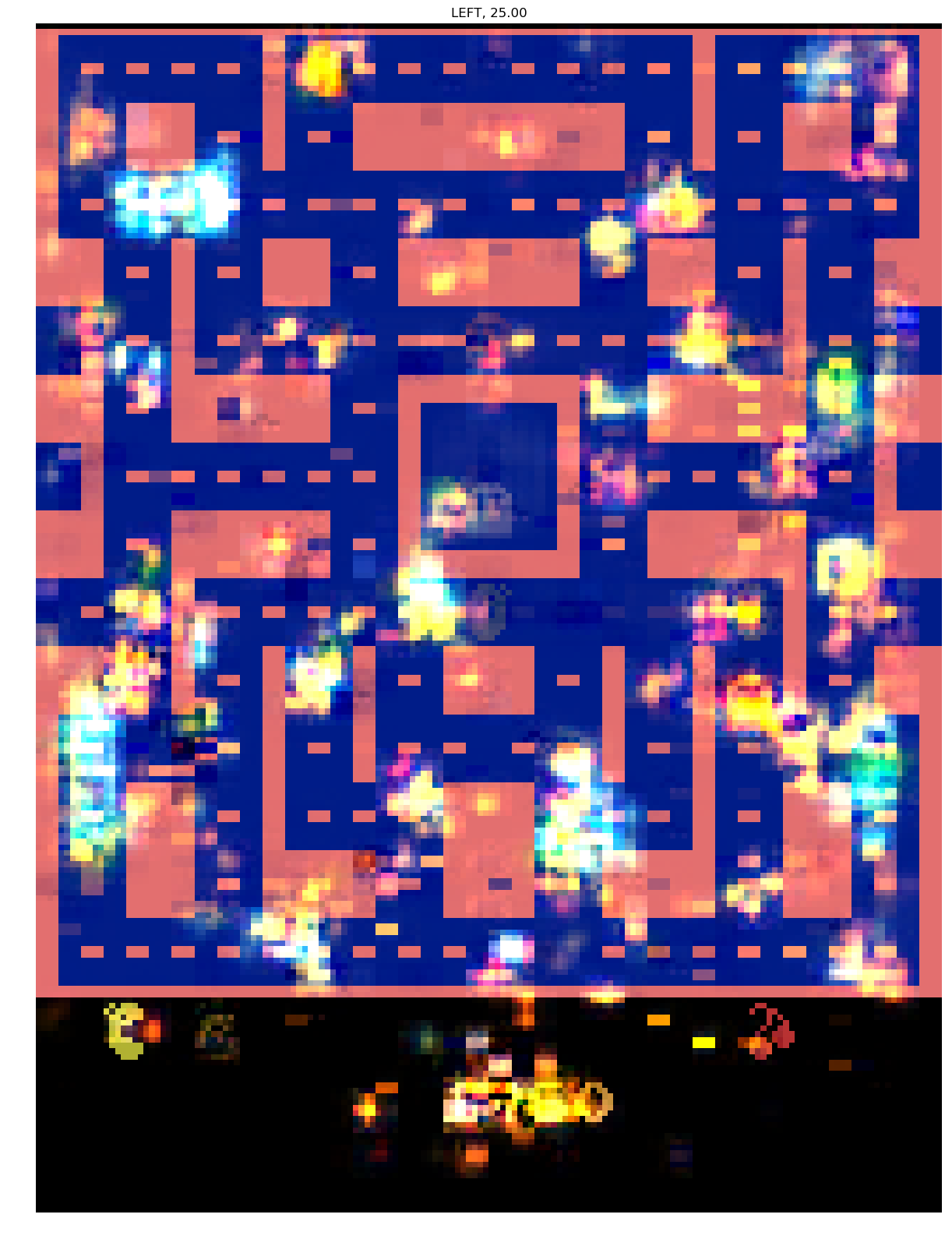}
    \caption{Left}
\end{subfigure} \hspace*{-0.25em}
\begin{subfigure}[t]{0.12\textwidth}
  \centering
    \includegraphics[width=1.0\textwidth]{pacman-actions/UP-act-19.png}
    \caption{Up}
\end{subfigure} \hspace*{-0.25em}
\begin{subfigure}[t]{0.12\textwidth}
  \centering
    \includegraphics[width=1.0\textwidth]{pacman-actions/UPLEFT-act-19.png}
    \caption{Upleft}
\end{subfigure} \hspace*{-0.25em}
\begin{subfigure}[t]{0.12\textwidth}
  \centering
    \includegraphics[width=1.0\textwidth]{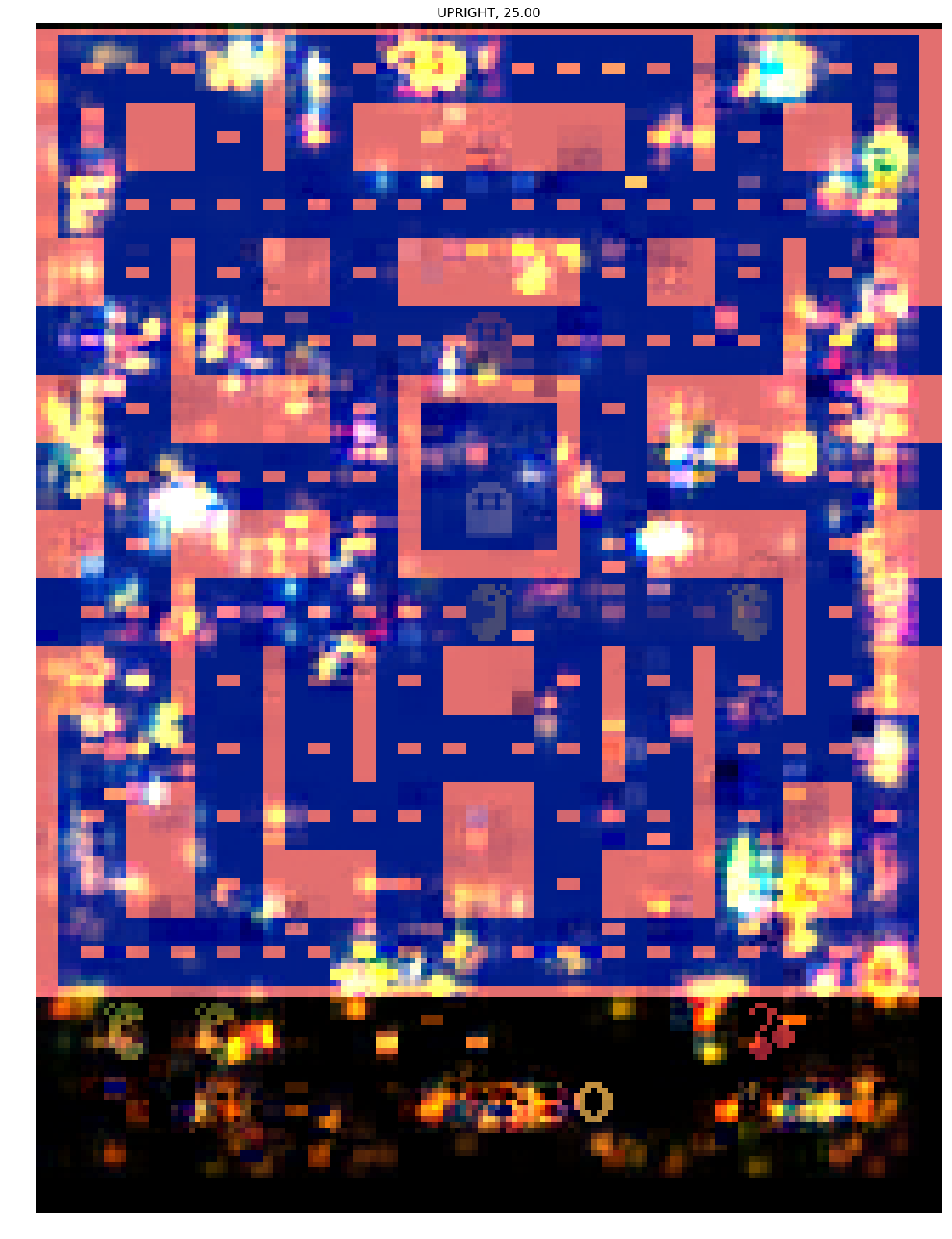}
    \caption{Upright}
\end{subfigure} \hspace*{-0.25em}
\begin{subfigure}[t]{0.12\textwidth}
  \centering
    \includegraphics[width=1.0\textwidth]{pacman-actions/RIGHT-act-19.png}
    \caption{Right}
    \label{fig:invert-key-25-pacman-right-supp}
\end{subfigure}
\caption{MsPacman, Inverting keys for different actions with Q-value 25}
\label{fig:invert-key-25-pacman-supp}
\begin{subfigure}[t]{0.12\textwidth}
  \centering
    \includegraphics[width=1.0\textwidth]{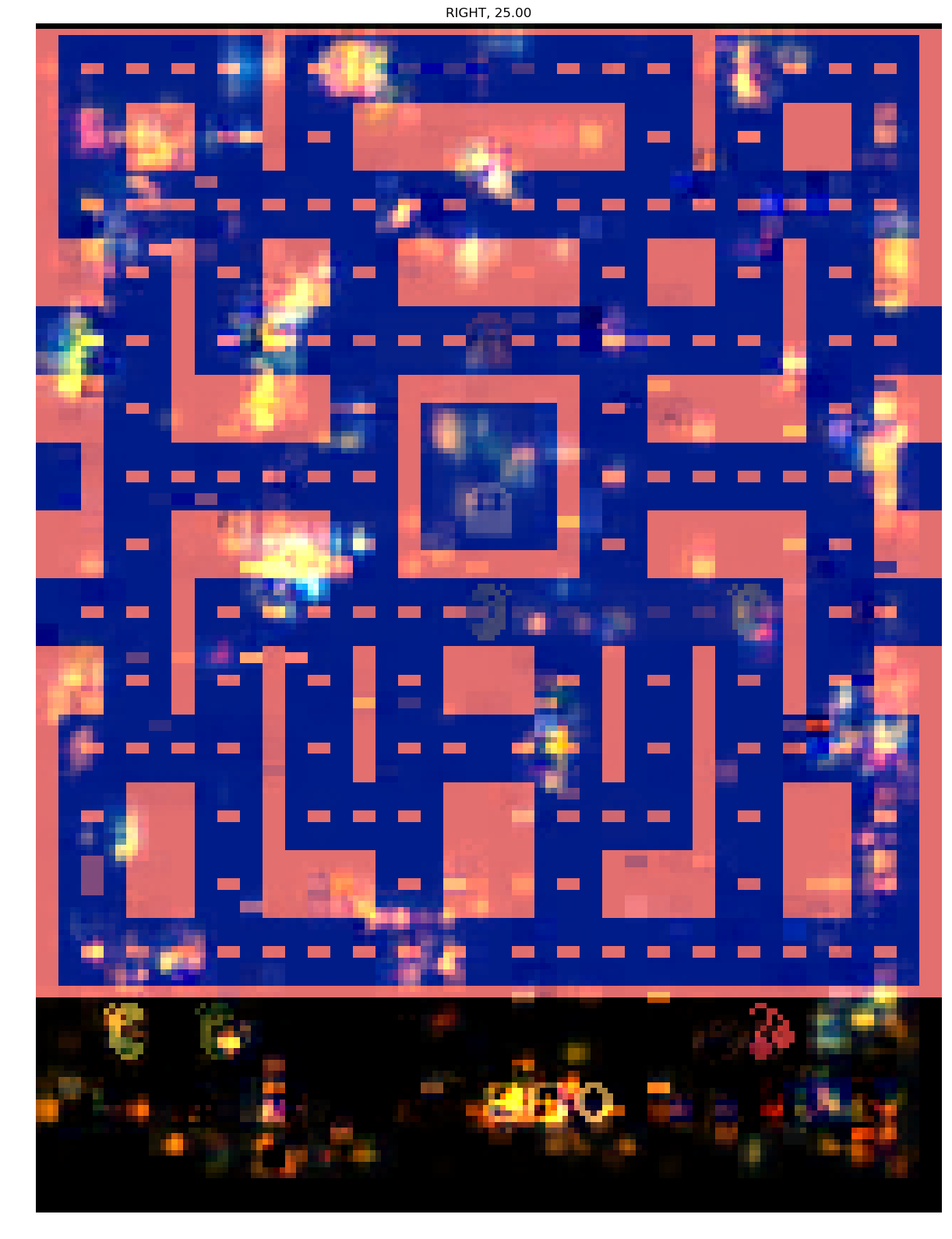}
\end{subfigure} \hspace*{-.1em}
\begin{subfigure}[t]{0.12\textwidth}
  \centering
    \includegraphics[width=1.0\textwidth]{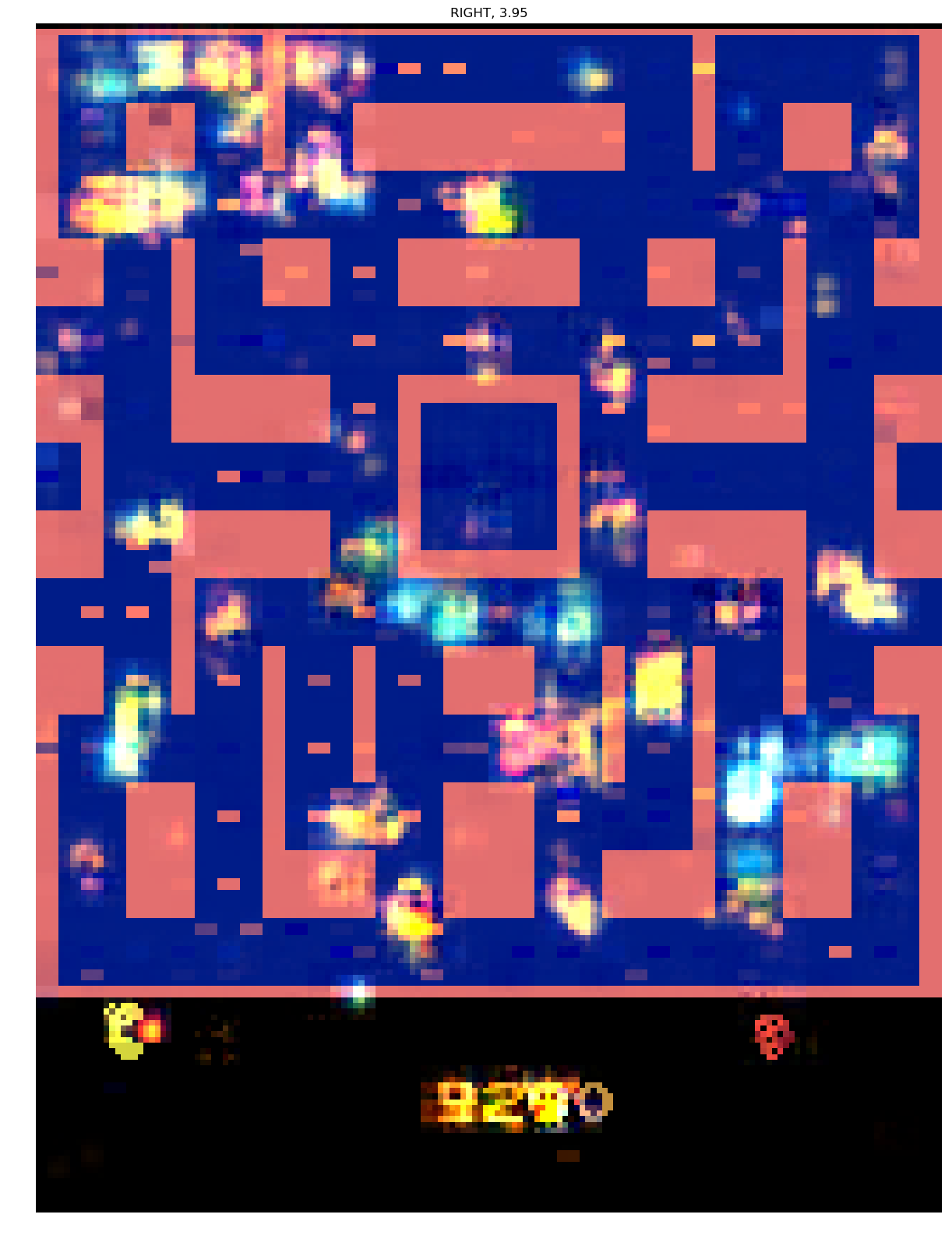}
\end{subfigure} \hspace*{-.1em}
\begin{subfigure}[t]{0.12\textwidth}
  \centering
    \includegraphics[width=1.0\textwidth]{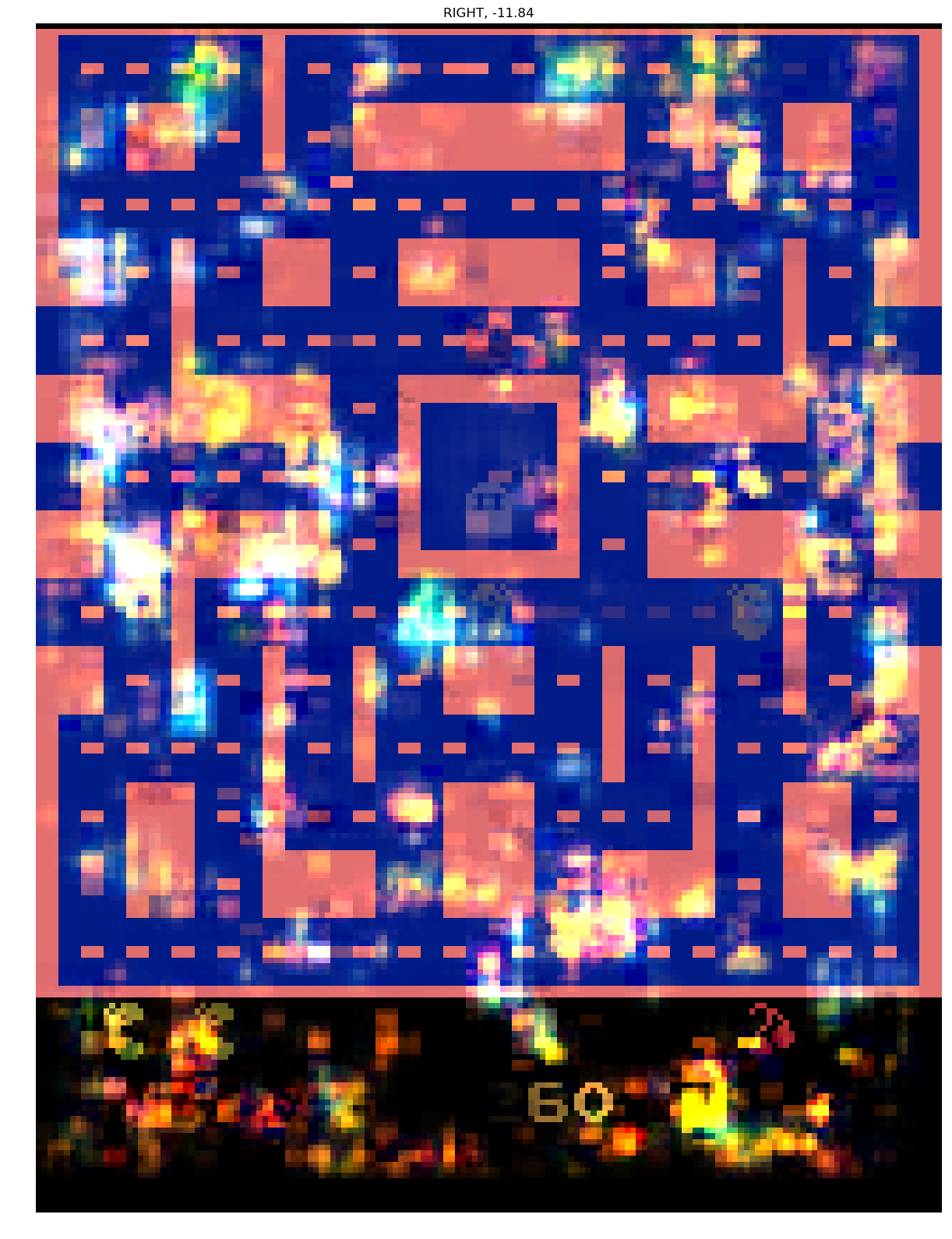}
\end{subfigure} \hspace*{-.1em}
\begin{subfigure}[t]{0.12\textwidth}
  \centering
    \includegraphics[width=1.0\textwidth]{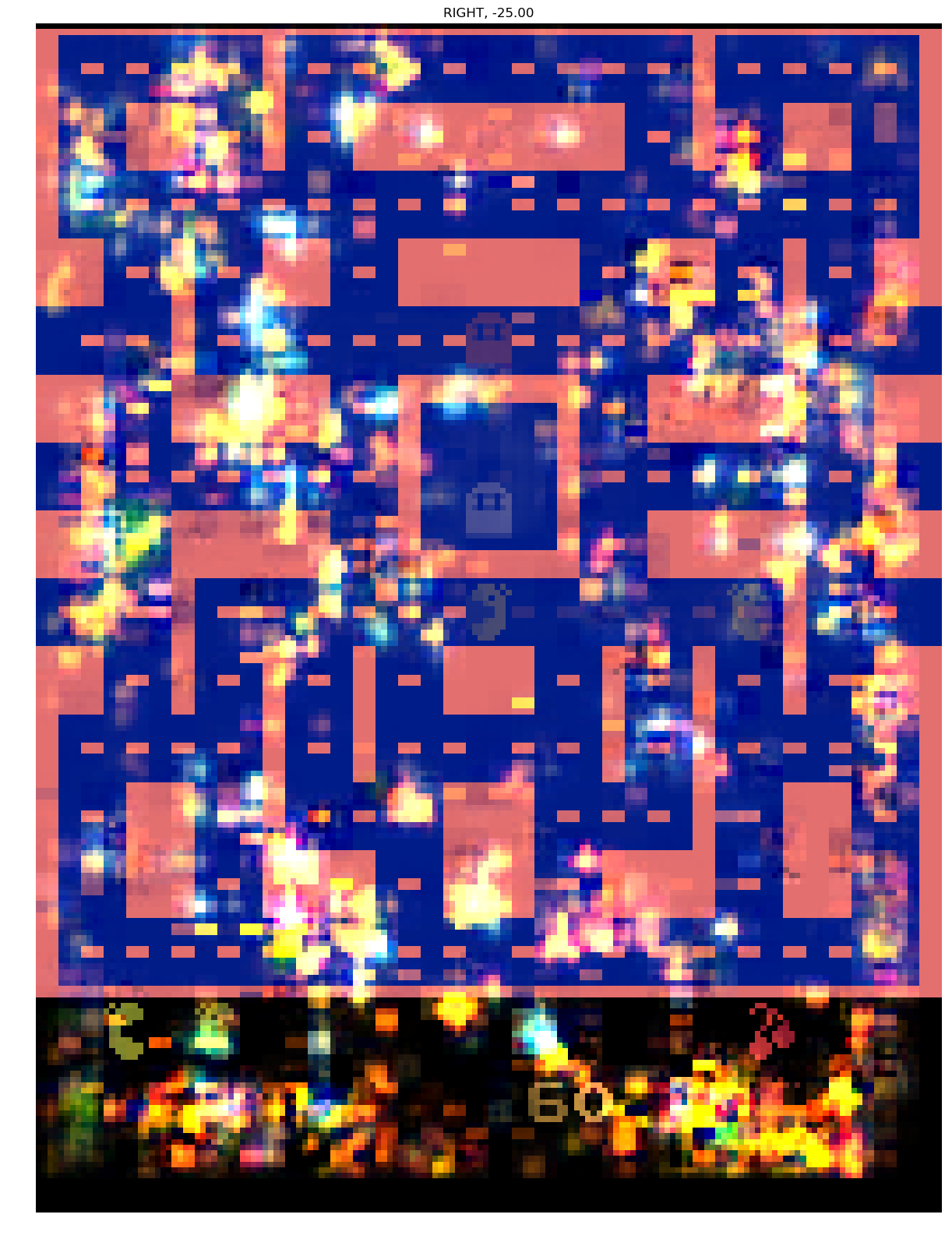}
\end{subfigure}
\caption{Inverting keys for action Right, Q-values $\{25, 3, -11, -25\}$}
\label{fig:invert-keys-right-pacman}
\end{figure*}

\begin{figure*}[t!]
\centering
\begin{subfigure}[t]{\textwidth}
\centering
\begin{subfigure}[t]{0.2\textwidth}
  \centering
    \includegraphics[width=1.0\textwidth]{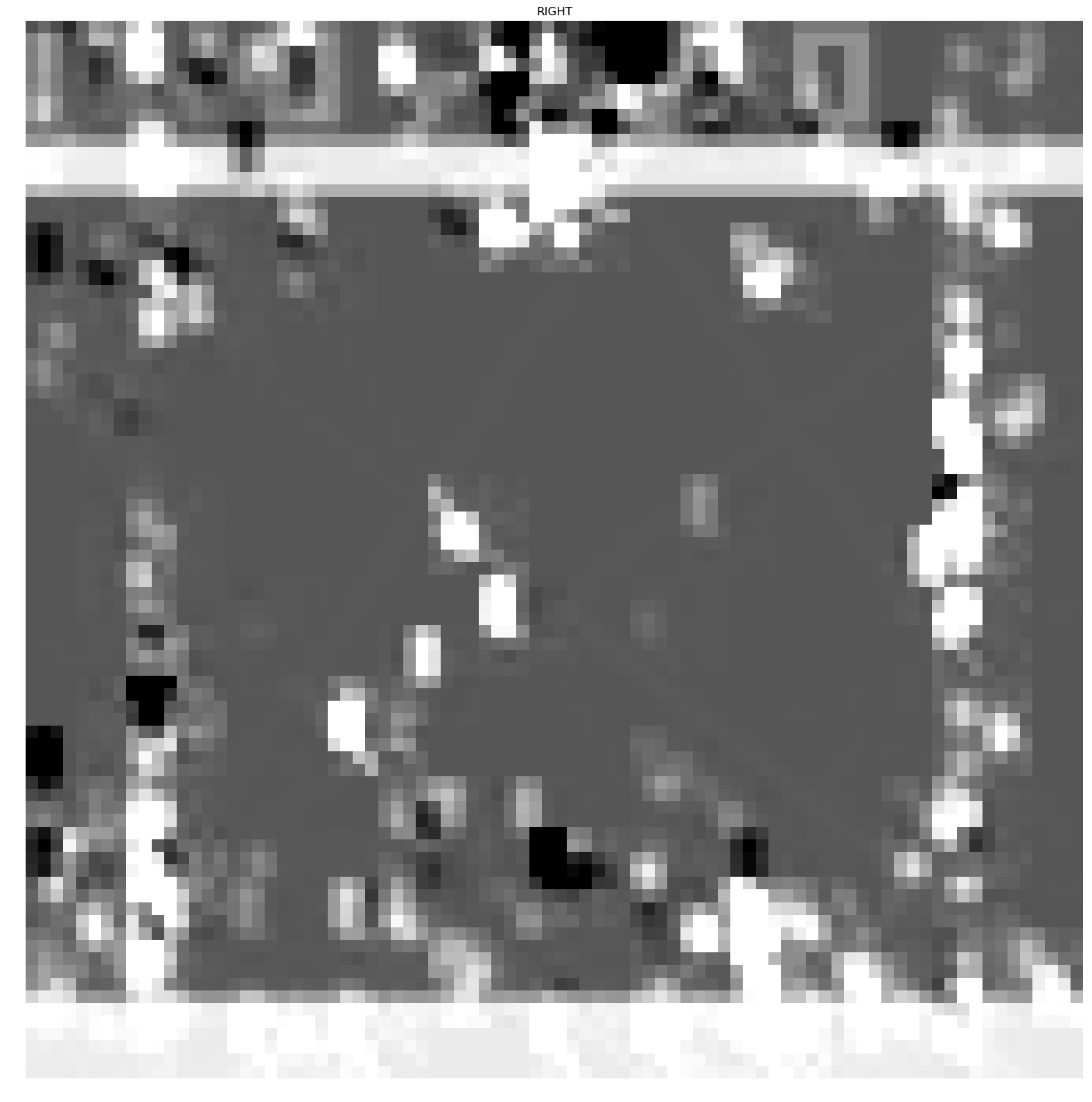}
    \captionsetup{labelformat=empty}
    \caption{Action: Right}
\end{subfigure} \hspace*{-.1em}
\begin{subfigure}[t]{0.2\textwidth}
  \centering
    \includegraphics[width=1.0\textwidth]{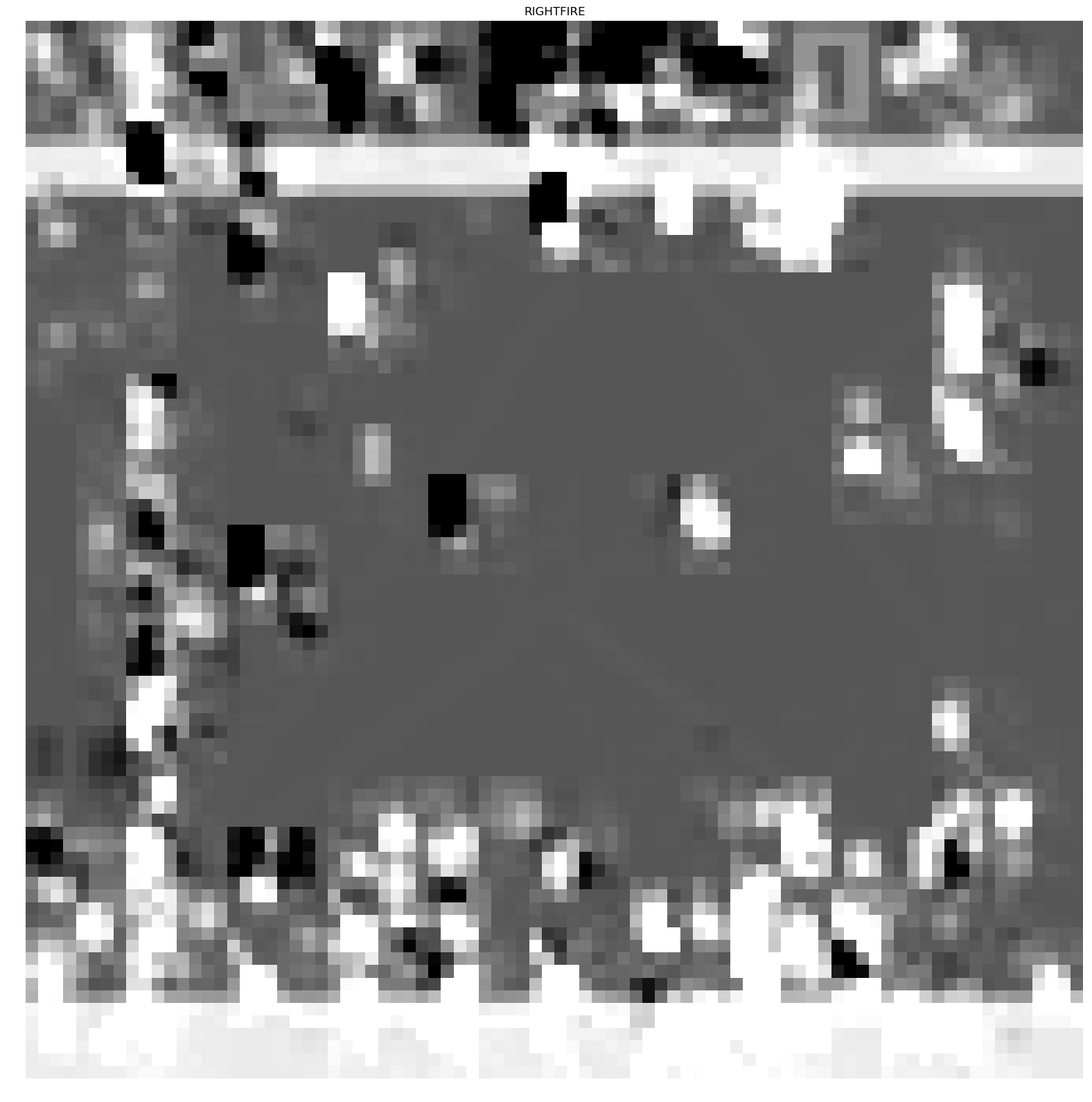}
    \captionsetup{labelformat=empty}
    \caption{Action: Right-Fire}
\end{subfigure} \hspace*{-.1em}
\begin{subfigure}[t]{0.2\textwidth}
  \centering
    \includegraphics[width=1.0\textwidth]{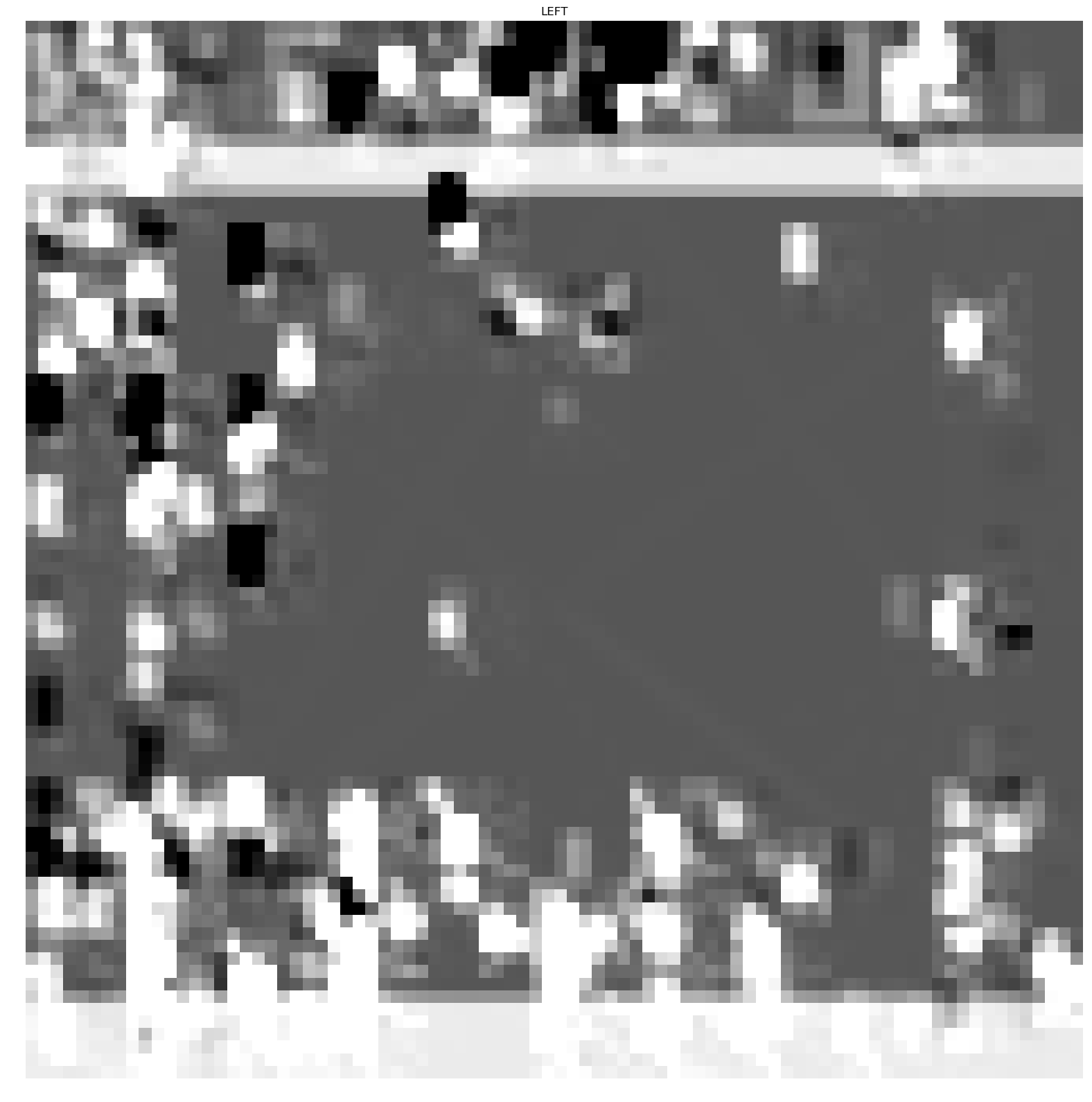}
    \captionsetup{labelformat=empty}
    \caption{Action: Left}
\end{subfigure} \hspace*{-.1em}
\begin{subfigure}[t]{0.2\textwidth}
  \centering
    \includegraphics[width=1.0\textwidth]{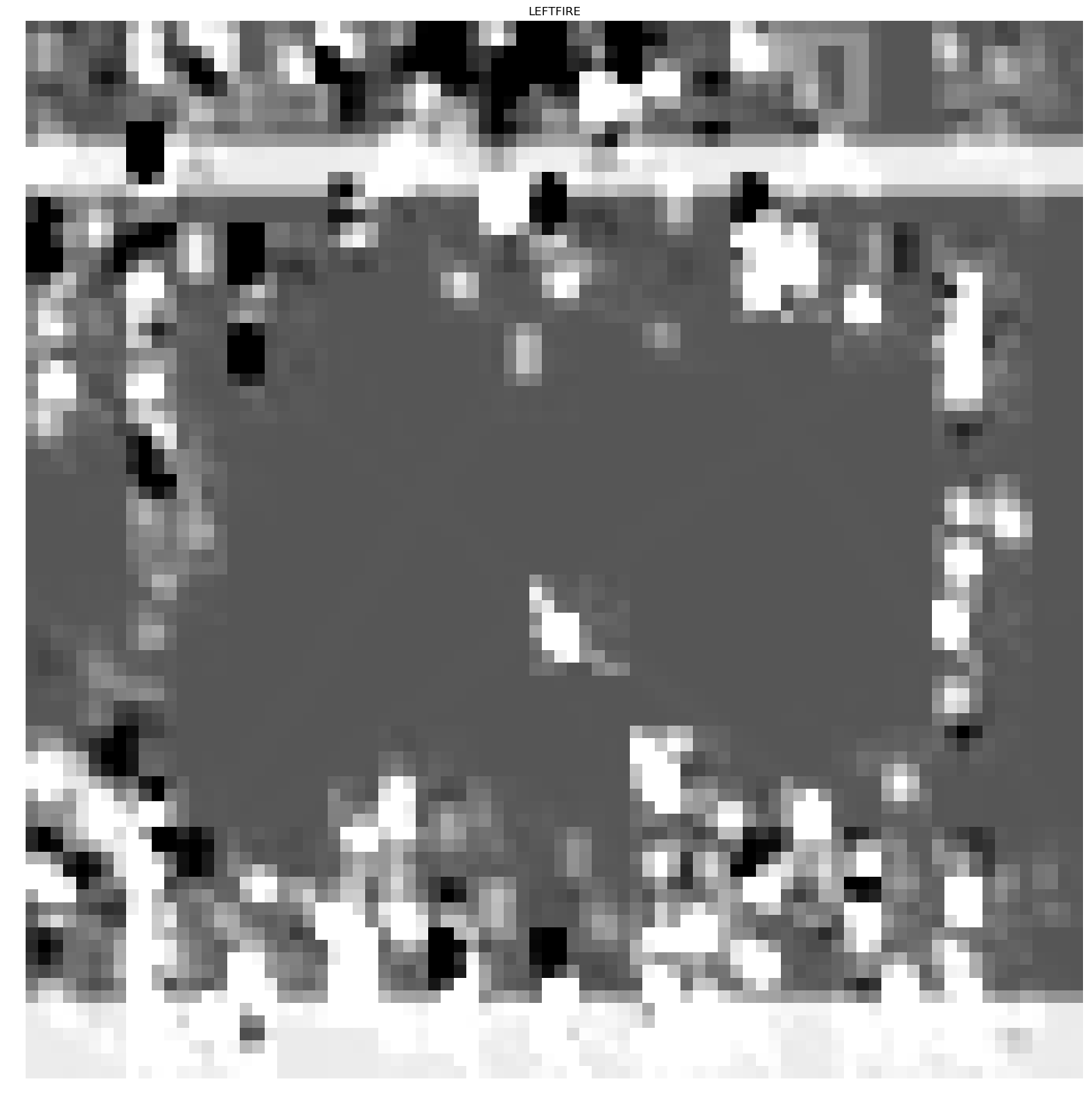}
    \captionsetup{labelformat=empty}
    \caption{Action: Left-Fire}
\end{subfigure}
\setcounter{subfigure}{0}
\caption{Pong, Inverting keys for different actions with Q-value: 25}
\label{fig:invert-keys-25-pong}
\end{subfigure}\\
\begin{subfigure}[t]{\textwidth}
\centering
\begin{subfigure}[t]{0.24\textwidth}
  \centering
    \includegraphics[width=1.0\textwidth]{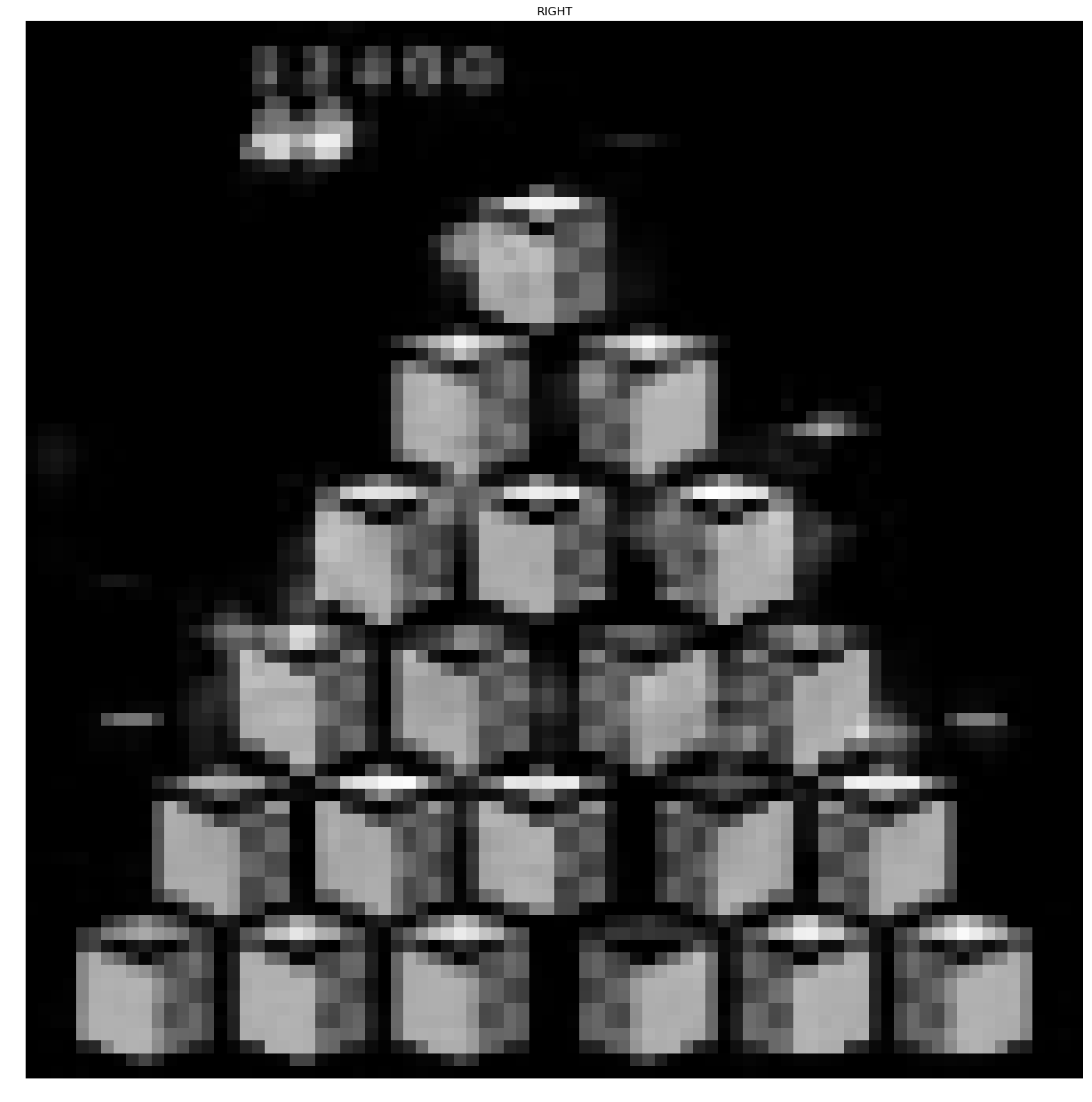}
    \captionsetup{labelformat=empty}
    \caption{Action: Right}
\end{subfigure} \hspace*{-.1em}
\begin{subfigure}[t]{0.24\textwidth}
  \centering
    \includegraphics[width=1.0\textwidth]{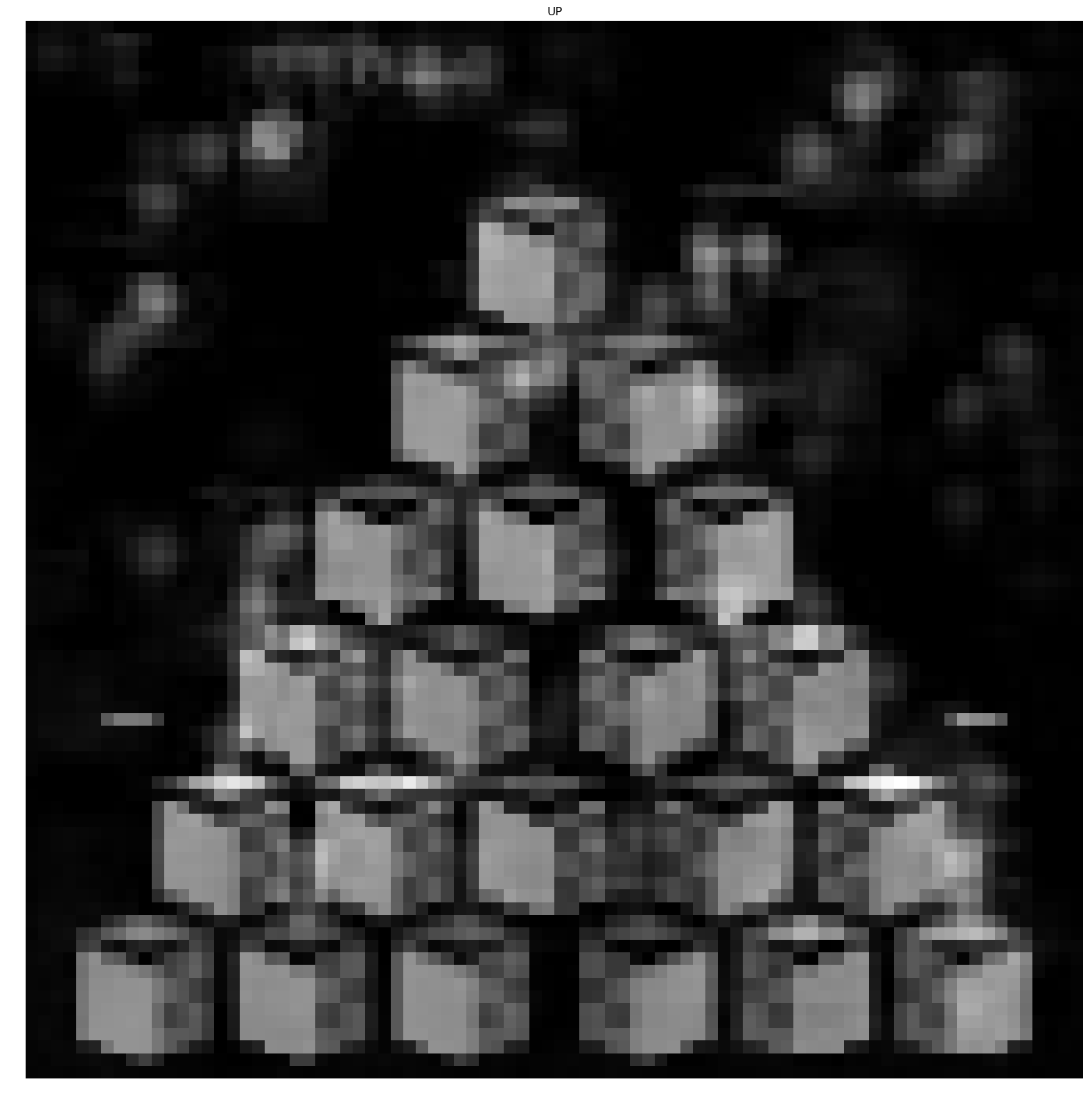}
    \captionsetup{labelformat=empty}
    \caption{Action: Up}
\end{subfigure} \hspace*{-.1em}
\begin{subfigure}[t]{0.24\textwidth}
  \centering
    \includegraphics[width=1.0\textwidth]{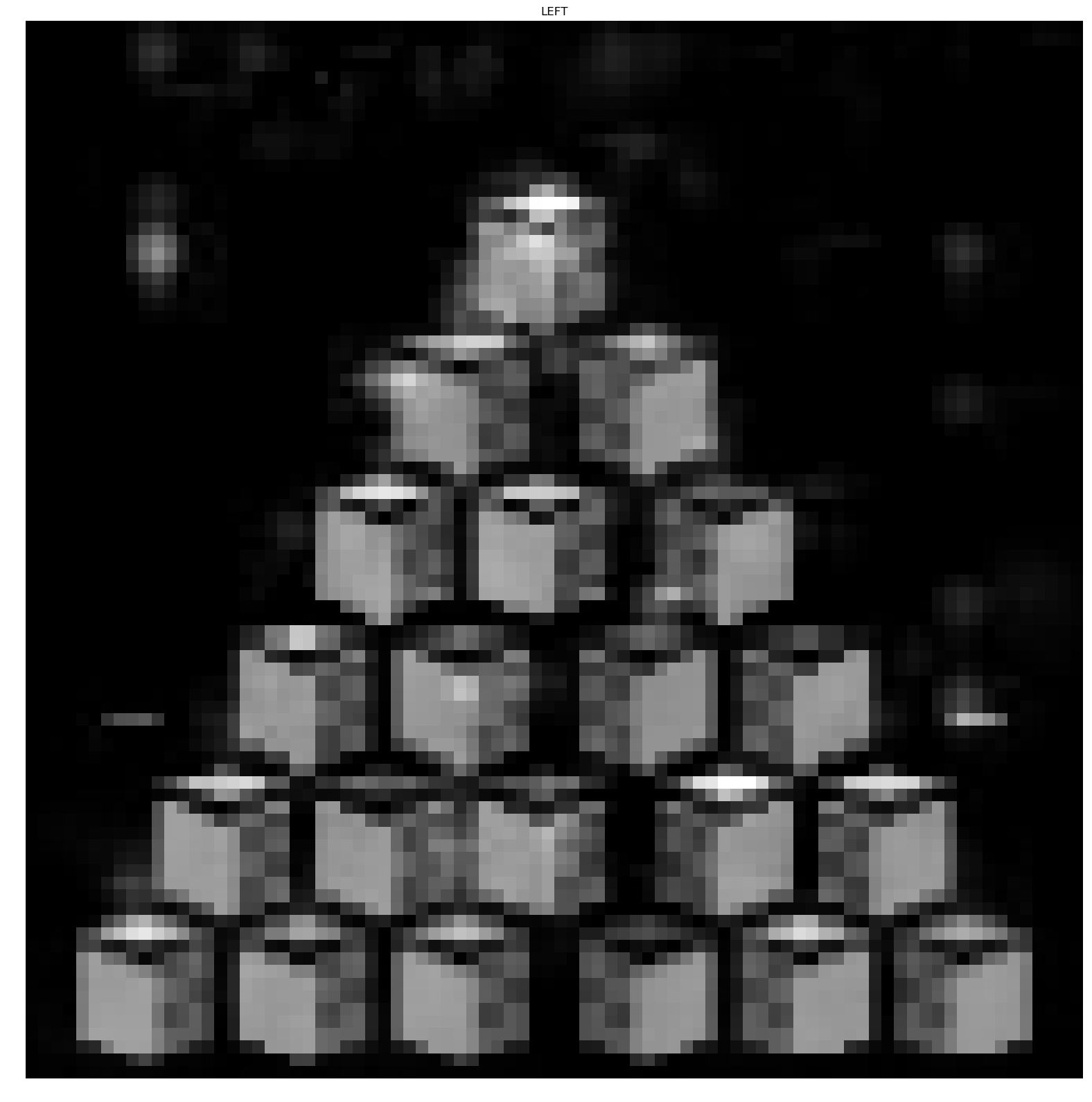}
    \captionsetup{labelformat=empty}
    \caption{Action: Left}
\end{subfigure} \hspace*{-.1em}s
\begin{subfigure}[t]{0.24\textwidth}
  \centering
    \includegraphics[width=1.0\textwidth]{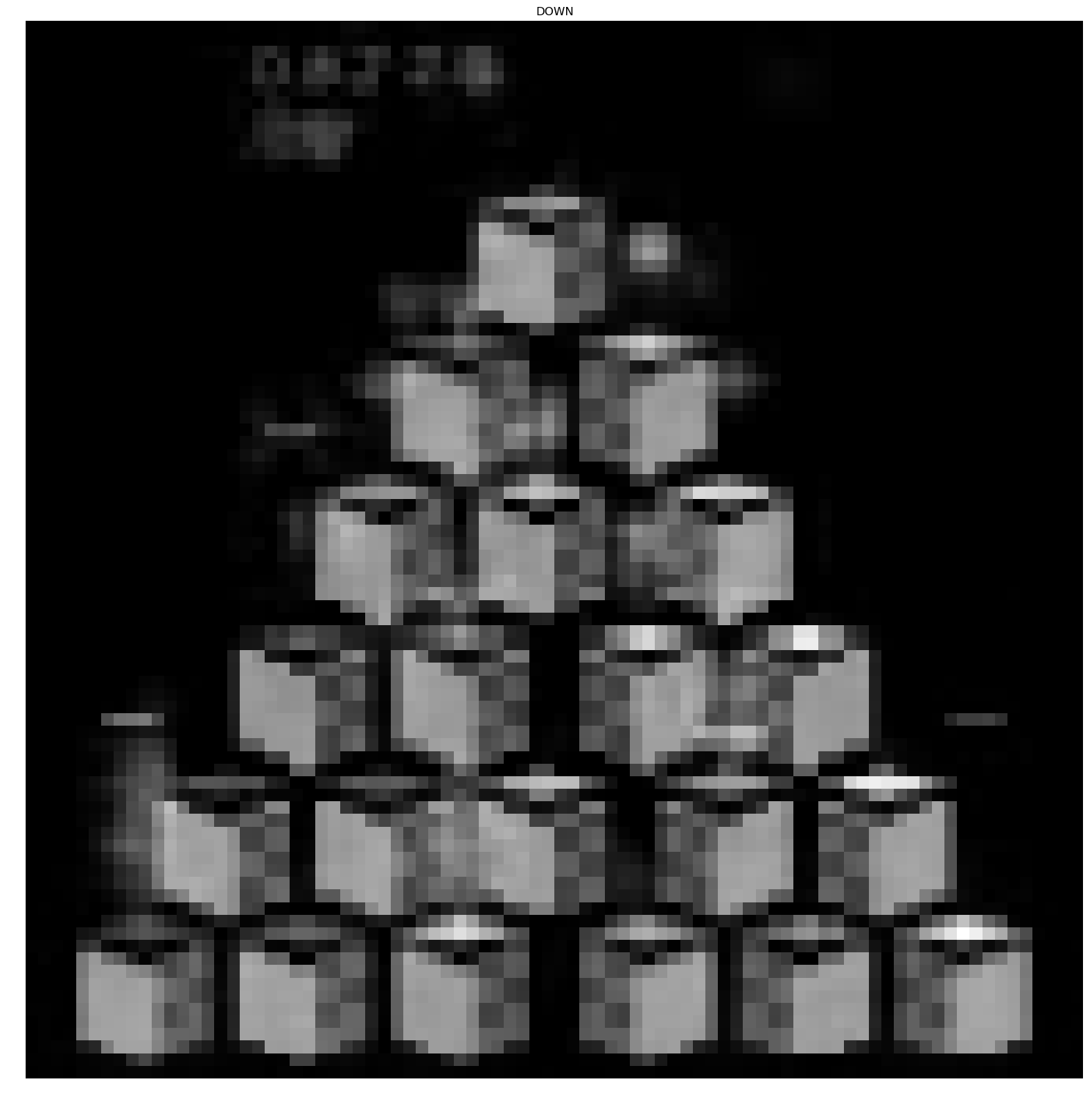}
    \captionsetup{labelformat=empty}
    \caption{Action: Down}
\end{subfigure}
\setcounter{subfigure}{1}
\caption{Qbert, Inverting keys for different actions with Q-value: 25}
\label{fig:invert-keys-25-qbert}
\end{subfigure}
\caption{Inverting Keys for Pong, Qbert}
\end{figure*}

\section{Inversion of Keys}
Figures~\ref{fig:invert-key-25-pacman},~\ref{fig:invert-keys-right-pacman} show the reconstructed images by inverting keys in MsPacman (some of these have been discussed in the paper). Figures~\ref{fig:invert-keys-25-pong},~\ref{fig:invert-keys-25-qbert} show the reconstructed images by inverting keys in Pong and Qbert. Most of these images are noisy, especially the ones in gray-scale but seem to indicate configurations of different objects in the game. Figure~\ref{fig:interpolate} shows the reconstructions by interpolating in the hidden space starting from a key embedding and moving towards a specific training state. We see that most of the hidden space can be reconstructed but the information about shapes and colors of the objects is not always clear while reconstructing the key embeddings.

\section{Adversarial Examples for MsPacman}
\begin{figure*}[t!]
\centering
\begin{subfigure}[t]{\textwidth}
\centering
\begin{subfigure}[t]{0.14\textwidth}
  \centering
    \includegraphics[width=1.0\textwidth]{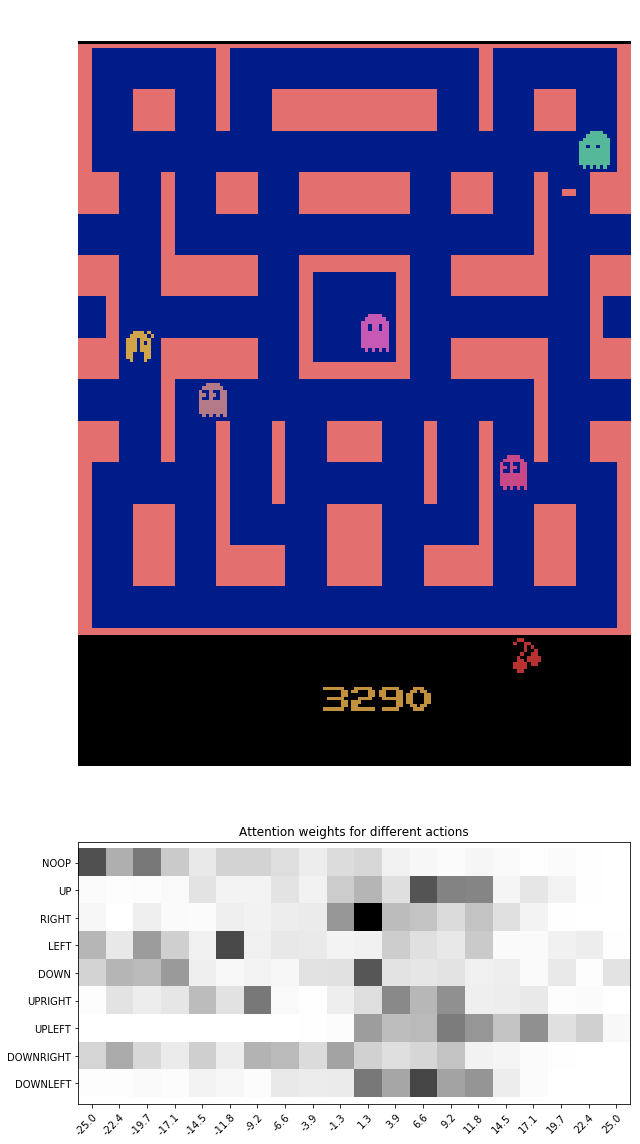}
\end{subfigure} \hspace*{-.5em}
\begin{subfigure}[t]{0.14\textwidth}
  \centering
    \includegraphics[width=1.0\textwidth]{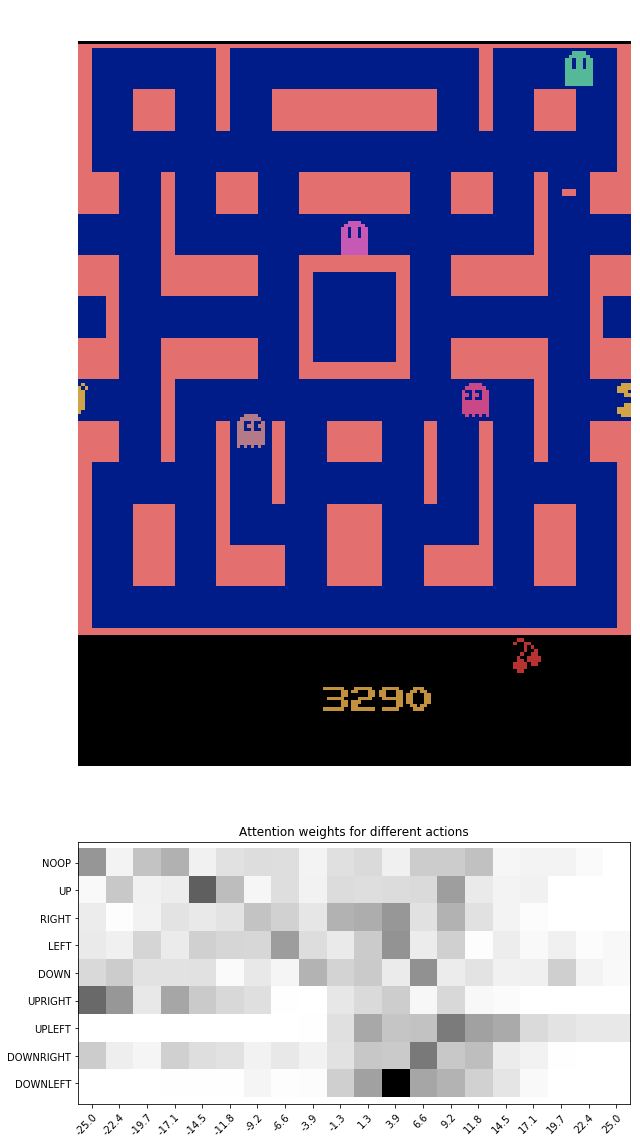}
\end{subfigure} \hspace*{-.5em}
\begin{subfigure}[t]{0.14\textwidth}
  \centering
    \includegraphics[width=1.0\textwidth]{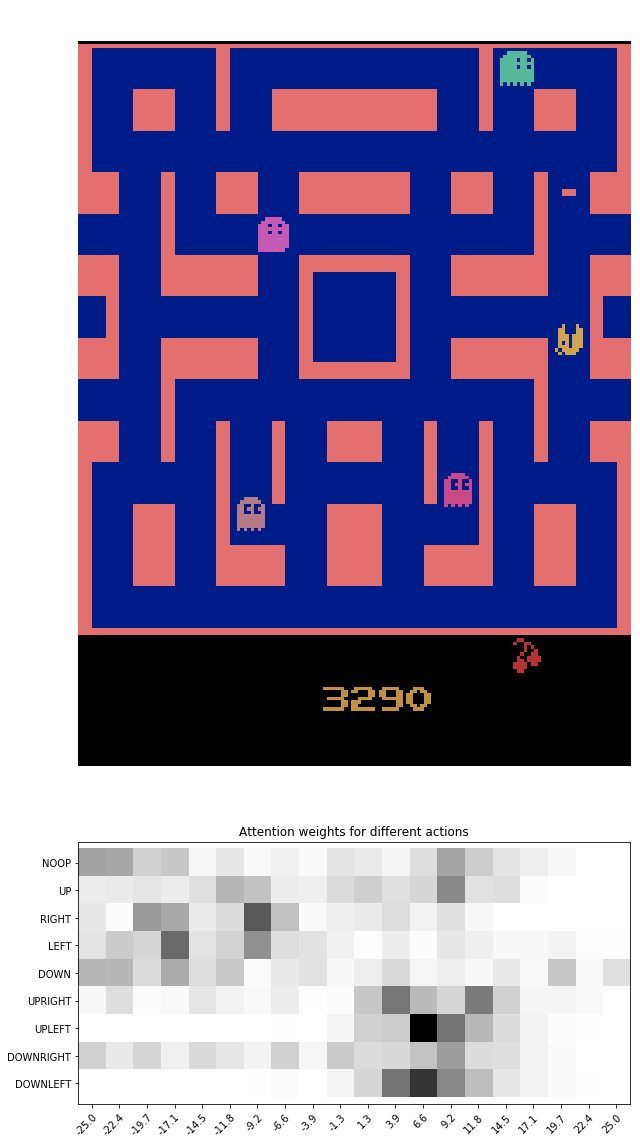}
\end{subfigure} \hspace*{-.5em}
\begin{subfigure}[t]{0.14\textwidth}
  \centering
    \includegraphics[width=1.0\textwidth]{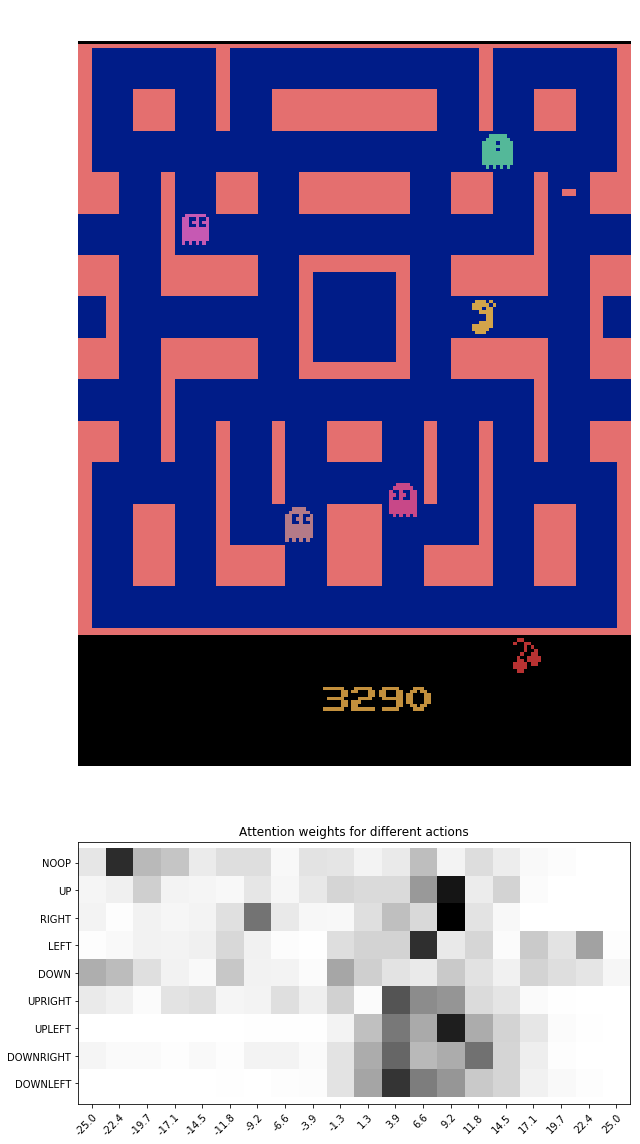}
\end{subfigure} \hspace*{-.5em}
\begin{subfigure}[t]{0.14\textwidth}
  \centering
    \includegraphics[width=1.0\textwidth]{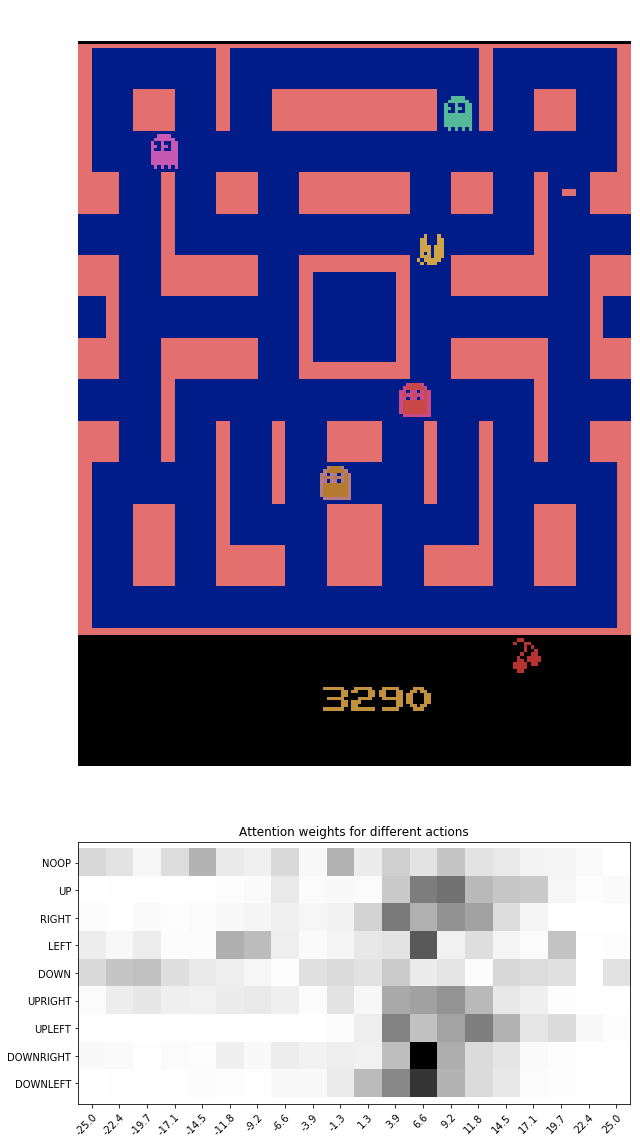}
\end{subfigure} \hspace*{-.5em}
\begin{subfigure}[t]{0.14\textwidth}
  \centering
    \includegraphics[width=1.0\textwidth]{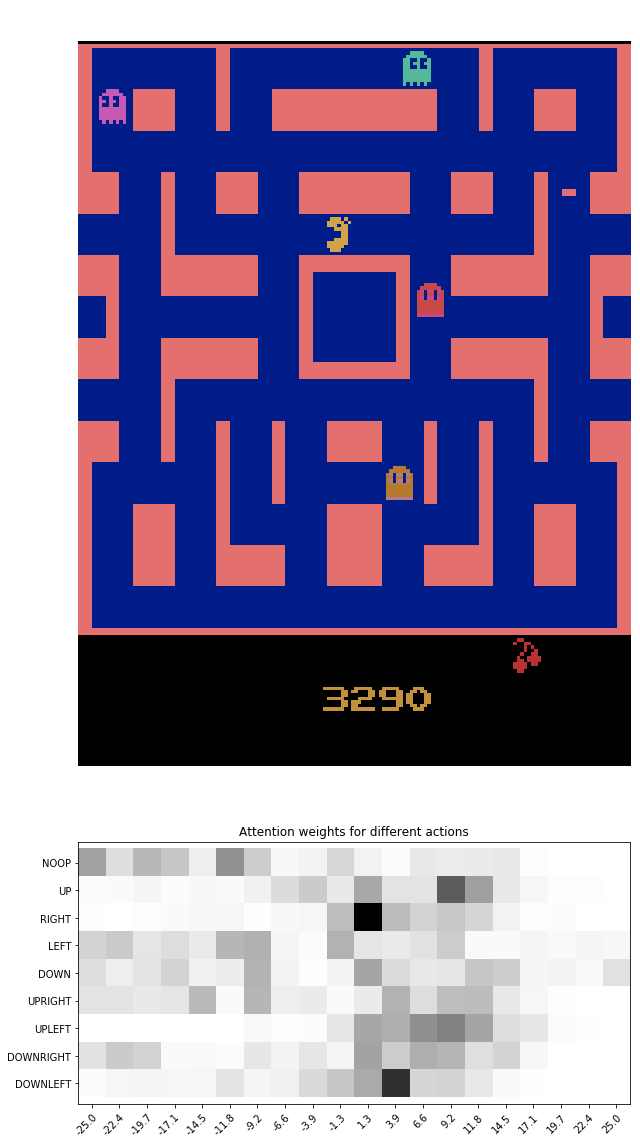}
\end{subfigure} \hspace*{-.5em}
\begin{subfigure}[t]{0.14\textwidth}
  \centering
    \includegraphics[width=1.0\textwidth]{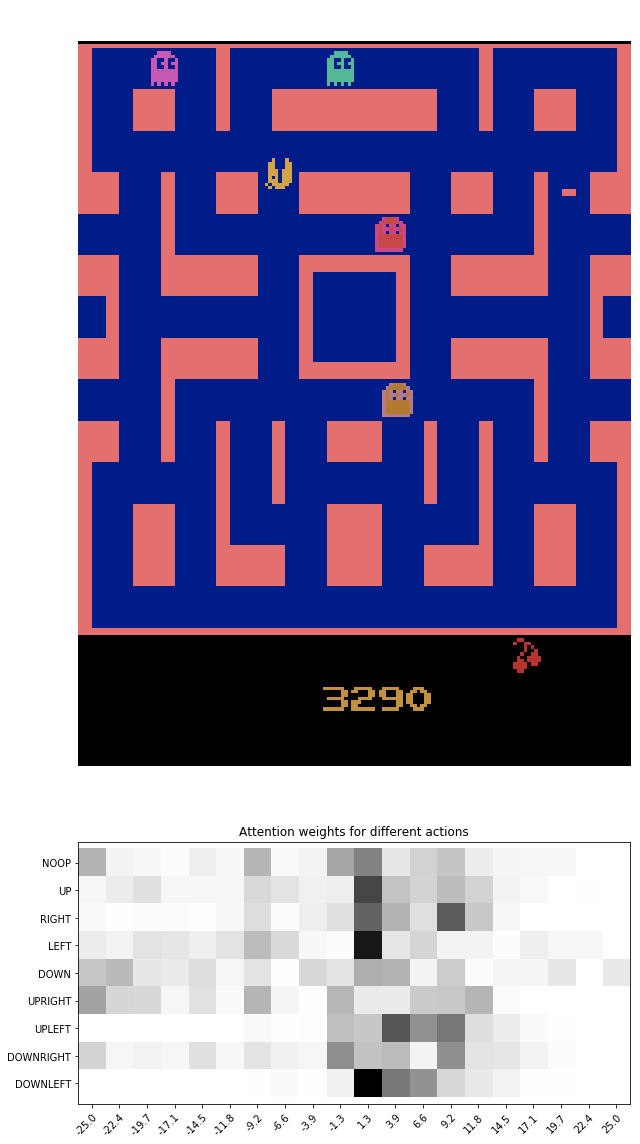}
\end{subfigure}
\caption{MsPacman, Adversarial example}
\end{subfigure}\\
\begin{subfigure}[t]{\textwidth}
\centering
\begin{subfigure}[t]{0.14\textwidth}
  \centering
    \includegraphics[width=1.0\textwidth]{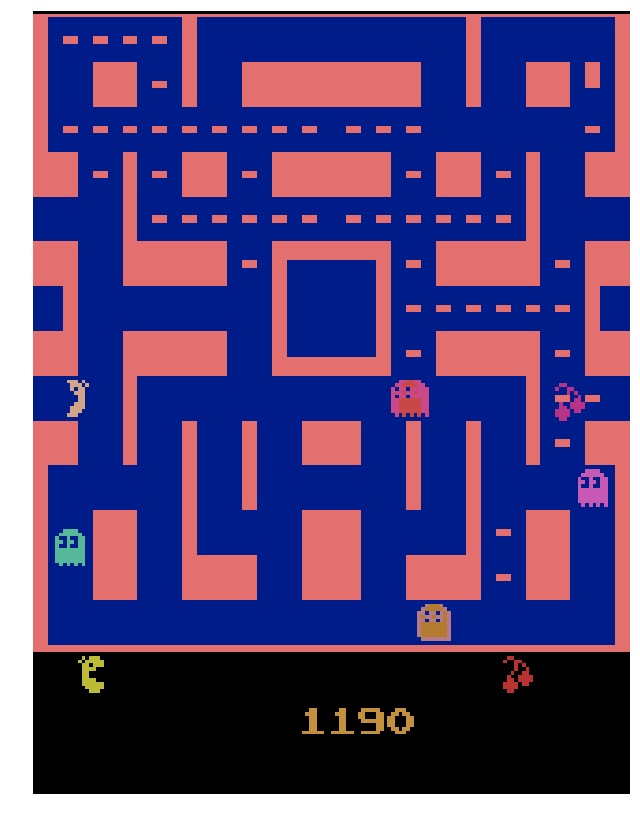}
\end{subfigure} \hspace*{-.5em}
\begin{subfigure}[t]{0.14\textwidth}
  \centering
    \includegraphics[width=1.0\textwidth]{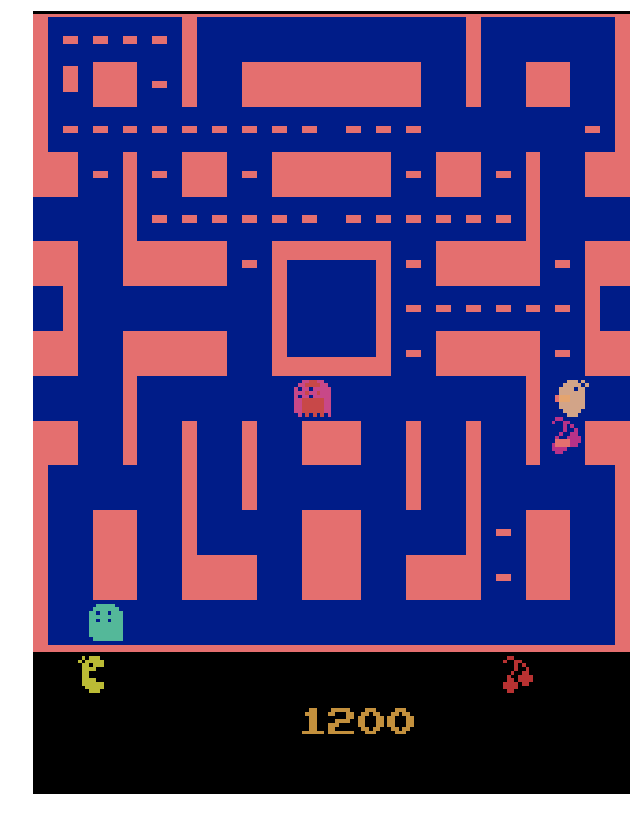}
\end{subfigure} \hspace*{-.5em}
\begin{subfigure}[t]{0.14\textwidth}
  \centering
    \includegraphics[width=1.0\textwidth]{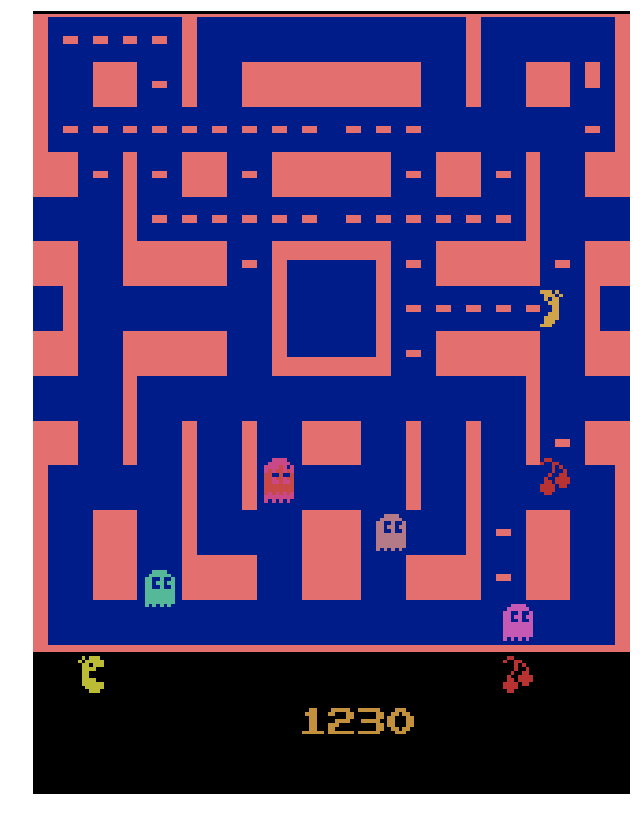}
\end{subfigure} \hspace*{-.5em}
\begin{subfigure}[t]{0.14\textwidth}
  \centering
    \includegraphics[width=1.0\textwidth]{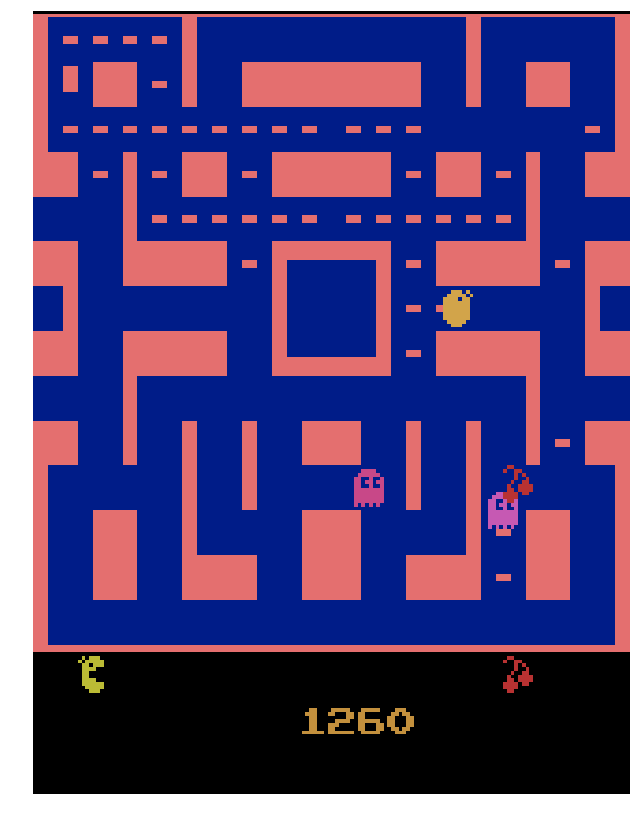}
\end{subfigure} \hspace*{-.5em}
\begin{subfigure}[t]{0.14\textwidth}
  \centering
    \includegraphics[width=1.0\textwidth]{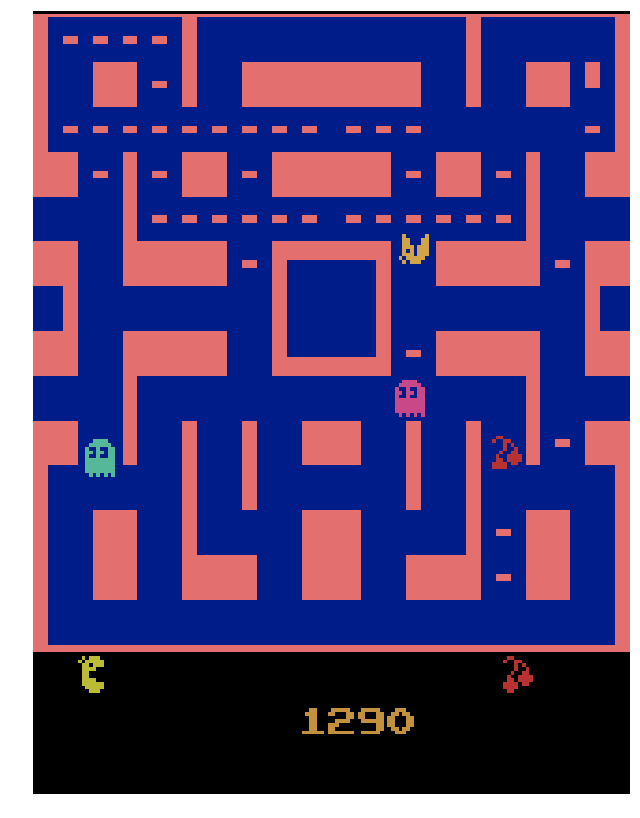}
\end{subfigure} \hspace*{-.5em}
\begin{subfigure}[t]{0.14\textwidth}
  \centering
    \includegraphics[width=1.0\textwidth]{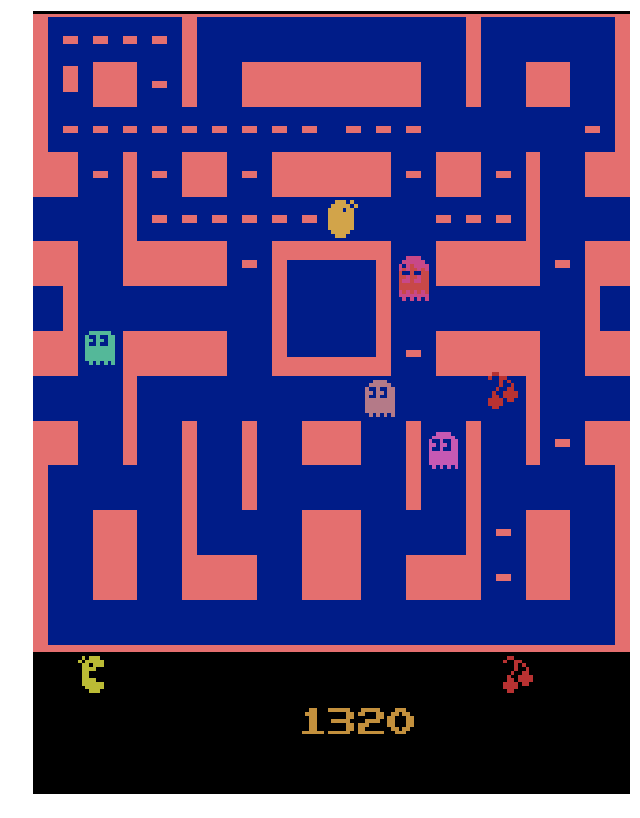}
\end{subfigure} \hspace*{-.5em}
\begin{subfigure}[t]{0.14\textwidth}
  \centering
    \includegraphics[width=1.0\textwidth]{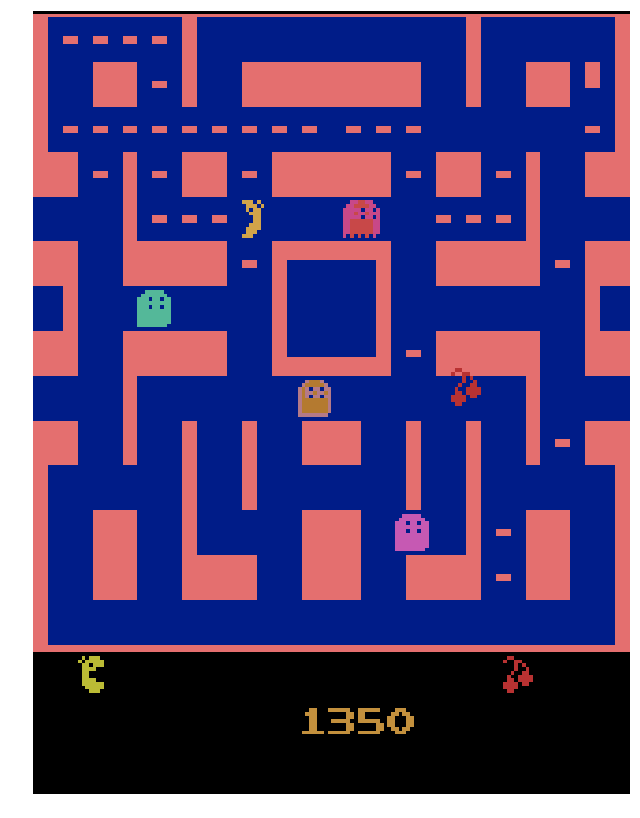}
\end{subfigure}
\caption{MsPacman, Trajectory during training}
\end{subfigure}\\
\begin{subfigure}[t]{\textwidth}
\centering
\begin{subfigure}[t]{0.14\textwidth}
  \centering
    \includegraphics[width=1.0\textwidth]{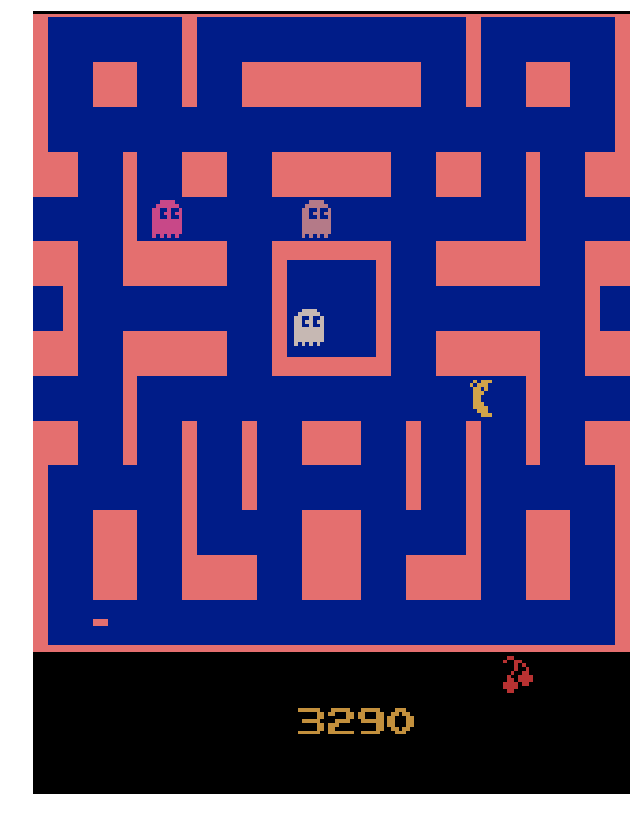}
\end{subfigure} \hspace*{-.5em}
\begin{subfigure}[t]{0.14\textwidth}
  \centering
    \includegraphics[width=1.0\textwidth]{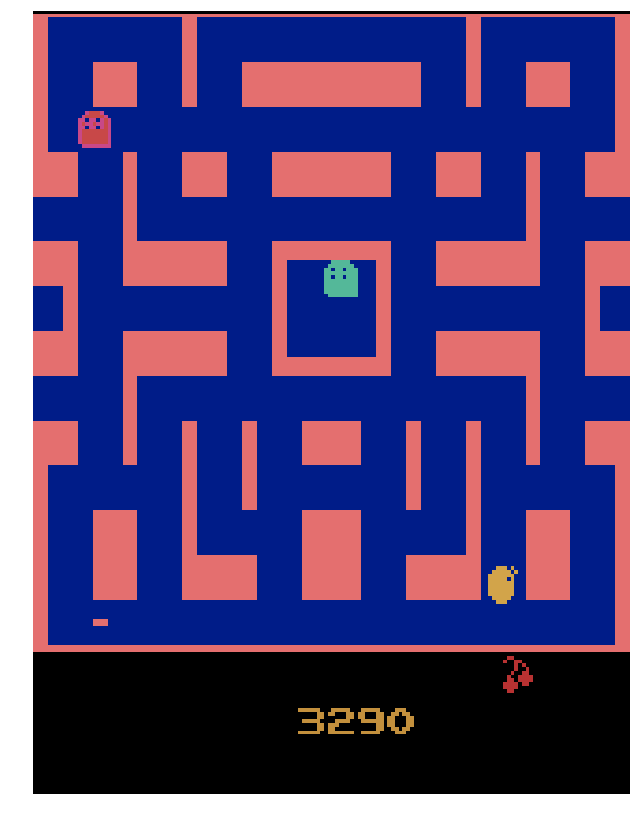}
\end{subfigure} \hspace*{-.5em}
\begin{subfigure}[t]{0.14\textwidth}
  \centering
    \includegraphics[width=1.0\textwidth]{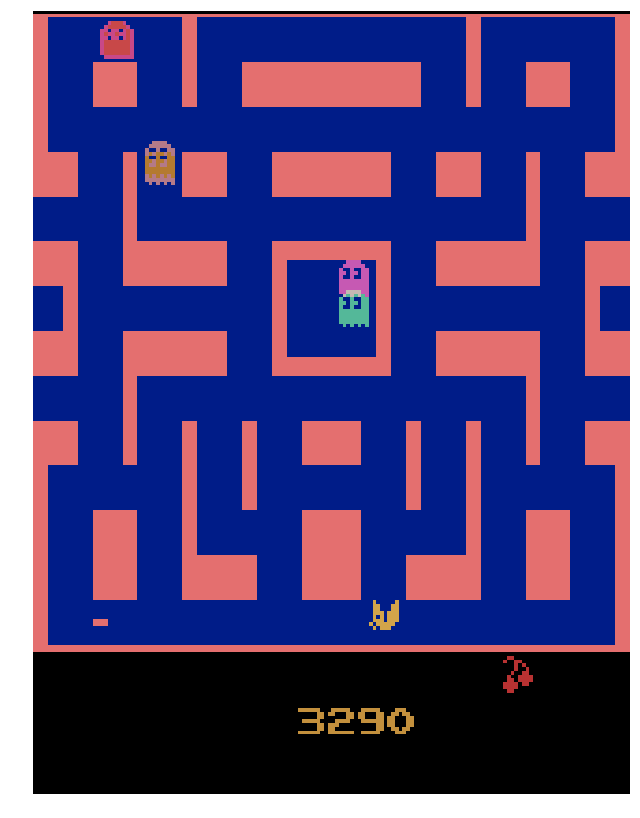}
\end{subfigure} \hspace*{-.5em}
\begin{subfigure}[t]{0.14\textwidth}
  \centering
    \includegraphics[width=1.0\textwidth]{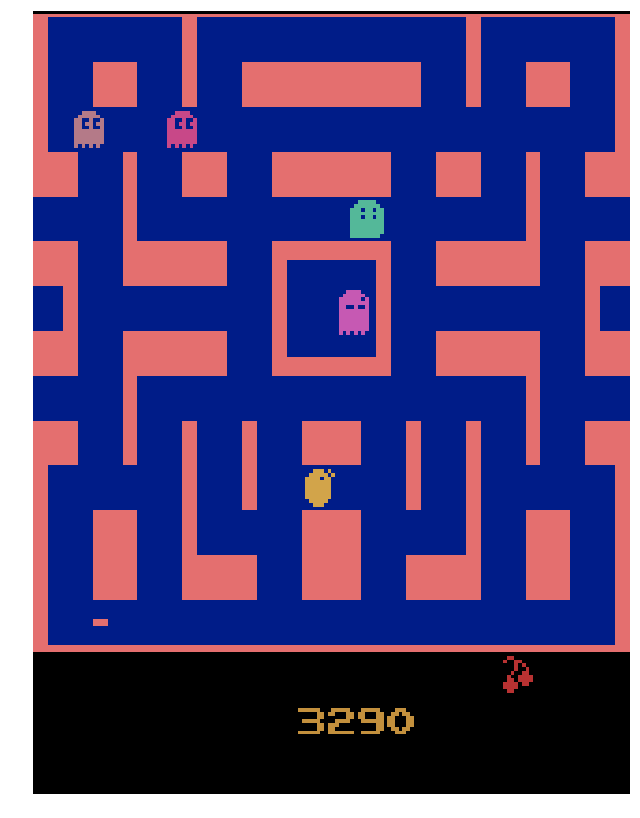}
\end{subfigure} \hspace*{-.5em}
\begin{subfigure}[t]{0.14\textwidth}
  \centering
    \includegraphics[width=1.0\textwidth]{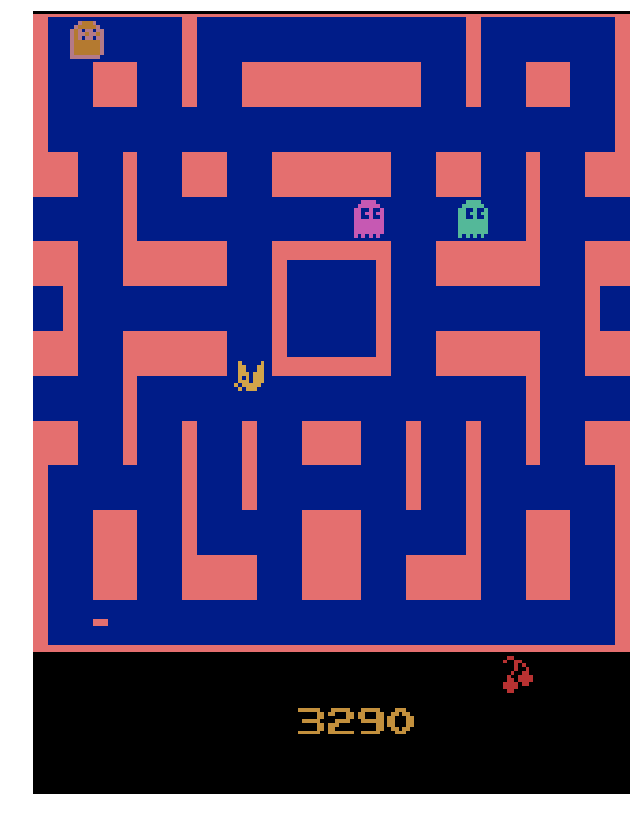}
\end{subfigure} \hspace*{-.5em}
\begin{subfigure}[t]{0.14\textwidth}
  \centering
    \includegraphics[width=1.0\textwidth]{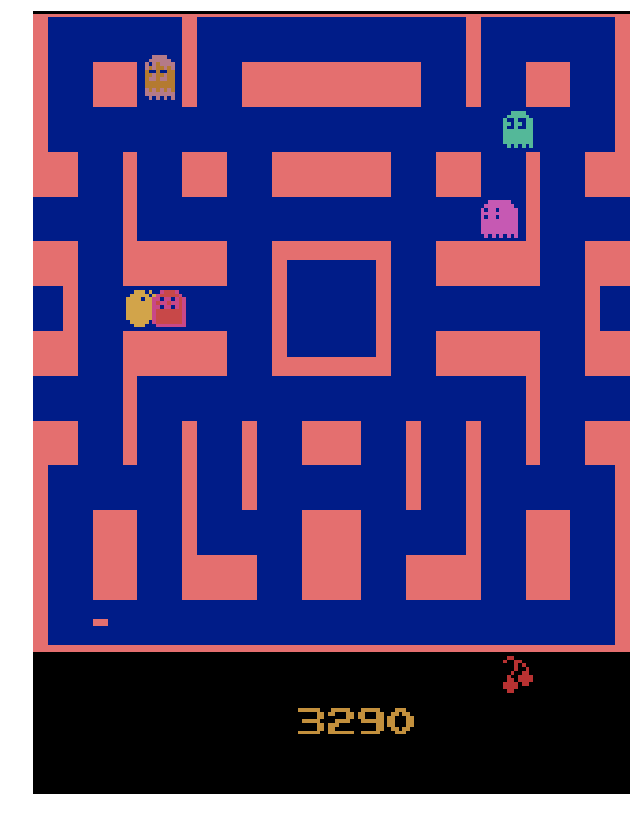}
\end{subfigure}
\begin{subfigure}[t]{0.14\textwidth}
  \centering
    \includegraphics[width=1.0\textwidth]{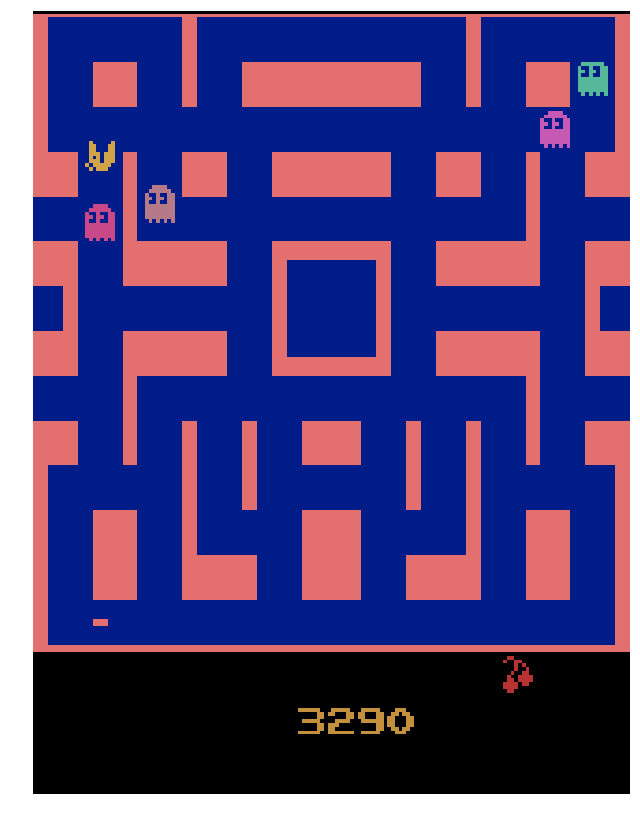}
\end{subfigure}
\caption{MsPacman, Adversarial Example}
\end{subfigure} \\
\begin{subfigure}[t]{\textwidth}
\centering
\begin{subfigure}[t]{0.14\textwidth}
  \centering
    \includegraphics[width=1.0\textwidth]{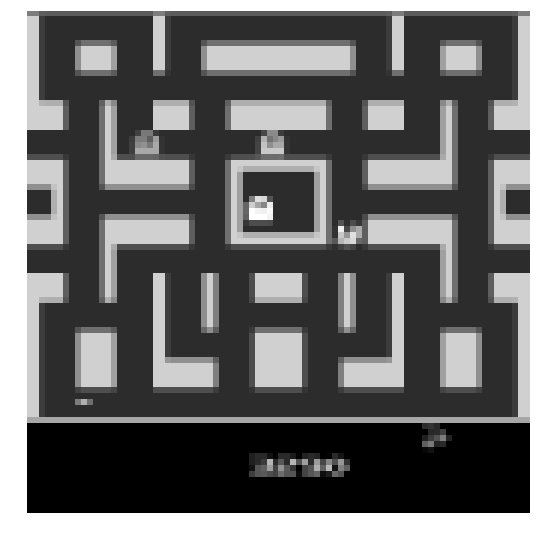}
\end{subfigure} \hspace*{-.5em}
\begin{subfigure}[t]{0.14\textwidth}
  \centering
    \includegraphics[width=1.0\textwidth]{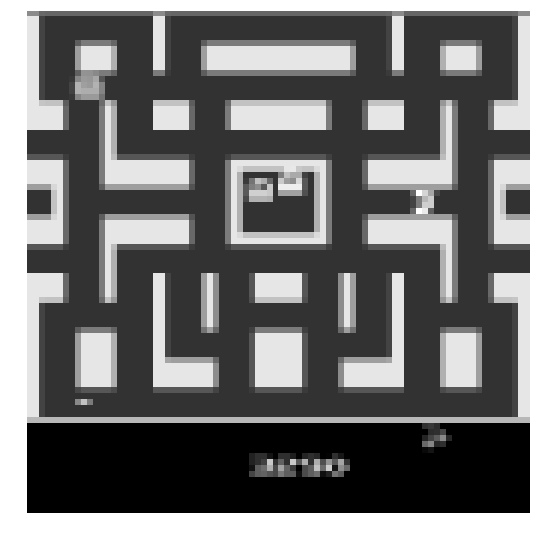}
\end{subfigure} \hspace*{-.5em}
\begin{subfigure}[t]{0.14\textwidth}
  \centering
    \includegraphics[width=1.0\textwidth]{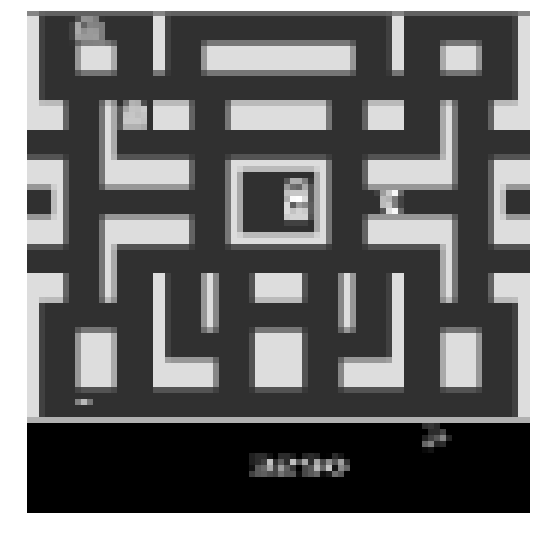}
\end{subfigure} \hspace*{-.5em}
\begin{subfigure}[t]{0.14\textwidth}
  \centering
    \includegraphics[width=1.0\textwidth]{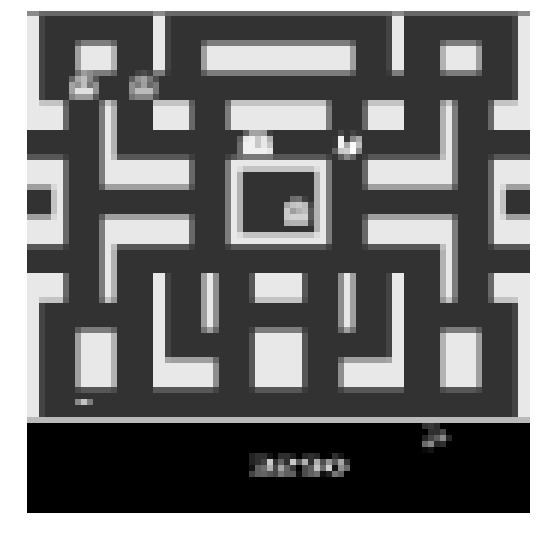}
\end{subfigure} \hspace*{-.5em}
\begin{subfigure}[t]{0.14\textwidth}
  \centering
    \includegraphics[width=1.0\textwidth]{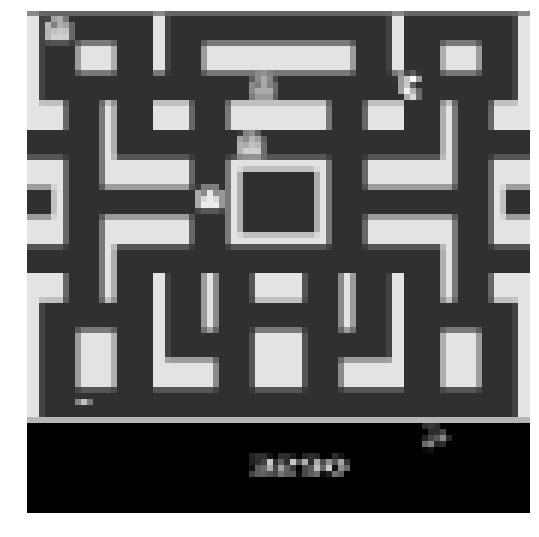}
\end{subfigure} \hspace*{-.5em}
\begin{subfigure}[t]{0.14\textwidth}
  \centering
    \includegraphics[width=1.0\textwidth]{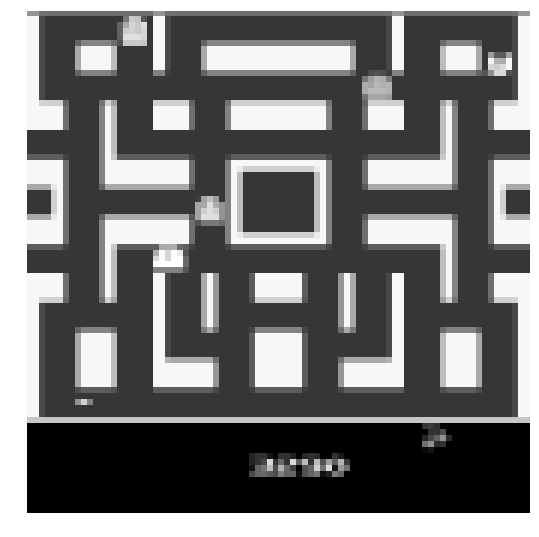}
\end{subfigure}
\begin{subfigure}[t]{0.14\textwidth}
  \centering
    \includegraphics[width=1.0\textwidth]{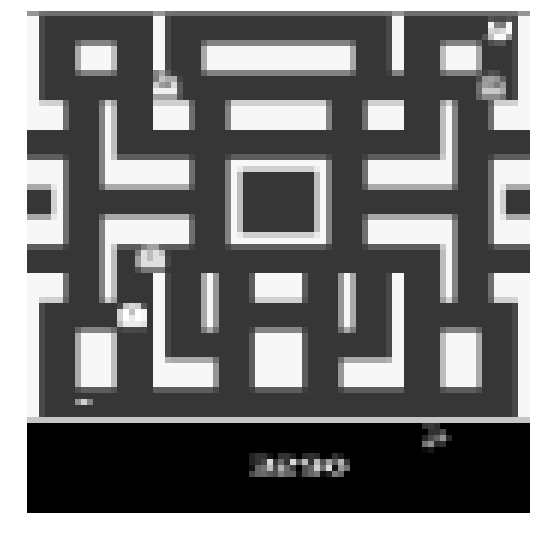}
\end{subfigure}
\caption{MsPacman, Adversarial Example for Double DQN model}
\end{subfigure}
\caption{Adversarial Examples}
\label{fig:adversarial_examples}
\end{figure*}

\begin{figure*}[t!]
\centering
\begin{subfigure}[t]{0.24\textwidth}
  \centering
    \includegraphics[width=1.0\textwidth]{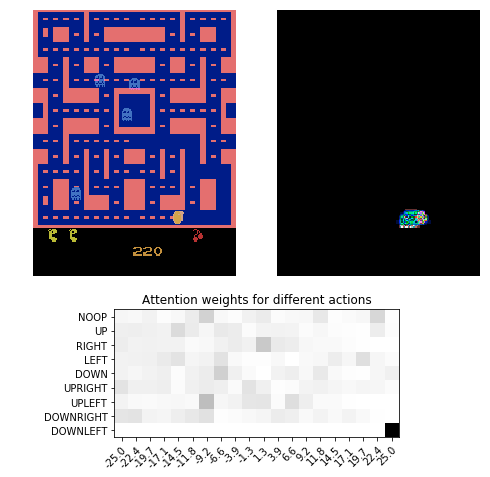}
    \caption{Action Selected: Downleft}
\end{subfigure} \hspace*{-.5em}
\begin{subfigure}[t]{0.24\textwidth}
  \centering
    \includegraphics[width=1.0\textwidth]{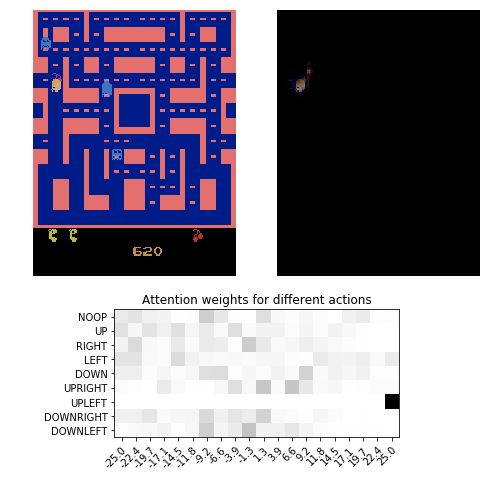}
    \caption{Action Selected: Upleft}
\end{subfigure} \hspace*{-.5em}
\begin{subfigure}[t]{0.24\textwidth}
  \centering
    \includegraphics[width=1.0\textwidth]{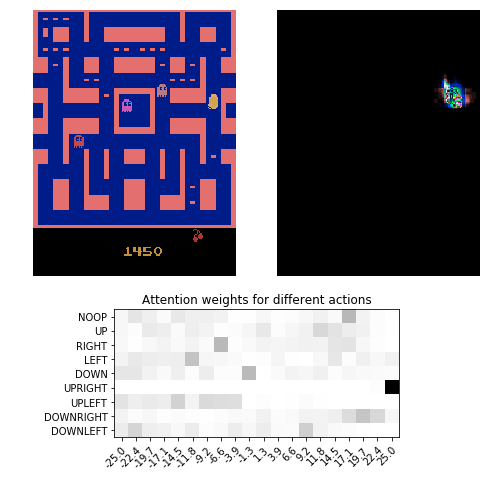}
    \caption{Action Selected: Upright}
\end{subfigure} \hspace*{-.5em}
\begin{subfigure}[t]{0.24\textwidth}
  \centering
    \includegraphics[width=1.0\textwidth]{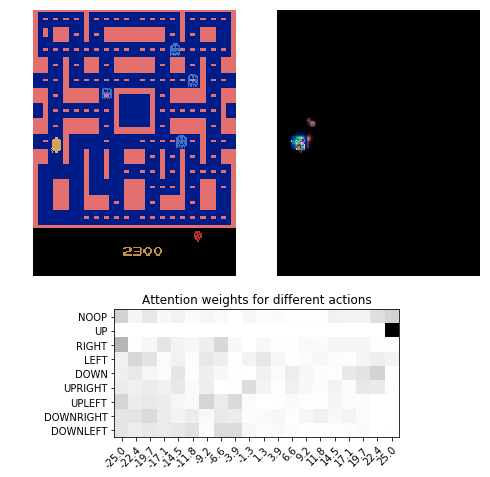}
    \caption{Action Selected: Up}
\end{subfigure} 
\caption{MsPacman, Samples from training trajectory (with attention and saliency maps)}
\label{fig:int-adv-training}
\begin{subfigure}[t]{0.24\textwidth}
  \centering
    \includegraphics[width=1.0\textwidth]{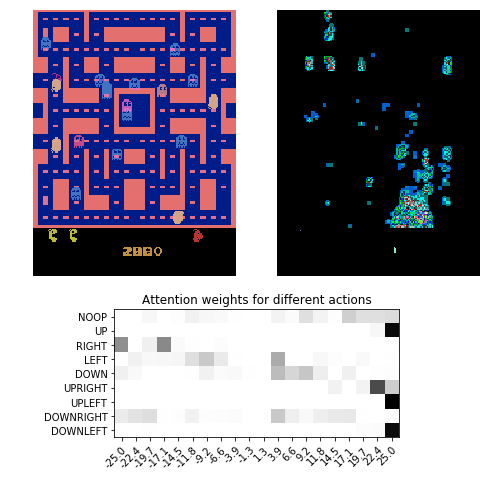}
\end{subfigure} \hspace*{-.5em}
\begin{subfigure}[t]{0.24\textwidth}
  \centering
    \includegraphics[width=1.0\textwidth]{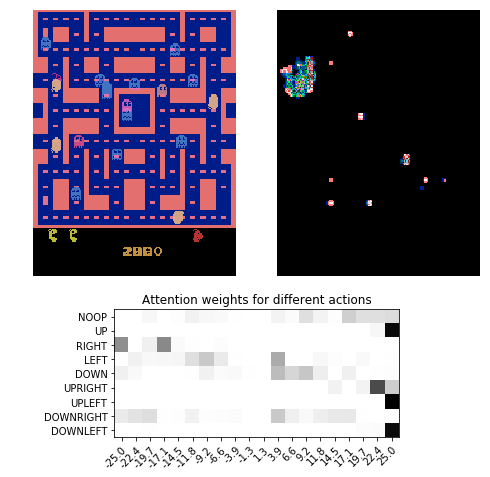}
\end{subfigure}
\begin{subfigure}[t]{0.24\textwidth}
  \centering
    \includegraphics[width=1.0\textwidth]{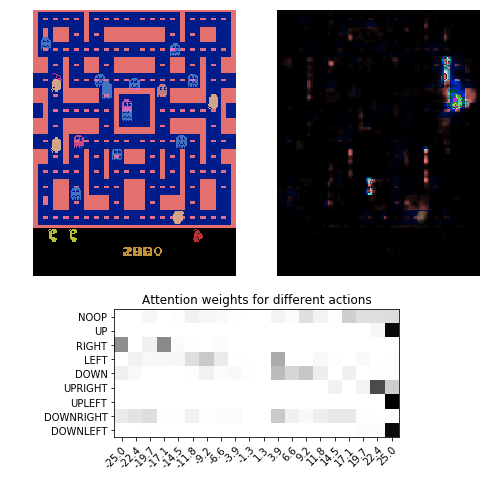}
\end{subfigure}
\begin{subfigure}[t]{0.24\textwidth}
  \centering
    \includegraphics[width=1.0\textwidth]{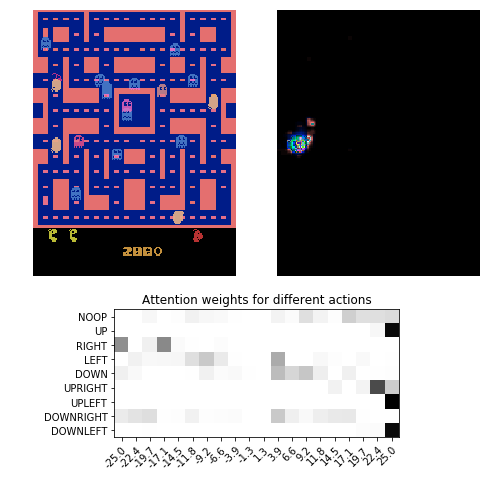}
\end{subfigure}
\caption{MsPacman, Adversarial example generated by pixelwise-max operation over samples in Figure~\ref{fig:int-adv-training}. Such a state although unseen during training, is probably meaningless. But the saliency maps can still recover the same regions almost perfectly (with extra noise) by taking gradients with respect to the particular actions.}
\label{fig:int-adversarial-example}
\end{figure*}

\begin{figure*}[t!]
\centering
\begin{subfigure}[t]{\textwidth}
    \centering
    \includegraphics[width=1.0\textwidth]{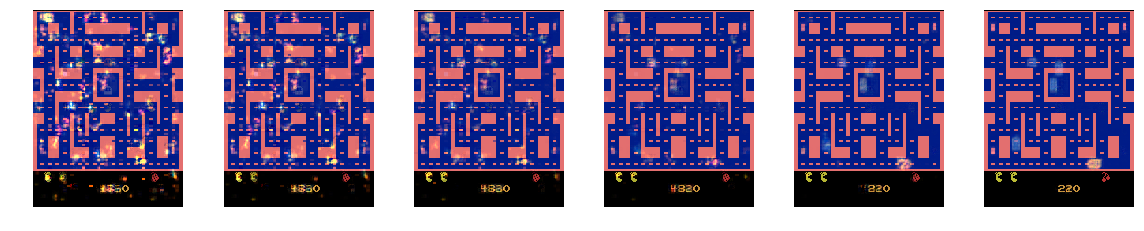}
\end{subfigure}
\begin{subfigure}[t]{\textwidth}
    \centering
    \includegraphics[width=1.0\textwidth]{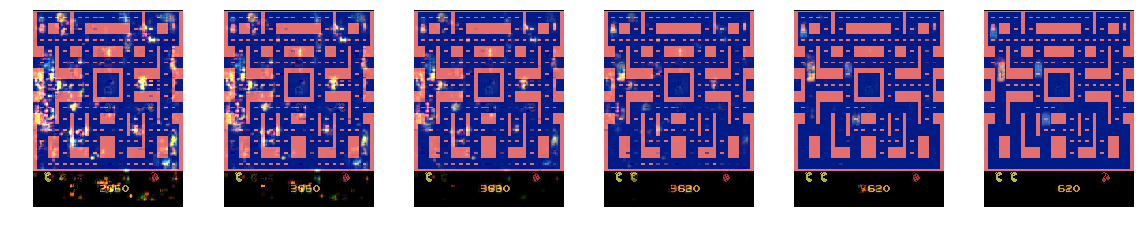}
\end{subfigure}
\begin{subfigure}[t]{\textwidth}
    \centering
    \includegraphics[width=1.0\textwidth]{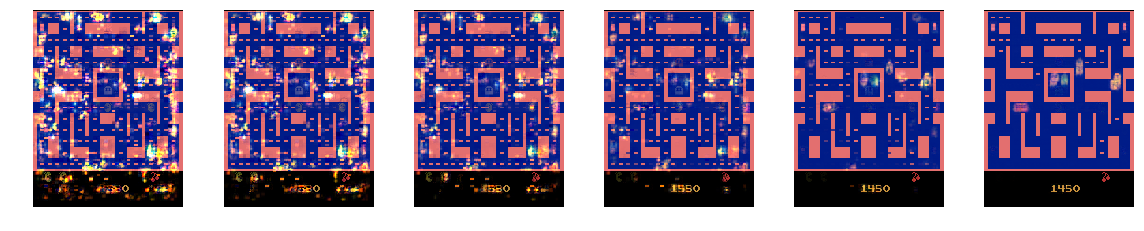}
\end{subfigure}
\begin{subfigure}[t]{\textwidth}
    \centering
    \includegraphics[width=1.0\textwidth]{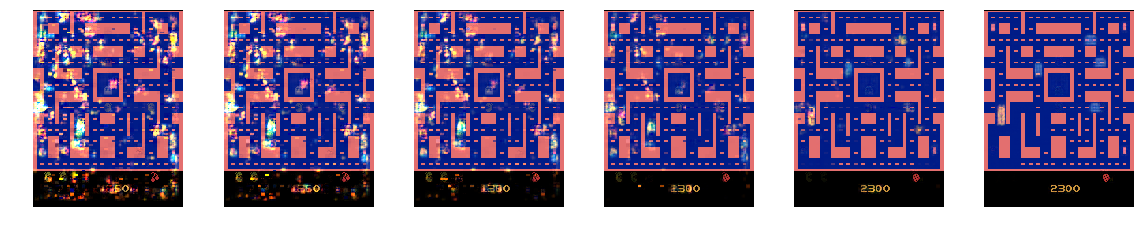}
\end{subfigure}
\caption{MsPacman, Interpolating in the hidden space; start embeddings are keys for Downleft, Upleft, Upright, Up (First Column); final embeddings are the same states as those in Figure~\ref{fig:int-adv-training} with channel-wise max over 4 frames instead of single frame (Last Column); We use a linear interpolation scheme as follows, $embedding = start + \lambda(final - start) $, $\lambda = \{0, 0.2, 0.4, 0.6, 0.8, 1.0\}$ and invert each $embedding$.}
\label{fig:interpolate}
\end{figure*}

In Figure~\ref{fig:adversarial_examples}, we show some more examples for MsPacman where the agent does not learn to clear a pellet or goes around repeating action sequences performed during training. Note that Open AI gym environment does not provide access to the state-space and the required out-of-sample state needs to be generated via human gameplay. Figures~\ref{fig:int-adv-training},~\ref{fig:int-adversarial-example} present an interesting adversarial example created by taking the pixel-wise maximum over real training states. Even though this state is not a valid state, we see that in the attention-maps, all of the four selected actions are strongly highlighted. In fact, by computing the saliency maps with respect to each of these actions, we can almost get back the same regions that are highlighted in Figure~\ref{fig:int-adv-training}. This suggests that the key embeddings seem to independently capture specific regions of the board where Pacman is present without modeling any high level dependencies among objects.

\end{document}